%% file: main.tex
\pdfoutput=1
\def\PREPRINT{}
\documentclass[10pt]{article}
\newif\ifpreprint
\ifdefined\PREPRINT\preprinttrue\fi
\ifpreprint\usepackage[preprint]{tmlr}\else\usepackage{tmlr}\fi
\usepackage{amsmath,amssymb,amsthm,mathtools}
\usepackage{graphicx,booktabs,xcolor}
\usepackage[colorlinks=true,linkcolor=blue,citecolor=blue,urlcolor=blue]{hyperref}
\usepackage{cleveref}

\newtheorem{proposition}{Proposition}
\newtheorem{lemma}{Lemma}
\newtheorem{corollary}{Corollary}
\theoremstyle{definition}
\newtheorem{definition}{Definition}
\newtheorem{assumption}{Assumption}
\theoremstyle{remark}
\newtheorem{remark}{Remark}

\newcommand{\E}{\mathbb{E}}
\newcommand{\KL}{\mathrm{KL}}
\newcommand{\kl}{\mathrm{kl}}
\newcommand{\Bern}{\mathrm{Bern}}
\newcommand{\Bin}{\mathrm{Bin}}
\newcommand{\BetaD}{\mathrm{Beta}}
\newcommand{\eCE}{\mathrm{eCE}}
\newcommand{\lev}{\mathrm{lev}}
\newcommand{\nlev}{|\mathcal X|}
\ifpreprint
\newcommand{\pkgname}{mitarget}
\newcommand{\pkgurl}{\url{https://github.com/damianghl/mitarget}}
\else
\newcommand{\pkgname}{sparsete}
\fi

\title{When is simple target encoding optimal?\\
\large Moment exactness for sparse categorical features, and the limits of estimating their information}
\author{\name Dami\'an G. Hern\'andez \email damian.hernandez@ib.edu.ar \\
\addr Department of Applications of Physics and Biology to Health Sciences, Centro At\'omico Bariloche, Argentina \\
\addr Instituto Balseiro, Centro At\'omico Bariloche, Argentina}

\begin{document}
\maketitle

\input{sections/00_abstract}
\input{sections/01_introduction}
\input{sections/02_related_work}
\input{sections/03_setup}
\input{sections/04_moment_exactness}

\input{sections/05_popularity}
\input{sections/06_mutual_information}
\input{sections/07_experiments}
\input{sections/08_mi_experiments}
\input{sections/09_discussion}

\bibliographystyle{tmlr}
\bibliography{refs}

\appendix
\input{sections/appendix}

\end{document}

%% file: sections/00_abstract.tex
\begin{abstract}
Target encoding replaces each level of a high-cardinality categorical feature by a smoothed estimate of the target rate. We ask how well any encoder can do when most test rows fall on levels seen zero, one or two times, what a simple encoder misses on real data, and whether the information $I(X;Y)$ the feature carries can be estimated in the same regime. The Bayes prediction after $n$ observations depends only on the first $n+1$ moments of the prior of the target rates. Hence, when the target rate of a level does not depend on its popularity (exchangeability), a beta prior matched to two moments is exactly the oracle on levels seen at most once; fitting it by type-II maximum likelihood recovers it asymptotically when no level is seen more than twice, and stays numerically close otherwise when levels are equally popular. A moment test rejects exchangeability on 21 of 23 real features. Without it, the oracle prior for levels seen $n$ times is the law of the target rate among levels seen $n+1$ times; this shift gives a correction for levels seen at most once that is read off the singleton and doubleton counts, and a frequency-aware prior that closes half to four-fifths of the remaining gap on the real-data subsampling oracles with popularity dependence. Information is harder: from levels seen at most $L$ times, $I(X;Y)$ is only partially identified, within a worst-case half-width of $0.038$ nats for $L=2$ and about $0.22/L^2$ nats for larger even $L$, and without exchangeability only a trivial bound is identified for the singletons' share. In single-feature prediction on 24 real features, the fitted priors (beta, NPMLE, posterior-mean GLMM, frequency-aware, beta with the correction) differ from each other by less than $0.01$ nats per test row, though by up to $15\%$ of the beta's gain on some sparse features; larger losses come from a fixed smoothing strength far from the fitted one, such as the common default $m=20$ on sparse click features, and from predicting a GLMM at its conditional mode. A likelihood-region interval for $I(X;Y)$ covers in our synthetic studies and on all but one exchangeable real-data pool; on real features, where exchangeability fails, stratified identified intervals, which assume it only within groups of level sizes and ignore sampling error, are several hundredths of a nat wide on sparse features and up to $0.45$ nats on the sparsest.
\end{abstract}

%% file: sections/01_introduction.tex
\section{Introduction}\label{sec:intro}

Categorical features with very many levels are everywhere in applied prediction: user, device and advertiser identifiers in click logs, product codes, postal codes, job titles. One-hot encoding them is wasteful or impossible, and the standard alternative is \emph{target encoding}: each level $x$ is replaced by a smoothed estimate of the target rate $P(Y=1\mid X=x)$, which a downstream model then uses as a single numeric feature \citep{micci2001}. The usual estimate shrinks the level's empirical rate $k_x/n_x$ towards the global rate $\hat\mu$,
\[
r_x=\frac{k_x+m\,\hat\mu}{n_x+m},
\]
with a smoothing strength $m$ that is fixed by hand or tuned by cross-validation. Benchmarks find that this regularisation matters: plain level means overfit, and a random-intercept GLMM encoder was the best method in the comparison of \citet{pargent2022}.

In the regime where these features are hard, the levels that carry the test mass are small. By the Good--Turing identity \citep{good1953}, the share of test rows that fall on levels seen $n$ times in training is estimated by $(n+1)F_{n+1}/N$, where $F_j$ is the number of levels seen exactly $j$ times among $N$ training rows. The share on levels seen at most once is therefore $(F_1+2F_2)/N$, computable from the training counts alone. On the click-log features and Book-Crossing ISBNs of \cref{sec:bakeoff} it is between $0.17$ and $0.42$ on the full data: a sixth to two fifths of all test predictions are made for levels with zero or one training observation. This raises three questions.\footnote{\ifpreprint The author used Claude (Anthropic), a large language model, to assist with writing, coding and analysis. The author reviewed all content and takes full responsibility for it.\else The authors used Claude (Anthropic), a large language model, to assist with writing, coding and analysis. The authors reviewed all content and take full responsibility for it.\fi}
\begin{enumerate}
\item[(a)] What is the best log-loss any encoder can achieve on such a feature, and how close does a simple smoothed encoder get?
\item[(b)] When a simple encoder falls short on real data, what is it missing?
\item[(c)] Can the information the feature carries about the target, $I(X;Y)$, be estimated in the same regime?
\end{enumerate}

\begin{figure}[t]
\centering
\includegraphics[width=\linewidth]{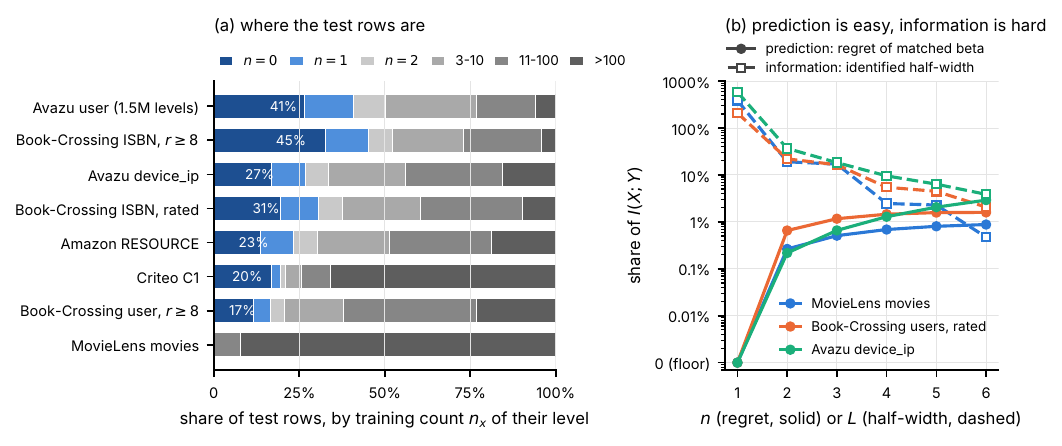}
\caption{Prediction is easy, information is hard. (a) Share of held-out test rows by the number of times their level was seen in training, for real high-cardinality features (5-fold cross-validation, \cref{sec:bakeoff}). On levels seen at most once (blue) a beta prior matched to two moments is exactly the Bayes oracle when the conditionals do not depend on popularity (\cref{prop:exact}); when they do, this is the mass on which the corrections of \cref{sec:popularity} act. (b) On the row-weighted (popularity-blind) distribution of the conditionals $q_x$ of three real-data pools (\cref{sec:oracle}), with infinitely many levels: regret of the moment-matched beta on a level seen $n$ times (solid; zero at $n\le1$, plotted at the bottom), and half-width of the interval of values of $I(X;Y)$ consistent with levels seen at most $L$ times (dashed), both as a share of $I(X;Y)$. These pools are exchangeable by construction. The matched beta loses under $1\%$ of $I(X;Y)$ on levels seen twice, while levels seen at most twice leave $I(X;Y)$ in an interval of half-width $19$--$37\%$ of its value, not centred on it (\cref{tab:teaser}); the solid curve is one prior's regret, the dashed one a worst case over priors with the same moments.}\label{fig:teaser}
\end{figure}

\paragraph{Prediction is easy, information is hard.} Our answer to (a) has a theoretical and an empirical part. Theoretically, on levels seen at most once the simple encoder, with $m$ fitted rather than fixed, is asymptotically the Bayes oracle when the target rate of a level does not depend on how popular the level is (exchangeability, Assumption~\ref{ass:exch}); this assumption is rarely satisfied on real data. Empirically, on 24 real features richer fitted priors and popularity corrections beat it significantly on about half the tasks, but by at most about $0.01$ nats per test row. The Bayes prediction for a level seen $n$ times depends on the prior of the conditionals $q_x$ only through its first $n+1$ moments (\cref{lem:moments}). A beta prior has two free moments, so a beta matched to the mean and variance of the true prior gives exactly the oracle prediction on every level seen at most once, and on levels seen twice when the prior is symmetric about $1/2$ (\cref{prop:exact,cor:sym}). Fitting its concentration by type-II maximum likelihood finds this beta when every level is seen at most twice (\cref{prop:ml}), and in our numerical checks with equally popular levels stays within $5\times10^{-5}$ nats of it on the levels seen at most once otherwise; heavy-tailed level sizes pull the fit further away (\cref{rem:mixed}). Moments beyond the largest level size cannot be learned by any encoder (\cref{prop:lower}).

Exchangeability is a strong assumption, and a moment test of it rejects on 21 of the 23 full real features where it can run (\cref{sec:mi-pop}). We use it as an idealised benchmark that explains when and why the beta does well. The results that do not need it are the $n+1$ shift and the corrections built on it (\cref{sec:popularity}), the non-identification of the singletons' share (\cref{prop:nonid}) and the predictive lower bound on $I(X;Y)$ (Remark~\ref{rem:mi-lower}). For (b), the theory names the two places where the remaining gap can live, and the experiments find both. The first is the \emph{shape} of the prior on large levels, where higher moments matter; a nonparametric prior, the Kiefer--Wolfowitz NPMLE, closes it once levels are large enough. The second, which dominates on several real features, is dependence between a level's popularity and its target rate. The right prior for levels seen $n$ times is then the law of $q$ among levels seen $n+1$ times (\cref{prop:shift}), a Robbins-type shift \citep{robbins1956}. This is our main positive result for real data. Unseen levels should then be predicted by the label rate of the levels seen once (the \emph{unseen correction}) and levels seen once from the label pairs of the levels seen twice; together these give the \emph{Robbins correction} for $n\le1$, which needs only the singleton and doubleton counts and no fitted prior, and is asymptotically exact on those levels for any joint law of popularity and target (\cref{cor:robbins}).

For (c), the answer is negative in a precise sense. From levels seen at most $L$ times the data determine only $L$ moments of the prior, and $I(X;Y)$ is known only up to an interval whose worst-case half-width is the best polynomial approximation error of the binary entropy, about $0.22/L^2$ nats for even $L$ (\cref{prop:pi}); singletons alone say nothing about $I$ beyond the label rate (\cref{cor:singletons}). A well-calibrated encoder therefore does not imply a well-determined information estimate. The asymmetry runs deeper than moments: the prediction on levels seen at most once needs no exchangeability, while the information does. Without it, the conditionals of levels that are never seen twice are identified only through their mean, so only a trivial bound is identified for the singletons' share of $H(Y\mid X)$ (\cref{prop:nonid}).

\paragraph{What is and is not new.} The encoder is classical: the posterior mean of a beta-binomial empirical-Bayes model, the smoothed encoding of \citet{micci2001} with its pseudo-count chosen by marginal likelihood, and the estimator of \citet{hernandez2019}. Several ingredients are known: exactness at $n\le1$ is B\"uhlmann credibility (\cref{rem:credibility}); \cref{lem:moments} and the limit to $T$ moments from $T$ draws per unit are in \citet{tian2017}; the duality behind \cref{prop:pi} is that of \citet{wu2016}; the likelihood interval is an instance of the profile sets of \citet{chen2018}; and the risk of add-constant rules at large $n$ is classical \citep{braess2004}. For symmetric priors, the marginal-likelihood fit when levels are seen at most twice was already analysed by \citet{hernandez2019}; \cref{prop:ml} treats an arbitrary mean and proves consistency. What is new is the account of \emph{when} and \emph{why} this encoder is exact, what is left over on real data, and what cannot be learned; we do not propose an encoder that beats the state of the art.

\paragraph{Contributions.} (1) Under exchangeability, exactness of the moment-matched beta on levels seen at most once (twice for symmetric priors), consistency of its type-II ML fit when levels are seen at most twice, a second-order formula for the gap on larger levels and a lower bound beyond the largest level size, with $(F_1+2F_2)/N$ as a label-free estimate of the test mass this covers (\cref{sec:exact}). (2) Under popularity dependence, the $n+1$ shift with a bound on the reweighting between training and test strata, the unseen and Robbins corrections, and a frequency-aware prior with a likelihood-ratio fallback (\cref{sec:popularity}). (3) For $I(X;Y)$, partial identification with the polynomial-approximation bound, non-identification of the singletons' share without exchangeability, a likelihood-region interval with its asymptotic coverage, and a moment test of exchangeability (\cref{sec:mi}). (4) Evidence from synthetic priors, real-data subsampling oracles with known conditionals, a held-out bakeoff on 24 real features against the NPMLE, a GLMM encoder, CatBoost's target statistic and fixed smoothing, and coverage studies (\cref{sec:experiments,sec:mi-exp}); the absolute stakes are small, at most about $0.01$ nats per test row between fitted priors. (5) The open-source package \texttt{\pkgname{}}\ifpreprint\ (\pkgurl)\fi, which reproduces every figure and table.

\paragraph{Scope and organisation.} The theory is for a binary target; the multi-class case is discussed in \cref{sec:discussion}. \Cref{sec:related} reviews related work, \cref{sec:setup} sets up the model and the loss, \cref{sec:exact,sec:popularity,sec:mi} give the theory, \cref{sec:experiments,sec:mi-exp} the experiments, and \cref{sec:discussion} discusses a practical recipe and the limitations.

%% file: sections/02_related_work.tex
\section{Related work}\label{sec:related}

\paragraph{Target encoding.} \citet{micci2001} proposed replacing each level by a blend of its own target rate and the global rate, motivated as an empirical-Bayes estimate; the $m$-estimate $(k_x+m\hat\mu)/(n_x+m)$ is the special case of a beta prior with pseudo-count $m$. \citet{pargent2022} benchmarked encoders for high-cardinality features and found regularised target encoding best, with a GLMM (random-intercept logistic) encoder on top. \Citet{matteucci2023} benchmark more encoders on more data sets, recommend one-hot, sum, binary and weight-of-evidence encodings, note that this differs from the conclusion of \citet{pargent2022}, and find that rankings depend on the choice of data sets. CatBoost's ordered target statistics compute each row's encoding from the rows before it in a random permutation, to avoid target leakage in boosting \citep{prokhorenkova2018}. Conjugate Bayesian encoders have been proposed for production use \citep{slakey2019,larionov2020}; they fix or tune the prior rather than study its distance to the oracle. Libraries also offer a James--Stein encoder, a fitted-$m$ estimate under a normal rather than a beta-binomial model \citep{james1961,efron1973}, and a leave-one-out encoder, which is the device by which our cross-validated $m$ is tuned. Random effects can also be built into the downstream learner \citep{sigrist2022,simchoni2021}, and string categories can be encoded by their morphology \citep{cerda2018,cerda2020}; these replace the encoder or use side information, and we do not compare with them. Our question is not which encoder wins on average but how far any encoder can be from the Bayes oracle on a given feature.

\paragraph{Empirical Bayes and the NPMLE.} Encoding many levels at once is a compound estimation problem. Our baseline is parametric empirical Bayes with a beta prior \citep{efron1975}; on levels seen at most once its exactness is linear (B\"uhlmann) credibility, which is exact for conjugate priors \citep{buhlmann1967,jewell1974,diaconis1979}. \citet{robbins1956} wrote the Bayes rule for Poisson counts in terms of the frequencies of neighbouring counts; \cref{prop:shift} is a result of this kind for popularity, and its moment estimates are $f$-modelling \citep{efron2014}. Its $n=0$ case, predicting the unseen from the once-seen, has a precedent in the back-off of language models, which estimate the distribution of a word after an unseen context from the contexts seen once \citep{kneser1995,generet1995}, and in the observation of \citet{baayen1996} that hapax legomena are the best guide to unseen words. \citet{chen2026} studies empirical Bayes for normal means when the estimation precision predicts the parameter, the Gaussian analogue of our popularity dependence. The richer prior we use is the NPMLE of the mixing distribution \citep{kiefer1956,laird1978,koenker2014}, which approaches the oracle Bayes risk for normal means \citep{jiang2009}; for binomial units see \citet{vinayak2019,chenlei2025,han2025}, \citet{polyanskiy2021} give sharp regret bounds for Robbins' estimator, and \citet{shen2022} compare Robbins-type ($f$-modelling) and NPMLE-based ($g$-modelling) rules. We adopt the oracle-Bayes view of \citet{efron2019}.

\paragraph{Learning populations of parameters.} \citet{tian2017} show that from units each observed a fixed number $T$ of times, the population of binomial parameters is learnable only through its first $T$ moments. \Cref{prop:exact,prop:lower,prop:pi} are this phenomenon seen from the prediction and the information side, with a random, mostly tiny number of observations per level and a loss that weights levels by popularity, which creates the $n+1$ shift; a prior can also be exact for prediction while far from the truth in the Wasserstein distance they use (\cref{rem:w1}).

\paragraph{Entropy, information and partial identification.} No unbiased entropy estimator exists \citep{paninski2003}; minimax estimators rely on best polynomial approximation \citep{wu2016,jiao2015,wu2019}, and the duality between moment matching and best approximation that gives \cref{prop:pi} is that of \citet[Appendix~E]{wu2016}, with the $L^{-2}$ rate of \citet{timan1963}. \citet{valiant2017} estimate symmetric properties from the fingerprint by linear programming. Classical corrections \citep{miller1955,panzeri1996,chao2003} and Bayesian estimators \citep{nemenman2002,archer2013,archer2014} are compared with ours in Appendix~\ref{app:mi-point}. \Citet{hernandez2019} estimate $I(X;Y)$ for a large-alphabet $X$ and a small-alphabet $Y$ with the hierarchical beta prior studied here. Taking the range of a binomial empirical-Bayes functional over priors whose moments lie in a confidence set goes back to \citet{sivaganesan1993}, who use it for robust Bayesian credible sets and note that the data identify only finitely many moments. Our likelihood interval is an instance of the profile quasi-likelihood sets of \citet{chen2018}; \citet{ignatiadis2022} and \citet{gu2025} build confidence intervals for empirical-Bayes estimands that combine partial identification with sampling uncertainty, the latter for discrimination detection in the setting of \citet{kline2021}.

%% file: sections/03_setup.tex
\section{Setup}\label{sec:setup}

\paragraph{Levels, popularity and conditionals.} A categorical feature $X$ takes values (levels) in a large set $\mathcal X$, and the target is $Y\in\{0,1\}$. Level $x$ carries a pair $(p_x,q_x)$: its popularity $p_x=P(X=x)$ and its conditional $q_x=P(Y=1\mid X=x)$. A training sample of $N$ rows gives each level a count $n_x\sim\Bin(N,p_x)$ and, given $n_x$, a number of ones $k_x\sim\Bin(n_x,q_x)$. We write $F_n$ for the number of levels seen exactly $n$ times, $F_{\ge n}=\sum_{j\ge n}F_j$, and $F_{n,k}$ for those seen $n$ times with $k$ ones (the \emph{fingerprint}); $\nlev$ is the number of levels; $\mu=P(Y=1)$, and $h(q)=-q\log q-(1-q)\log(1-q)$ is the binary entropy in nats. We abbreviate $\kl(a\|b)=\KL(\Bern(a)\|\Bern(b))$.

\begin{assumption}[Exchangeable conditionals]\label{ass:exch}
The $q_x$ are i.i.d.\ draws from a prior $\pi$ on $[0,1]$, independent of the popularities $(p_x)$.
\end{assumption}

This is the structural assumption of \citet{hernandez2019}. Sections~\ref{sec:exact} and~\ref{sec:mi} use it; Section~\ref{sec:popularity} drops it and lets the pairs $(p_x,q_x)$ have any joint law.

\paragraph{Encoders and the loss.} An \emph{encoder} assigns to each level a predicted probability $r_x\in(0,1)$ of $Y=1$, computed from the training data; unseen levels ($n_x=0$) also get a prediction. We score it by the held-out log-loss, or equivalently by its \emph{excess cross-entropy}
\begin{equation}\label{eq:ece}
\eCE(r)=\E\Big[\sum_x p_x\,\kl(q_x\,\|\,r_x)\Big],
\end{equation}
the expected log-loss on a fresh test row minus that of the true conditional. The marginal predictor $r_x\equiv\mu$ has $\eCE=I(X;Y)$; the Bayes oracle defined below has the smallest $\eCE$ among encoders that treat levels exchangeably. We report results on the scale between the two, as the \emph{share of the gain lost},
\[
\frac{\eCE(r)-\eCE(\text{oracle})}{I(X;Y)-\eCE(\text{oracle})}.
\]

\paragraph{Moments.} For $a,b\in\mathbb N$ let $m_j=\E_\pi[q^j]$ and
\[
M_\pi(a,b)=\E_\pi\big[q^a(1-q)^b\big]=\sum_{i=0}^{b}\binom bi(-1)^i\,m_{a+i},
\]
a linear function of $m_0=1,m_1,\dots,m_{a+b}$. Write $v=m_2-m_1^2$. We assume throughout that $\pi$ is not a point mass and puts mass on $(0,1)$, which is equivalent to $0<v<\mu(1-\mu)$ and implies $M_\pi(a,b)>0$ for all $a,b$.

\paragraph{The baseline encoder.} For $\mu\in(0,1)$ and $s>0$ let $B_{\mu,s}=\BetaD(\mu s,(1-\mu)s)$, whose posterior predictive after $k$ ones in $n$ trials is $(k+\mu s)/(n+s)$: classical smoothed target encoding \citep{micci2001} with pseudo-count $s$. We write $m$ only for the pseudo-count of the $m$-estimate when it is fixed or tuned by hand (so the baseline is the $m$-estimate with $m=\hat s$), and $m_j$ for moments. The baseline sets $\mu=\hat\mu$, the empirical rate of $Y=1$, and $s=\hat s$, the maximiser of the beta-binomial marginal likelihood \citep{hernandez2019}. It is convenient to use
\[
t=\frac{1}{1+s}\in(0,1),
\]
the intra-level correlation of $Y$; under $B_{\mu,s}$, $v=\mu(1-\mu)\,t$.

\begin{definition}[Moment-matched beta]\label{def:matched}
The moment-matched beta of $\pi$ is $B^\pi=B_{\mu,s_\pi}$ with $s_\pi=\mu(1-\mu)/v-1$, equivalently $t_\pi=v/(\mu(1-\mu))$: the unique beta with the same $m_1,m_2$ as $\pi$.
\end{definition}

\paragraph{Per-level risk.} For a predictor $r(k,n)$ that depends on a level only through its counts, let
\[
R_n(r)=\E\big[\kl\big(q\,\|\,r(K,n)\big)\big],\qquad \Delta_n(r)=R_n(r)-R_n(p_\pi),
\]
with $q\sim\pi$ and $K\mid q\sim\Bin(n,q)$, where $p_\pi(k,n)=\E_\pi[q\mid K=k,n]$ is the Bayes (oracle) predictive. $\Delta_n$ is the regret of $r$ on a level seen $n$ times.

\begin{lemma}[Decomposition by level size]\label{lem:decomp}
Under Assumption~\ref{ass:exch}, for every encoder whose prediction for a level depends only on that level's $(k_x,n_x)$,
\[
\eCE(r)=\sum_{n\ge0}w_n\,R_n(r),\qquad
w_n=\E\Big[\sum_x p_x\binom Nn p_x^n(1-p_x)^{N-n}\Big]=\frac{n+1}{N+1}\,\E\big[F^{(N+1)}_{n+1}\big],
\]
where $F^{(N')}_j$ is the number of levels seen $j$ times in a sample of size $N'$. The weight $w_n$ is the probability that a test row lands in a level seen $n$ times in training, and $\sum_nw_n=1$.
\end{lemma}

The second form of $w_n$ is the Good--Turing identity \citep{good1953}: $w_n$ is estimated from the training fingerprint by $\hat w_n=(n+1)F_{n+1}/N$. In particular the test mass on levels seen at most once is estimated by
\begin{equation}\label{eq:diagnostic}
\hat w_0+\hat w_1=\frac{F_1+2F_2}{N},
\end{equation}
which needs no labels. Section~\ref{sec:exact} shows that on this mass a fitted beta is exactly as good as the oracle under Assumption~\ref{ass:exch}, and Section~\ref{sec:popularity} what changes without it. \Cref{sec:oracle} checks it on real-data pools.

%% file: sections/04_moment_exactness.tex
\section{Moment exactness}\label{sec:exact}

This section works under Assumption~\ref{ass:exch}. Proofs are in Appendix~\ref{app:proofs}.

\subsection{Two lemmas}

\begin{lemma}[Finitely many moments]\label{lem:moments}
Fix $n$. Then
\[
P_\pi(K=k\mid n)=\binom nk M_\pi(k,n-k),\qquad
p_\pi(k,n)=\frac{M_\pi(k+1,n-k)}{M_\pi(k,n-k)}.
\]
Hence the law of $K$ given $n$ depends on $\pi$ only through $m_1,\dots,m_n$, and the oracle predictive $p_\pi(\cdot,n)$ only through $m_1,\dots,m_{n+1}$.
\end{lemma}

The lemma is classical \citep{skibinsky1967,lord1969}; \citet[Section~2]{tian2017} state it in this form.

\begin{lemma}[Regret is a KL between predictives]\label{lem:regret}
For every predictor $r$,
\[
\Delta_n(r)=\sum_{k=0}^{n}P_\pi(k\mid n)\;\kl\big(p_\pi(k,n)\,\|\,r(k,n)\big)\ \ge 0,
\]
and the oracle risk is a conditional mutual information, $R_n(p_\pi)=I(q;Y_{\rm new}\mid K)$. In particular $R_0(p_\pi)=I(q;Y)$, which is $I(X;Y)$ when no level carries a non-negligible share of the mass.
\end{lemma}

Together with Lemma~\ref{lem:decomp}, these reduce the comparison with the oracle to a comparison of predictives, one level size at a time, and each predictive to finitely many moments of $\pi$.

\subsection{Exactness on sparse levels}

\begin{proposition}[Two moments suffice up to $n=1$]\label{prop:exact}
For every prior $\pi$, the moment-matched beta satisfies $p_{B^\pi}(k,n)=p_\pi(k,n)$, and hence $\Delta_n(p_{B^\pi})=0$, for $n\in\{0,1\}$ and all $k\le n$. Moreover $P_{B^\pi}(k\mid n)=P_\pi(k\mid n)$ for $n\le2$.
\end{proposition}

\begin{remark}[Credibility]\label{rem:credibility}
At $n\le1$ the count $K$ takes at most two values, so every predictor is affine in $K$ and the Bayes rule coincides with the best linear (B\"uhlmann credibility) predictor, which depends only on $m_1,m_2$ \citep{buhlmann1967}. That the beta prior's posterior mean is exactly this linear rule for all $n$ is the exact-credibility property of conjugate priors \citep{jewell1974,diaconis1979}. Proposition~\ref{prop:exact} is therefore elementary; what follows (the $n=2$ condition, the type-II ML fit, the size of the remaining gap and the lower bound) is what decides how much of a real feature this covers.
\end{remark}

At $n=2$ the predictive needs $m_3$. The beta's third central moment is
\begin{equation}\label{eq:k3}
\kappa_3(B_{\mu,s})=\frac{2\,v\,(1-2\mu)}{s+2}.
\end{equation}

\begin{proposition}[The $n=2$ case]\label{prop:n2}
Let $\delta_3=m_3(\pi)-m_3(B^\pi)=\kappa_3(\pi)-\kappa_3(B^\pi)$. Then $\Delta_2(p_{B^\pi})=0$ if and only if $\delta_3=0$. Each predictive at $n=2$ is off by $\pm\delta_3$ divided by a quantity that depends only on $m_1,m_2$, and $\Delta_2(p_{B^\pi})$ is of order $\delta_3^2$ (explicit forms in Appendix~\ref{app:proofs-exact}).
\end{proposition}

\begin{corollary}[Symmetric priors]\label{cor:sym}
If $\pi$ is symmetric about $1/2$, then $\kappa_3(\pi)=0=\kappa_3(B^\pi)$ and the moment-matched beta is exact for all $n\le2$. This covers the junk-fraction priors $f\,\delta_{1/2}+\frac{1-f}{2}(\delta_{q_0}+\delta_{1-q_0})$ of \cref{sec:synthetic}.
\end{corollary}

Exact symmetry arises in label-randomised designs, such as a match-up feature whose orientation is assigned at random. Symmetry does not extend exactness beyond $n=2$, and on real data the relevant quantity is $\delta_3$, which enters the regret squared (\cref{prop:n2}).

\begin{corollary}[Where the gap can live]\label{cor:where}
Under Assumption~\ref{ass:exch},
\[
\eCE(p_{B^\pi})-\eCE(p_\pi)=\sum_{n\ge2}w_n\,\Delta_n(p_{B^\pi})
\qquad\Big(=\textstyle\sum_{n\ge3}w_n\Delta_n \text{ if } \delta_3=0\Big).
\]
The test mass on which the moment-matched beta is exactly the oracle is $w_0+w_1$, estimated by $(F_1+2F_2)/N$ (add $3F_3/N$ when $\delta_3=0$).
\end{corollary}

\subsection{The type-II ML fit recovers the matched beta}

\begin{proposition}[Consistency in the all-sparse regime]\label{prop:ml}
Suppose every level has $n_x\le2$, let $F_2=\#\{x:n_x=2\}\to\infty$ and $\hat\mu\to m_1$ in probability. Fix $\hat\mu$ in the beta-binomial marginal likelihood and let $\hat t$ maximise it over $t\in[0,1]$. Then:
\begin{enumerate}
\item given $\hat\mu$, levels with $n_x\le1$ do not affect $\hat t$;
\item the log-likelihood is $\ell(t)=\sum_{x:n_x=2}\log P_t(k_x)$ with
\[
P_t(2)=\hat\mu^2+\hat\mu(1-\hat\mu)t,\quad P_t(1)=2\hat\mu(1-\hat\mu)(1-t),\quad P_t(0)=(1-\hat\mu)^2+\hat\mu(1-\hat\mu)t,
\]
affine in $t$, so $\ell$ is strictly concave with a unique maximiser on $[0,1]$;
\item $\hat t\to t_\pi$ in probability; equivalently $\hat s\to s_\pi$.
\end{enumerate}
Consequently the fitted encoder converges to $p_{B^\pi}$, and its regret on levels with $n\le1$ ($n\le2$ for symmetric $\pi$) tends to zero.
\end{proposition}

\begin{remark}[Levels with $n_x\ge3$]\label{rem:mixed}
With larger levels present the likelihood is misspecified, and $\hat t$ converges to a pseudo-true value that need not match $m_2$. In the equal-popularity cases of Table~\ref{tab:avg} the pseudo-true fit is no worse than the moment-matched beta in total: it trades a loss of at most $5\times10^{-5}$ nats on levels seen at most once for a gain on denser ones. Heavy-tailed level sizes let large levels dominate: under lognormal popularity the loss on levels seen at most once grows to $4\times10^{-4}$ nats, still under $0.1\%$ of $I$, and in an extreme design with $10\%$ of the levels seen 200 times the per-level regret at $n=1$ reaches $7\%$ of $I$ (\cref{tab:heavy}). Fitting $t$ on the levels with $n_x\le2$ alone restores \cref{prop:ml} for the predictions at $n\le1$.
\end{remark}

\subsection{The gap for $n\ge2$}

Let $\tilde\pi$ be any prior (in practice the fitted beta) and $\delta_j=m_j(\pi)-m_j(\tilde\pi)$.

\begin{proposition}[Second-order expansion]\label{prop:expansion}
As $\delta\to0$, $\Delta_n(p_{\tilde\pi})$ is, up to $O(\|\delta\|^3)$, a positive semidefinite quadratic form in $(\delta_1,\dots,\delta_{n+1})$, given explicitly in Appendix~\ref{app:proofs-exact}. For the moment-matched beta, $\delta_1=\delta_2=0$ and the form involves only $\delta_3,\dots,\delta_{n+1}$.
\end{proposition}

The expansion is informative only while the mismatches are small; at large $n$ the regret is governed by the shape of $\pi$ near $0$ and $1$, and for atomic priors it decays like $1/(2n)$ while the oracle's risk decays exponentially, which is where the NPMLE pays off (\cref{rem:largen}, Appendix~\ref{app:exact-more}).

\subsection{No encoder can do uniformly better beyond the data}

\begin{proposition}[Non-identifiability bound]\label{prop:lower}
Suppose every level has $n_x\le n_{\max}$. Let $\pi_1,\pi_2$ share $m_1,\dots,m_{n_{\max}}$, a vector interior to the moment space (so that such pairs exist), and differ in $m_{n_{\max}+1}$. Then the training data have the same law under $\pi_1$ and $\pi_2$, and every data-driven encoder $\hat r$ satisfies
\[
\max_{i=1,2}\ \E\big[\Delta^{\pi_i}_{n_{\max}}(\hat r)\big]\ \ge\ \frac12\sum_{k}P(k\mid n_{\max})\,\min_{r\in(0,1)}\Big\{\kl\big(p_{\pi_1}(k)\,\|\,r\big)+\kl\big(p_{\pi_2}(k)\,\|\,r\big)\Big\}>0 .
\]
\end{proposition}

\Citet{sivaganesan1993} compute the range of binomial empirical-Bayes posterior means over all priors with given moments, the quantity this bound controls from below.

This is the finite-sample form of the identifiability limit for binomial mixtures \citep{teicher1963,tian2017}: moments beyond the largest level size are invisible, so the oracle's advantage on the largest levels cannot be learned. In the all-sparse regime ($n_{\max}=2$) the beta encoder already uses everything that is identifiable, and is exact wherever the identifiable moments determine the oracle.

\paragraph{Numerical check.} Exact sums over binomial outcomes confirm that the regret vanishes at $n\le1$ to machine precision and that, with equally popular levels, the type-II ML fit with larger levels present stays within $5\times10^{-5}$ nats of the oracle there; on the true distributions of $q$ of the six real-data pools of \cref{sec:oracle} the moment-matched beta's regret at $n=2$ is under $1\%$ of $I(X;Y)$ (Appendix~\ref{app:exact-more}).

%% file: sections/05_popularity.tex
\section{Popularity and the $n+1$ shift}\label{sec:popularity}

On real data the remaining gap of the baseline is mostly not a matter of higher moments: it comes from dependence between a level's popularity and its target rate (\cref{sec:oracle}). This section drops Assumption~\ref{ass:exch} and lets the pairs $(p_x,q_x)$ have any joint law. Then a level's count $n_x$ is informative about $q_x$, and the question is which prior a level of a given size should use.

\subsection{Stratification and the shift identity}

\begin{lemma}[Stratification]\label{lem:strat}
Let $w_n=\E\sum_xp_x\Bin(n\mid N,p_x)$ and define the \emph{test-weighted prior} of stratum $n$,
\[
\pi^*_n(dq)=\frac1{w_n}\,\E\sum_x p_x\,\Bin(n\mid N,p_x)\,\mathbf 1\{q_x\in dq\}.
\]
Then $\eCE(r)=\sum_nw_nR^{(n)}(r)$, where $R^{(n)}(r)=\E\,\kl(q\|r(K,n))$ with $q\sim\pi^*_n$ and $K\mid q\sim\Bin(n,q)$.
\end{lemma}

Equivalently, if $X^*$ is the level of a test row, $\pi^*_n$ is the law of $q_{X^*}$ given $n_{X^*}=n$: the prior of the levels the loss scores at stratum $n$, with each level weighted by its chance $p_x$ of receiving the test row. Each stratum is therefore the problem of Section~\ref{sec:exact} with its own prior $\pi^*_n$, and Lemmas~\ref{lem:moments}--\ref{lem:regret} and Propositions~\ref{prop:exact}, \ref{prop:n2}, \ref{prop:expansion} and~\ref{prop:lower} apply inside each stratum with $\pi\to\pi^*_n$. The oracle for a level seen $n$ times is the posterior mean under $\pi^*_n$ and uses its moments up to order $n+1$. Under Assumption~\ref{ass:exch}, $\pi^*_n=\pi$ for every $n$. The Good--Turing form of $w_n$ in Lemma~\ref{lem:decomp} never used independence, so it survives unchanged.

The key fact is that $\pi^*_n$ is revealed by the training data, one stratum up.

\begin{proposition}[Shift identity]\label{prop:shift}
$\pi^*_n$ is the level-weighted law of $q$ among levels seen $n+1$ times in a sample of $N+1$ rows:
\[
\pi^*_n(dq)=\frac{\E\sum_x\Bin(n+1\mid N+1,p_x)\,\mathbf 1\{q_x\in dq\}}{\E\sum_x\Bin(n+1\mid N+1,p_x)}=:\pi^{\lev}_{n+1}(dq).
\]
Equivalently, the oracle has the Robbins form
\begin{equation}\label{eq:robbins}
p^*(k,n)=\frac{(k+1)\,\E F^{(N+1)}_{n+1,k+1}}{(k+1)\,\E F^{(N+1)}_{n+1,k+1}+(n+1-k)\,\E F^{(N+1)}_{n+1,k}} .
\end{equation}
\end{proposition}

Here $\pi^{\lev}_{n+1}$ is the law of $q$ for a level drawn uniformly at random from those seen $n+1$ times, so each level gets weight $1$ instead of $p_x$; it is what the table $F_{n+1,k}$ reveals, as a mixture of $\Bin(n+1,q)$. Adding the test row to a level seen $n$ times makes it a level seen $n+1$ times, and the test row is one of its $n+1$ exchangeable rows. The loss weights levels by $p_x$, and $p\,\Bin(n\mid N,p)\propto\Bin(n+1\mid N+1,p)$, so the right prior for a level seen $n$ times is the population of levels seen $n+1$ times. Training levels are seen out of $N$ rows rather than $N+1$. Exactly, $\Bin(n+1\mid N+1,p)/\Bin(n+1\mid N,p)=(N+1)(1-p)/(N-n)$, so the identity reweights each level by a factor proportional to $1-p_x$. The next lemma shows that this reweighting is negligible.

\begin{lemma}[Training strata]\label{lem:train-shift}
Let $\nu_{n+1}$ be the level-weighted law of $(p,q)$ among levels seen $n+1$ times in the $N$ training rows, and $\varepsilon_n=\E_{\nu_{n+1}}p$. For every $g$ with values in $[0,1]$,
\[
\big|\E_{\pi^*_n}g(q)-\E_{\nu_{n+1}}g(q)\big|\le\frac{\varepsilon_n}{1-\varepsilon_n},\qquad
\varepsilon_n=\frac{(n+2)\,\E F^{(N+1)}_{n+2}}{(N+1)\,\E F_{n+1}}\le\frac1{\E F_{n+1}}.
\]
\end{lemma}

So every moment $\E q^j$ of the training stratum differs from that of $\pi^*_n$ by $O(1/F_{n+1})$, an order below its sampling error $O(F_{n+1}^{-1/2})$, whatever the joint law of $(p_x,q_x)$. A level seen $\ell$ times with $k$ ones gives the unbiased estimate $\binom kj/\binom{\ell}j$ of $q^j$ for $j\le \ell$, so
\begin{equation}\label{eq:momest}
\hat m^*_{n,j}=\frac1{F_{n+1}}\sum_kF_{n+1,k}\binom kj\Big/\binom{n+1}j,\qquad j\le n+1,
\end{equation}
are exactly the moments the oracle at stratum $n$ needs \citep[this is $f$-modelling in the sense of][]{efron2014}.

\subsection{Levels seen at most twice}

Asymptotic statements in this subsection are along a sequence of populations in which the numbers of singletons and doubletons grow, $F_1,F_2\to\infty$, with no level carrying a non-vanishing share of either (for a fixed law and $N\to\infty$, $w_0$ and $w_1$ vanish instead). Regrets are per stratum: on levels seen $n$ times, divided by $w_n$.

\paragraph{Unseen levels ($n=0$).} The oracle for an unseen level needs one number, the mean $m^*_{0,1}$ of $\pi^*_0=\pi^{\lev}_1$.

\begin{corollary}[Unseen correction]\label{cor:unseen}
The oracle predicts unseen levels by the mean of $q$ among levels seen once, estimated by the singletons' label rate $\hat m^*_{0,1}=(F_{1,1}+\hat\mu)/(F_1+1)$, shrunk toward $\hat\mu$ by one pseudo-observation so that it is never $0$ or $1$ (unsmoothed, $\E\,\kl$ below would be infinite). A constant prediction $c$ for unseen levels has regret $w_0\,\kl(m^*_{0,1}\|c)$. The blind baseline uses $c=\hat\mu$, an estimate of $\sum_nw_nm^*_{n,1}$, and has regret $w_0\,\kl(m^*_{0,1}\|\mu)+O(1/N)$. The singletons' rate has regret
\[
w_0\,\E\,\kl\big(m^*_{0,1}\,\|\,\hat m^*_{0,1}\big)\le\frac{w_0}{2F_1}\big(1+o(1)\big)\approx\frac1{2N}
\]
per test row, using $w_0\approx F_1/N$. The correction therefore helps whenever $\kl(m^*_{0,1}\|\mu)\gtrsim1/(2F_1)$, and costs at most about $1/(2N)$ nats per test row when $q$ does not depend on popularity.
\end{corollary}

The baseline's $\mu$ is the right average of the stratum means but generally the wrong value for this stratum, which carries the most test weight in sparse data. A prior fitted per level size and evaluated at the level's own size $n=0$ is also wrong: it predicts the mean of the unseen levels, which are mostly the least popular ones, while a test row that lands in an unseen level comes from the more popular end of that set.

\paragraph{Levels seen once ($n=1$).} The oracle needs the mean $\mu^*_1$ and correlation $t^*_1=v^*_1/(\mu^*_1(1-\mu^*_1))$ of $\pi^*_1$:
\[
p^*(1,1)=\mu^*_1+(1-\mu^*_1)t^*_1,\qquad p^*(0,1)=\mu^*_1(1-t^*_1).
\]

\begin{proposition}\label{prop:pop-n1}
A beta encoder is exact on levels seen once if and only if the beta it uses for them has $\mu=\mu^*_1$ and $t=t^*_1$, the mean and correlation of $q$ among \emph{doubletons}. In terms of the doubleton table,
\[
\hat p^*(1,1)=\frac{2F_{2,2}}{2F_{2,2}+F_{2,1}},\qquad\hat p^*(0,1)=\frac{F_{2,1}}{F_{2,1}+2F_{2,0}}.
\]
\end{proposition}

The unseen correction and these doubleton rates together need no prior at all.

\begin{corollary}[Robbins correction for $n\le1$]\label{cor:robbins}
Let $\hat r$ predict unseen levels by the singletons' rate $\hat m^*_{0,1}$ of \cref{cor:unseen}, levels seen once by the doubleton rates $\hat p^*(\cdot,1)$ of Proposition~\ref{prop:pop-n1}, each shrunk toward the baseline's prediction by one pseudo-pair, and every other level as the baseline does. Then, whatever the joint law of $(p_x,q_x)$, the per-stratum regret of $\hat r$ on levels seen at most once tends to zero as $F_1,F_2\to\infty$, up to the shift from $N+1$ to $N$ rows bounded by \cref{lem:train-shift}. When $q$ does not depend on popularity, the doubleton rates are noisy estimates of what the fitted beta already predicts, and cost $O(w_1/F_2)=O(1/N)$ nats per test row.
\end{corollary}

The cost matters only when the number of doubletons is small. Our implementation (\texttt{method="baseline\_robbins"}) therefore uses the doubleton rates only when a likelihood-ratio test rejects the fitted beta for the doubletons' label pairs (two degrees of freedom, level $5\%$), and otherwise keeps the baseline on once-seen levels. For $k_y>2$ classes the same construction predicts, for a level seen once with label $c$, the label distribution of the other row of the doubletons that contain a $c$.

\paragraph{Levels seen twice and symmetric conditionals.} At $n=2$ the oracle needs three moments of $\pi^*_2$, the law among tripletons, and a beta is exact there under the $\kappa_3$ condition of Proposition~\ref{prop:n2} applied to $\pi^*_2$. If every $\pi^*_n$ is symmetric about $1/2$, each stratum is described by one number, and the blind baseline stays exact on unseen and once-seen levels when every level is seen at most twice, despite the popularity dependence (Appendix~\ref{app:sym}, with a summary of the three cases in \cref{tab:pop-n2}).

\subsection{A frequency-aware prior}

The shift identity calls for a prior that varies with level size, evaluated one size up. We use a beta-binomial regression,
\[
q_x\sim\BetaD\big(\mu_x s_x,(1-\mu_x)s_x\big),\qquad \operatorname{logit}\mu_x=a+b\,z_x,\qquad \log s_x=c\ \ (\text{or } c+d\,z_x),\qquad z_x=\log(1+n_x),
\]
with all parameters fitted by type-II ML on the training levels. With $b=d=0$ it reduces to the baseline with a free mean. Two choices follow from the theory:
\begin{itemize}
\item \textbf{Encoding at $n+1$.} The encoder evaluates the curve at $z=\log(2+n_x)$ (the $n+1$ shift). Unseen levels get $\mu$ at $\log 2$, which tracks the singletons' label rate.
\item \textbf{Likelihood-ratio fallback.} The regression is used only when the LR test of $b=d=0$ rejects at level $5\%$; otherwise the encoder is the baseline.
\end{itemize}
\begin{remark}[All-sparse consistency]\label{rem:freq-consistent}
If every level is seen at most twice, the shifted frequency-aware prior with $d=0$ is asymptotically exact on unseen and once-seen levels as $F_1,F_2\to\infty$: $(a,b,c)$ saturate the singletons' mean and the doubletons' mean and correlation, which are exactly what the $n=0$ and $n=1$ predictions need (Proposition~\ref{prop:pop-n1}; proof in Appendix~\ref{app:proofs}).
\end{remark}
The encoder is \texttt{\pkgname{}.fit\_freq\_prior} (method \texttt{"freq\_prior"}).

On a synthetic popularity-linked prior with $2\times10^5$ levels, reading the curve at $\log(1+n)$ instead of $\log(2+n)$ makes it worse than the marginal predictor at $N=20{,}000$ (eCE $0.155$ against $0.123$ nats), while the shifted reading beats both and the baseline ($0.097$ against $0.106$; \cref{fig:shift}, Appendix~\ref{app:figs}).

If a popularity covariate is known from outside the training counts (e.g.\ a full catalogue), then given it $n_x$ is independent of $q_x$ again and Section~\ref{sec:exact} applies within each popularity stratum with no shift; the shift arises only because $n_x$ is a noisy proxy for $p_x$.

%% file: sections/06_mutual_information.tex
\section{Mutual information: partial identification}\label{sec:mi}

The same moment structure that makes prediction easy on sparse levels makes the mutual information hard. This section works under Assumption~\ref{ass:exch}, except for \cref{sec:mi-pop}.

\subsection{The estimand}

\begin{lemma}[Count-weighted representation]\label{lem:rep}
$I(X;Y)=h(\mu)-H(Y\mid X)$ with
\[
H(Y\mid X)=\sum_xp_x\,h(q_x)=\E\Big[\frac1N\sum_xn_x\,h(q_x)\Big]=\sum_{n\ge1}\frac{n\,\E F_n}{N}\,\E_{\pi^{\lev}_n}[h(q)],
\]
where $\pi^{\lev}_n$ is the level-weighted law of $q$ among levels seen $n$ times, as in Proposition~\ref{prop:shift}. Under Assumption~\ref{ass:exch}, $\pi^{\lev}_n=\pi$ for every $n$ and $H(Y\mid X)=\E_\pi h(q)$.
\end{lemma}

Under Assumption~\ref{ass:exch} we take the estimand to be the superpopulation value $I=h(m_1)-\E_\pi h(q)$, the limit of $I(X;Y)$ when no level carries a non-negligible share; \cref{prop:pi,prop:lik} are about it.

Two consequences frame the rest of the section. First, unseen levels are not the obstacle: the weight $p_x$ that $H(Y\mid X)$ puts on a level is the expected share $n_x/N$ of the sample it receives, so unseen levels drop out of the estimand. This is why the decomposition $I=h(\mu)-H(Y\mid X)$ of \citet{hernandez2019} escapes the support-size problem that makes $H(X)$ hard (\cref{sec:mi-vv}). Second, the obstacle is $h$ on each seen level. $\E[g(K)\mid q]$ is a polynomial of degree $n$ in $q$ for every $g$, and $h$ is not a polynomial, so no unbiased per-level estimate of $h(q_x)$ exists \citep{paninski2003}. Pooling over levels turns the problem into one about $\pi$.

\subsection{What the data identify}

By Lemma~\ref{lem:moments} the law of the counts of a level of size $n$ depends on $\pi$ only through $m_1,\dots,m_n$, and $\binom kj/\binom nj$ is an unbiased estimate of $m_j$ for every $j\le n$. So the training data depend on $\pi$ only through $m_1,\dots,m_{n_{\max}}$, and $m_j$ is estimated from the $F_{\ge j}$ levels with at least $j$ observations, with standard error of order $F_{\ge j}^{-1/2}$. Define the identified set
\[
\mathcal I_L(m_1,\dots,m_L)=\big\{h(m_1)-\E_{\tilde\pi}h(q):\ m_j(\tilde\pi)=m_j,\ j\le L\big\}=[I_{\rm lo},I_{\rm hi}],
\]
an interval because the constraint set is convex and $\tilde\pi\mapsto\E_{\tilde\pi}h$ is linear.

\begin{proposition}[Partial identification]\label{prop:pi}
Suppose every level has $n_x\le L$. For every estimator $\hat I$ built from the data and every moment vector $m$,
\[
\sup_{\tilde\pi:\ m_{1:L}(\tilde\pi)=m}\E_{\tilde\pi}\big|\hat I-I(\tilde\pi)\big|\ \ge\ \tfrac12\,(I_{\rm hi}-I_{\rm lo}),
\]
however many levels there are. Over all priors,
\[
\sup_m\ \tfrac12\big(I_{\rm hi}(m)-I_{\rm lo}(m)\big)=E_L(h):=\min_{g\in\mathcal P_L}\ \max_{q\in[0,1]}|h(q)-g(q)|,
\]
where $\mathcal P_L$ is the set of polynomials of degree at most $L$: $E_L(h)$ is the error of the best uniform approximation of $h$ on $[0,1]$ by such a polynomial.
\end{proposition}

\begin{corollary}[Singletons carry no information about $I$]\label{cor:singletons}
With $L=1$, $\mathcal I_1=[0,h(\mu)]$: a point mass at $\mu$ gives $I=0$, and the two-point law on $\{0,1\}$ gives $I=h(\mu)$. All information about $I(X;Y)$ beyond the label rate comes from coincidences, i.e.\ levels seen at least twice.
\end{corollary}

\Citet[Section~4]{hernandez2019} observed that once-seen levels leave the posterior over the beta's concentration unchanged; that coincidences carry the information is the classical idea behind coincidence-based entropy estimation \citep{ma1981,nemenman2011}.

The second statement is the duality between moment matching and best polynomial approximation of \citet[Appendix~E]{wu2016}, applied to $\E_\pi h$ under the binomial kernel. Because $h$ is symmetric about $1/2$, $E_{2j+1}=E_{2j}$. Numerically (\cref{tab:EL} in Appendix~\ref{sec:mi-vv}), $E_1=\tfrac12\log2$ and, for even $L$, $L^2E_L$ rises from $0.15$ at $L=2$ to $0.218$ at $L=8$ and $0.226$ at $L=32$, towards a limit of about $0.227$ (by extrapolation; the $L^{-2}$ rate for $x\log x$ on $[0,1]$ is classical, \citealp{timan1963}); so $E_L\approx0.22/L^2$ for even $L$ of practical size. This is an identification floor, with $L$ capped by the data. It is not comparable to the minimax rates of \citet{wu2016} and \citet{jiao2015} for entropy, where the degree $L\asymp\log n$ is the statistician's choice and the error also includes sampling variance; the two results are consistent.

The worst case is pessimistic; for a given $\pi$ the interval can be much narrower. Table~\ref{tab:id} gives it for several priors, computed by linear programming over priors on a grid of 2001 atoms (\texttt{\pkgname{}.identified\_interval}). With doubletons only, $I$ is known to within $\pm0.01$ to $\pm0.035$ nats, roughly $\pm10$--$30\%$ of its value. Each extra pair of moments typically shrinks the width by a factor of 3 to 4, and discrete priors with $r$ atoms are identified exactly once $L\ge2r$: an $r$-atom measure with atoms in $(0,1)$ has a moment vector on the boundary of the moment space, and so a unique representing measure, only from $L=2r$ on \citep{karlin1966}. \Cref{fig:interval} in Appendix~\ref{app:figs} shows the shrinkage for three of these priors.

\begin{table}[t]
\centering\small
\begin{tabular}{lcccccc}
\toprule
Prior $\pi$ & true $I$ & beta $I$ & $L=2$ & $L=3$ & $L=4$ & $L=6$\\
\midrule
junk $f=0.5$, $q_0=0.1$ & 0.1840 & 0.1845 & [0.170, 0.222] & same & [0.184, 0.197] & 0.184\\
junk $f=0.8$, $q_0=0.1$ & 0.0736 & 0.0681 & [0.065, 0.089] & same & [0.074, 0.080] & 0.074\\
$\{.2,.3,.85\}$; $\{.3,.4,.3\}$ & 0.1634 & 0.1755 & [0.161, 0.213] & [0.162, 0.207] & [0.163, 0.166] & 0.163\\
$\BetaD(0.5,0.5)$ & 0.3069 & 0.3069 & [0.277, 0.347] & same & [0.298, 0.318] & [0.303, 0.312]\\
$\BetaD(1,3)$ & 0.1040 & 0.1040 & [0.085, 0.142] & [0.095, 0.133] & [0.100, 0.113] & [0.102, 0.107]\\
\bottomrule
\end{tabular}
\caption{Identified interval $[I_{\rm lo},I_{\rm hi}]$ (nats) given the true first $L$ moments, with infinitely many levels. ``Beta $I$'' is the value implied by the moment-matched beta, the limit of the beta estimator when all levels have $n\le2$. ``Same'': for these priors, symmetric about $1/2$, the extremal priors at $L=2$ can be symmetrised because $h$ is symmetric, so adding $m_3$ does not narrow the interval.}\label{tab:id}
\end{table}

\subsection{Estimators}

\begin{itemize}
\item \textbf{Minimax.} With known $m_{1:L}$, the midpoint of $[I_{\rm lo},I_{\rm hi}]$ minimises the worst-case error, which then equals the half-width (Proposition~\ref{prop:pi}); a polynomial-approximation estimator in the style of \citet{wu2016} and \citet{jiao2015} has bias at most $E_L(h)$ for every $\pi$.
\item \textbf{Moment interval.} Minimise and maximise $\E_{\tilde\pi}h$ over priors whose moments lie in a joint $95\%$ confidence region of the U-statistics $\hat m_{1:L}$ \citep[in the style of][]{valiant2017,sivaganesan1993}; simple, but its width is dominated by the noise in $\hat m_3,\hat m_4$, and $L$ must be chosen (\texttt{\pkgname{}.mi\_interval}; Appendix~\ref{app:mi-point}).
\item \textbf{Likelihood interval.} The moments are a lossy summary of the table $F_{n,k}$: the U-statistics weight every level alike, and $L$ must be chosen. Under Assumption~\ref{ass:exch} the table has the log-likelihood $\ell(\tilde\pi)=\sum_{n,k}F_{n,k}\log P_{\tilde\pi}(k\mid n)$ with $P_{\tilde\pi}(k\mid n)=\int\binom nk q^k(1-q)^{n-k}\,d\tilde\pi(q)$ linear in $\tilde\pi$, so $\ell$ is concave and its maximiser is the NPMLE. The likelihood interval is the range of $I$ over the convex set $\{\tilde\pi:\ell(\tilde\pi)\ge\ell(\hat\pi_{\rm NPMLE})-c\}$; for fixed $m_1$, $I$ is linear in $\tilde\pi$, so both ends are convex programmes (\texttt{mi\_interval(region="likelihood")}). Every cell then counts with its own binomial precision, the many small levels most, and levels of every size contribute, so no $L$ is chosen. At the population table the set reduces to the identified set, so only the sampling part of the width can shrink. Confidence sets for the marginal turned into intervals for an empirical-Bayes functional are the F-localization of \citet{ignatiadis2022}. We use the profile-likelihood cutoff $c=\chi^2_{1,0.95}/2$, which \cref{prop:lik} below justifies asymptotically.
\item \textbf{Bayes.} The estimator of \citet{hernandez2019} is the posterior-mean rule for the beta family. The family fixes all moments beyond the second, so its posterior spread reflects uncertainty in $s$ only and ignores the identification width; for non-beta priors its credible interval under-covers once that width exceeds the posterior standard deviation. A richer family such as the NPMLE moves the selected point but cannot shrink the identified set; when every level has $n_x\le L$ the NPMLE is not even unique, and mirror ascent from the uniform prior converges to the maximum-entropy prior with the fitted moments (\cref{prop:npmle-maxent}, Appendix~\ref{app:npmle}); our EM returns the same point to within $2\times10^{-4}$ nats.
\end{itemize}

Under exchangeability the likelihood interval is the one we recommend, and its large-sample behaviour can be stated exactly. Write $\mathcal M_L=\{m_{1:L}(\tilde\pi)\}\subset\mathbb R^L$ for the moment space of priors on $[0,1]$, $\ell(m)$ for the log-likelihood of the table as a function of the moments (\cref{prop:lik}(i) shows it is one), and $F=\sum_{n\ge1}F_n$ for the number of seen levels.

\begin{proposition}[The likelihood interval]\label{prop:lik}
Let Assumption~\ref{ass:exch} hold with $\pi((0,1))>0$, and condition on the level sizes, all at most $L$ (after thinning). Let $F\to\infty$ with the share of seen levels of size $n$ converging to $\omega_n$, $\omega_L>0$; let $m=m_{1:L}(\pi)$, and let $\hat C_c$ be the likelihood interval with cutoff $c$.
\begin{enumerate}
\item[(i)] $\ell(\tilde\pi)$ depends on $\tilde\pi$ only through $m_{1:L}(\tilde\pi)$. Hence $\hat C_c=\bigcup_{m'\in\hat M_c}\mathcal I_L(m')$, where $\hat M_c=\{m'\in\mathcal M_L:\ell(m')\ge\max\ell-c\}$ is the likelihood-ratio confidence set for the moments, and $\hat C_c\supseteq\mathcal I_L(m)$ whenever $m\in\hat M_c$.
\item[(ii)] For every $c$, $\liminf P\big(\mathcal I_L(m)\subseteq\hat C_c\big)\ge P(\chi^2_L\le 2c)$. With $c=\chi^2_{L,1-\alpha}/2$ the interval covers the whole identified set, and so $I(X;Y)$, with asymptotic probability at least $1-\alpha$, whether or not $m$ is on the boundary of $\mathcal M_L$.
\item[(iii)] For every $c=o(F)$ and $\epsilon>0$, $P\big(\hat C_c\subseteq[I_{\rm lo}(m)-\epsilon,I_{\rm hi}(m)+\epsilon]\big)\to1$: the interval shrinks onto the identified set. If $m$ is interior to $\mathcal M_L$ and $c$ is fixed, the excess width is $O_p(F^{-1/2})$.
\item[(iv)] Let $m$ be interior to $\mathcal M_L$ and $c=\chi^2_{1,1-\alpha}/2$. If $I_{\rm lo}(m)<I(X;Y)<I_{\rm hi}(m)$, the coverage tends to $1$. If $I(X;Y)$ is an end of a non-degenerate identified set, $I_{\rm hi}(m)$ or $I_{\rm lo}(m)$, and that end is differentiable at $m$ with a non-zero gradient, the coverage tends to $1-\alpha/2$. $I_{\rm hi}$ is concave on $\mathcal M_L$ and $I_{\rm lo}$ is a difference of convex functions, so both are differentiable at almost every $m$; at a kink the coverage can fall below $1-\alpha$.
\end{enumerate}
\end{proposition}

So the cutoff that guarantees coverage by projection is $\chi^2_L/2$, far wider than the $\chi^2_1/2$ profile cutoff we use when $L$ is large (after thinning $L$ can be 200). At an interior moment vector where the ends are differentiable, (iv) says the $\chi^2_1$ interval is asymptotically valid, even conservative, as observed by \citet{imbens2004}; it is an instance of the profile quasi-likelihood set of \citet{chen2018}, whose conditions \cref{prop:lik} checks directly in this model. The guarantee is pointwise, it can fail at a kink of an end (coverage $0.938$ for a quadrant-shaped cone at nominal $0.95$), and it excludes boundary moment vectors, which include every atomic prior once $L\ge2r$; for those, the coverage studies of \cref{sec:mi-exp} are the only evidence (Appendix~\ref{app:lik-more}).

\begin{remark}[A model-free lower bound]\label{rem:mi-lower}
With no assumption on the joint law of $(p_x,q_x)$, every encoder $r$ fitted on training data gives $I(X;Y)\ge h(\mu)-\mathrm{CE}_{\rm test}(r)$, since the held-out cross-entropy is $H(Y\mid X)+\eCE(r)$. The slack is at least the oracle's eCE. Under Assumption~\ref{ass:exch} that includes $w_0\,I$ from unseen test levels (\cref{lem:regret}), so even a near-oracle encoder recovers at most a fraction of about $1-w_0\approx1-F_1/N$ of $I$ this way; anything above that comes from extrapolating $\pi$ to unseen levels.
\end{remark}

\subsection{Without exchangeability}\label{sec:mi-pop}

When $q$ depends on popularity, Lemma~\ref{lem:rep} still holds stratum by stratum, but the strata no longer share one prior: levels seen $n$ times identify $n$ moments of their own $\pi^{\lev}_n$. The singleton stratum carries weight $F_1/N$ of $H(Y\mid X)$, and its own data identify only the mean $\mu^{\lev}_1$ of $\pi^{\lev}_1$. The strata are linked, since each $\pi^{\lev}_n$ is a Poisson-weighted mixture over popularity of the same joint law, but the link can be empty: levels so light that they are never seen twice show up only as singletons.

\begin{proposition}[No identification of the singletons' share]\label{prop:nonid}
Take any joint law of $(p_x,q_x)$ on a set of base levels of total popularity $1-\rho$, and add $M$ light levels of popularity $\rho/M$ each, with $q_x$ i.i.d.\ from a law $G$ with mean $\mu_G$, independent of everything else. Then:
\begin{enumerate}
\item[(i)] for two laws $G,G'$ with the same mean, the laws of the training fingerprint $(F_{n,k})$ differ in total variation by at most $N^2\rho^2/(2M)$;
\item[(ii)] the expected number of light levels seen once tends to $N\rho$ as $M\to\infty$;
\item[(iii)] as $G$ ranges over the laws with mean $\mu_G$, $I(X;Y)$ ranges over an interval of length $\rho\,h(\mu_G)$.
\end{enumerate}
Hence, as $M\to\infty$, every estimator $\hat I$ with values in $[0,\log2]$ has worst-case error $\E|\hat I-I|$ at least $\tfrac12\rho\,h(\mu_G)-o(1)$ over these populations: on a row share $\rho$ that the data see only as singletons, only the trivial bound $[0,h(\mu_G)]$ on the conditional entropy is identified.
\end{proposition}

The width is the singleton part of the stratified bound: if nothing links the strata, each stratum $n$ has its own prior $\pi^{\lev}_n$ identified only through its first $n$ moments, and the identified interval for $I$ is
\[
\frac{F_1}{N}\,h(\mu^{\lev}_1)+\sum_{n\ge2}\frac{nF_n}{N}\,\big(\text{width of }\E_{\pi^{\lev}_n}h\text{ given }m_{1:n}(\pi^{\lev}_n)\big)
\]
wide (in the limit of many levels per stratum), which is large when $F_1/N$ is. Under Assumption~\ref{ass:exch} the light levels would share $\pi$ with the levels that repeat, which is what makes $I(X;Y)$ estimable in this regime; a model linking $q$ to popularity, such as the frequency-aware prior, is the other way.

\paragraph{Testing exchangeability.} The assumption is testable through the moments it constrains. Under Assumption~\ref{ass:exch} every $\pi^{\lev}_n$ equals $\pi$, so for any set $S$ of level sizes the averages of $k_x/n_x$ and of $k_x(k_x-1)/(n_x(n_x-1))$ over the levels with $n_x\in S$ are unbiased estimates of the same $m_1$ and $m_2$. The \emph{stratum moment test} (\texttt{\pkgname{}.stratum\_moment\_test}) is a Wald test that these averages agree across strata of level sizes (construction in Appendix~\ref{app:exch}). It uses no model of the alternative. The likelihood-ratio test of the frequency-aware prior (\cref{sec:popularity}) is the parametric alternative, with power against a logit mean that drifts with $\log n$. On simulated exchangeable data with lognormal popularity the stratum test rejects $6\%$ of the time at level $5\%$. Both tests answer whether levels of different sizes differ, not whether the difference moves $I$ by more than the interval's width. What sets the stakes for $I$ is the singletons' share: by \cref{prop:nonid} the part of $I(X;Y)$ that rests on the assumption is about $(F_1/N)\,h(\hat\mu^{\lev}_1)$ wide, with $\hat\mu^{\lev}_1=F_{1,1}/F_1$. When this is small (below about $0.01$ nats, as on the dense features of \cref{tab:mi-bounds}) a rejection is harmless for $I$, since well-sampled strata identify their own priors; when it is large, any point estimate rests on the assumption whatever the test says, because the singletons' spread is invisible to it.

\paragraph{Relation to estimating the unseen.} The identified set is a moment linear programme, as in the estimation of symmetric properties from the fingerprint \citep{valiant2017}, but for the law of the conditionals rather than of the popularities, so unseen levels drop out and the support size of $X$ never enters (Appendix~\ref{sec:mi-vv}).

%% file: sections/07_experiments.tex
\section{Experiments: prediction}\label{sec:experiments}

We test three predictions of the theory about the encoder; \cref{sec:mi-exp} tests those about the information. Under exchangeability the type-II ML beta should be as good as the oracle on sparse levels (\cref{sec:exact}); its remaining gap should sit on levels seen many times and shrink only for priors that match higher moments (\cref{prop:expansion,prop:lower}); and when the target depends on popularity, the gap should move to the smallest levels, where the corrections of \cref{sec:popularity} act. The loss is the excess cross-entropy (\cref{eq:ece}) in nats. Three ratios put it on a common scale. The \emph{share of gain lost}, $(\eCE_{\rm method}-\eCE_{\rm oracle})/(I(X;Y)-\eCE_{\rm oracle})$, is the fraction of the oracle's improvement over the marginal predictor that a method gives up. The \emph{share of the gap closed}, $(\eCE_{\rm baseline}-\eCE_{\rm method})/(\eCE_{\rm baseline}-\eCE_{\rm oracle})$, is how much of what the baseline leaves a method recovers (averaged as a ratio of means over the training sets). In the bakeoff, which has no oracle, the \emph{share of the baseline's gain} is a method's held-out log-loss difference to the baseline divided by the baseline's improvement over the prior-only predictor. All three are ratios of small numbers, so we give nats next to them and read shares only where the denominator is not negligible.

\paragraph{Methods.} The \emph{baseline} is the beta prior centred on $\hat\mu$ with $\hat s$ fitted by type-II ML. The \emph{unseen correction} replaces its prediction for unseen levels by the singletons' label rate, the \emph{Robbins correction} also replaces the once-seen predictions by the doubleton rates when a test rejects the fitted beta (\cref{cor:unseen,cor:robbins}), and the \emph{frequency prior} is the shifted regression of \cref{sec:popularity} with its LR fallback. The richer prior is the Kiefer--Wolfowitz NPMLE on a 400-point grid with posterior-mean encoding \citep{jiang2009}, alone and with the Robbins correction on top. Competitors are the random-intercept logistic GLMM, best in the benchmark of \citet{pargent2022}, fitted by adaptive Gauss--Hermite quadrature and predicted at its conditional mode (as usual) or by its posterior mean (Appendix~\ref{app:glmm}); the $m$-estimate with $m=1$, to which CatBoost's ordered target statistic reduces at prediction time \citep{prokhorenkova2018}; fixed smoothing with the common default $m=20$; and the $m$-estimate with $m$ tuned by leave-one-out log-loss. scikit-learn's \texttt{TargetEncoder} is left out because its held-out log-loss is set by clipping (Appendix~\ref{app:details}). \Cref{tab:methods} in Appendix~\ref{app:data} lists every method with its \texttt{method=} string, tuning and gate.

\subsection{Synthetic: near-oracle and where it fails}\label{sec:synthetic}

We take the three-atom ``junk'' family of \citet{hernandez2019}: $\nlev=2500$ equally popular levels, a fraction $f$ with $q_x=1/2$ and the rest split between $q_0$ and $1-q_0$, with $q_0\sim U[0,1]$ drawn afresh for each data set and $(f,q_0)$ unknown to every method. The reference, the \emph{family oracle}, fits $(f,q_0)$ by marginal likelihood. We sweep $f\in\{0.1,\dots,0.9\}$ and $\bar n=N/\nlev$ from $0.2$ to $2$, plus $\bar n=20$, with 40 data sets per cell. This prior is atomic and far from a beta, the hardest case for the baseline that keeps exchangeability. At $\bar n\le2$ the baseline is the oracle in practice: its gap is at most about $0.001$ nats in every cell, and $0$--$9\%$ of the gain where the oracle gains at least $0.005$ nats. The NPMLE is worse than the baseline in 15 of 16 such cells: with 2,500 levels of one or two rows its extra freedom is fitted to noise. At $\bar n=20$ the higher-moment gap of \cref{prop:expansion} opens (the baseline loses $4.5$--$37\%$ of the gain) and the NPMLE is within $0.0001$ nats of the oracle (\cref{fig:junk}, Appendix~\ref{app:details}).

\subsection{Real-data subsampling oracle}\label{sec:oracle}

\paragraph{Design.} To know the true $q_x$ on real data, we keep the levels of a feature with at least 100 rows and treat the empirical distribution of their rows as the population: a row falls in level $x$ with probability $p_x$ equal to its row share and has $y\sim\Bern(q_x)$, with $q_x$ the level's rate. Training sets of $N=\bar n\,\nlev$ rows are drawn from it for $\bar n$ from $0.25$ to $20$, 50 per cell, every method is fitted to the same sets, and $\eCE$ is exact. The \emph{full oracle} is the Bayes rule under the true joint law of $(p_x,q_x)$ over the pool; the \emph{popularity-blind oracle} uses the true law of $q$ but ignores $n_x$, and the \emph{moment-matched beta} matches its first two moments. A \emph{shuffled control} permutes $q_x$ across levels, which removes the link to popularity so that exchangeability holds. The six pools (Appendix~\ref{app:data}) are MovieLens 1M movies and users ($y=\mathbb 1[\text{rating}\ge4]$), Book-Crossing users with $y=\mathbb 1[\text{rating}\ge8]$ and with $y=$ explicit versus implicit rating (\emph{rated}), Avazu \texttt{device\_ip} and Criteo \texttt{C1}, with $495$--$2{,}945$ levels. These head-only pools create sparsity by subsampling and leave out the real long tail; we therefore also build semi-synthetic \emph{head + tail} populations that keep every level with its real popularity and give the tail the real mean label rate per size bucket (Appendix~\ref{app:details}).

\paragraph{Prediction is easy where the theory says it is.} In the shuffled control the baseline is close to the moment-matched beta built from the true $q$ (\cref{fig:oracle} in Appendix~\ref{app:figs}, open squares): at $\bar n\ge1$ it loses at most $7$ percentage points of the gain more, and at most $0.0007$ nats on every pool but the smallest ($0.0023$). On MovieLens the median $\hat s$ sits on the moment-matched value (movies $4.5$ against $4.4$; users $7.7$ against $7.7$), as \cref{prop:ml} predicts. The diagnostic $(F_1+2F_2)/N$ matches the true test mass on levels seen at most once to within $0.008$ in all 96 pool-by-$\bar n$ cells.

\paragraph{What is left over has the two sources the theory names} (\cref{tab:oracle}; gaps of all methods in \cref{tab:oracle-gaps}, split by level size in \cref{fig:gap-by-n}).
\begin{itemize}
\item \emph{Prior shape.} Where $q$ hardly depends on popularity (Book-Crossing $r\ge8$, Avazu, Criteo; $|\mathrm{corr}(\log p,q)|\le0.03$), only $12$--$16\%$ of the baseline's gap is on levels seen at most once. The moment-matched beta loses nearly as much as the fitted baseline while the popularity-blind oracle loses half as much or less: the beta family, not the fit of $\hat s$, is the limit. The NPMLE closes $27\pm5\%$ of the gap on Avazu at $\bar n=2$ and $73\%$ at $\bar n=20$, but is worse than the baseline on the two smaller pools at $\bar n\le2$.
\item \emph{Popularity dependence.} Where $q$ correlates with popularity (MovieLens movies $+0.41$, users $-0.27$; Book-Crossing rated $-0.24$), $47$--$64\%$ of the gap is on levels seen at most once and the popularity-blind oracle is no better than the baseline. The frequency prior closes $50$--$81\%$ of the gap, the Robbins correction $34$--$59\%$ and the unseen correction alone $28$--$43\%$. Without popularity dependence the LR test keeps the frequency prior at the baseline, while the two corrections cost up to $14\%$ of the gap, because on pools of 500--2,300 levels the singleton and doubleton rates are noisy.
\end{itemize}

\paragraph{The long tail moves the gap to small levels.} With the tail kept (\cref{tab:oracle-sens}, Appendix~\ref{app:details}), the gap moves towards levels seen at most once on every pool, and the pools that looked popularity-free in the head are not: on Book-Crossing $r\ge8$ the share of the gap at $n_x\le1$ rises from $16\%$ to $40\%$ and the frequency prior closes $57\%$ instead of $0\%$; on Avazu it rises from $12\%$ to $23\%$, with the frequency prior closing $28\%$; on Book-Crossing rated the gap doubles to $13$ milli-nats, $86\%$ of it at $n_x\le1$, and the frequency prior closes $96\%$. With $10^5$--$10^6$ levels the NPMLE no longer loses: it closes $67$--$87\%$ of the gap on the four large pools, and $67$--$95\%$ with the Robbins correction on top.

\paragraph{Fixed smoothing.} A fixed $m$ loses in proportion to its distance from $\hat s$ (\cref{fig:fixed-m}a, Appendix~\ref{app:details}). At $\bar n=2$, $m=\hat s/2$ or $2\hat s$ gives up $0$--$30\%$ of the oracle's gain and $m=4\hat s$ $6$--$42\%$, depending on the pool. The fitted $\hat s$ itself ranges from about $3$ on Book-Crossing to over $2{,}000$ on the real Criteo pool, so the default $m=20$ gives up $23$--$40\%$ of the gain on MovieLens and Book-Crossing, but under $7\%$ on the shuffled click pools, where it is close to $\hat s$. Tuning $m$ by cross-validation instead of marginal likelihood changes the eCE by at most $1\%$ of the gain at $\bar n\ge1$ on every pool except Criteo, whose gain is tiny ($0.0013$ nats).

\begin{table}[t]
\centering
\small
\resizebox{\linewidth}{!}{\input{tables/tab_oracle}}
\caption{Real-data oracle at $\bar n=2$. ``gain'': the full oracle's improvement over the marginal predictor (milli-nats). ``baseline gap'': gap to the full oracle in milli-nats (share of the gain). ``gap at $n_x\le1$'': share of the baseline's gap on levels seen at most once. ``closed'': share of the baseline's gap closed by the unseen correction, the Robbins correction for $n\le1$, the frequency prior with the LR fallback, the NPMLE and the NPMLE with the Robbins correction, $\pm$ its standard error over the 50 paired training sets. The gaps of all methods and of the two references are in \cref{tab:oracle-gaps}.}\label{tab:oracle}
\end{table}

\subsection{Bakeoff against the literature}\label{sec:bakeoff}

\paragraph{Design.} Each feature is encoded alone and scored by held-out log-loss, so all comparisons here are of single-feature probabilities. The 24 single-feature tasks come from nine sources: the click logs Avazu (\texttt{device\_ip} and the ``user'' \texttt{device\_ip}$\times$\texttt{device\_model}) and Criteo (\texttt{C1}, \texttt{C20}, \texttt{C22}), Book-Crossing (seven feature-target pairs), Amazon employee access (three columns), MovieLens (two) and seven OpenML features from four sources, including the four-class \texttt{employee\_salaries}, fitted one-vs-rest. Every task uses stratified 5-fold cross-validation repeated 5 times, the same 25 splits for every method; unseen levels get the encoder's prediction for a level with no data. Differences from the baseline are tested by the corrected resampled $t$-test \citep{nadeau2003,bouckaert2004} with Holm's correction over the 11 comparisons with the baseline on each task, at $p_{\rm Holm}<0.01$. Because tasks from one source share rows or targets, \cref{tab:bakeoff-summary} counts significant results both by task and by source and gives medians over all 24 tasks and over the nine sources, in milli-nats and as relative log-loss. Shares of the baseline's gain are summarised over a fixed set of 20 tasks that leaves out the four where the baseline gains less than $0.5\%$ of $H(Y)$. Per-task results, with corrected 95\% intervals in milli-nats, are in Appendix~\ref{app:details}.

\begin{table}[t]
\centering
\small
\resizebox{\linewidth}{!}{\input{tables/tab_bakeoff_summary}}
\caption{Bakeoff summary: each method against the baseline on 24 real features, 25 paired splits per task. ``mnats'': held-out log-loss difference to the baseline in milli-nats per test row (negative is better); ``rel.\ \%'': the same as a percentage of the baseline's log-loss; ``median over 9 sources'': median of each source's median difference; ``\% of gain'': difference as a share of the baseline's gain over the prior-only predictor, median over the 20 tasks where that gain exceeds $0.5\%$ of $H(Y)$; ``better/worse'': tasks, and sources with at least one such task, where the method is significantly better / worse than the baseline ($p_{\rm Holm}<0.01$). Per-task results: \cref{tab:bakeoff,tab:bakeoff-full,tab:bakeoff-abs}.}\label{tab:bakeoff-summary}
\end{table}

\paragraph{Results.}
\begin{itemize}
\item \emph{Nothing beats the baseline by much, and the richer priors are equivalent to it or better.} The NPMLE is better than the baseline on 14 tasks from 6 sources and worse on none, by a median of $0.29$ milli-nats per test row ($0.06\%$ of the log-loss, $2.1\%$ of the baseline's gain), at most $6.9\%$ of the gain where the baseline gains more than $0.03$ nats (Book-Crossing rated, whose rates are bimodal). The frequency prior is about as good (14 tasks, 7 sources). The largest difference between fitted priors on any task is about $0.01$ nats. Taking a tenth of that, $1$ milli-nat per test row, as the margin of equivalence, the corrected 95\% intervals of \cref{tab:bakeoff-abs} exclude a loss larger than the margin on every task for the unseen and Robbins corrections, the frequency prior and the GLMM's posterior mean, and on all tasks but one for the NPMLE (\texttt{churn/state}, up to $1.7$ milli-nats).
\item \emph{A large part of their edge is on unseen levels, and the corrections capture it.} Summed over the fixed task set, $49\%$ of the NPMLE's improvement and $61\%$ of the frequency prior's come from test rows on unseen levels, where the baseline predicts $\hat\mu$ although their rate differs from it, as \cref{prop:shift} says. The unseen and Robbins corrections are each better than the baseline on 12 tasks and worse on none. Compared directly, the Robbins correction ties the NPMLE (better on 4 tasks, worse on 5; \cref{tab:wtl}); on top of the NPMLE it is better than the baseline on 14 tasks by a median of $0.50$ milli-nats, and better than the NPMLE on 7 and worse on none.
\item \emph{The GLMM is a good prior with a poor prediction rule.} Predicted at its conditional mode, with $\operatorname{expit}(b_0)$ for unseen levels as \texttt{lme4}'s \texttt{predict} does, the GLMM is better than the baseline on 8 tasks and worse on 7: it gains on dense features but gives back $16$--$31\%$ of the gain on Avazu and $10$--$13\%$ on Criteo, where most test rows fall on unseen and once-seen levels.  For a logit-normal prior $\operatorname{expit}(b_0)$ is the median of $q$, not its mean. The same fit predicted by the posterior mean is better than the baseline on 16 tasks and worse on 2, by $0.3\%$ and $0.1\%$ of the gain, close to the NPMLE.
\item \emph{Fitted smoothing ties; a poorly chosen fixed $m$ loses.} Leave-one-out cross-validation of $m$ is within $1\%$ of the baseline's gain on every task, at about three times the cost. Fixed $m=20$ is worse on 15 tasks, by a median of $0.55$ milli-nats ($3.4\%$ of the gain), but up to $31$--$41\%$ on Avazu. Over the tasks the loss grows with $|\log(m/\hat s)|$ (\cref{fig:fixed-m}b): a factor of two costs a median of $2$--$3\%$ of the gain and a factor of four $11$--$14\%$, while $\hat s$ ranges from about $1$ to $109$ across tasks (\cref{tab:fixed-m}). The $m$-estimate with $m=1$ (CatBoost's prediction-time statistic) is worse on 19 tasks.
\end{itemize}
In absolute terms the stakes are small, and the click features carry only $2$--$10\%$ of $H(Y)$. In click-through prediction an improvement of about $0.001$ in log-loss is often treated as meaningful; on the Avazu features the NPMLE with the Robbins correction gains $1.1$--$1.5$ milli-nats and fixed $m=20$ loses $11$--$19$, while on Criteo all fitted priors are within $0.1$ milli-nats of the baseline.

\subsection{Downstream models}\label{sec:downstream}

On one data set, Amazon employee access, with every column (and in a second set all 28 pairwise concatenations) replaced by its out-of-fold encoding (Appendix~\ref{app:downstream}), the single-feature ranking carries over to an untuned logistic regression only weakly: the NPMLE beats the baseline by $1.2$--$1.9$ milli-nats, while the GLMM at its conditional mode and fixed $m=20$ lose $4.5$ and $3.5$ milli-nats on the sparse pairwise columns. With gradient boosting the encoders are indistinguishable except the GLMM at its mode. A label-free count feature $\log(1+n_x)$ gains $1.8$--$4.9$ milli-nats with boosting and up to $3.3$ with logistic regression, more than most changes of prior, and adding the pairwise columns gains about ten times the spread between encoders. On this data set the choice of encoder does not drive downstream accuracy.

%% file: tables/tab_oracle.tex
\begin{tabular}{l r r r r r r r r r }
\toprule
pool & corr & gain & baseline gap & gap at $n_x\le1$ & closed: +unseen & +Robbins & freq.\ prior & NPMLE & NPMLE +Robbins \\
\midrule
MovieLens movies & +0.41 & 32.2 & 7.9 (24\%) & 64\% & 43 $\pm$ 1\% & 59 $\pm$ 1\% & 81 $\pm$ 1\% & 14 $\pm$ 1\% & 41 $\pm$ 2\% \\
MovieLens users & -0.27 & 18.2 & 2.1 (12\%) & 57\% & 32 $\pm$ 2\% & 36 $\pm$ 3\% & 81 $\pm$ 2\% & -9 $\pm$ 3\% & 9 $\pm$ 5\% \\
Book-Crossing users, rated & -0.24 & 76.3 & 6.1 (8\%) & 47\% & 28 $\pm$ 1\% & 34 $\pm$ 2\% & 50 $\pm$ 1\% & 21 $\pm$ 2\% & 35 $\pm$ 2\% \\
Book-Crossing users, $r\geq8$ & -0.03 & 71.5 & 3.9 (5\%) & 16\% & -9 $\pm$ 3\% & -9 $\pm$ 3\% & -0 $\pm$ 0\% & -72 $\pm$ 18\% & -79 $\pm$ 19\% \\
Avazu device\_ip & +0.02 & 9.8 & 2.2 (22\%) & 12\% & 0 $\pm$ 1\% & -1 $\pm$ 1\% & -1 $\pm$ 1\% & 27 $\pm$ 5\% & 24 $\pm$ 5\% \\
Criteo C1 & -0.02 & 1.3 & 0.7 (59\%) & 12\% & -11 $\pm$ 4\% & -14 $\pm$ 5\% & -0 $\pm$ 0\% & -17 $\pm$ 9\% & -34 $\pm$ 11\% \\
\bottomrule
\end{tabular}

%% file: tables/tab_bakeoff_summary.tex
\begin{tabular}{l r r r r r r}
\toprule
 & \multicolumn{2}{c}{median over 24 tasks} & median over & median \% of & tasks & sources\\
method & mnats & rel.\ \% & 9 sources (mnats) & gain (20 tasks) & better/worse & better/worse\\
\midrule
+unseen & -0.08 & -0.032 & -0.11 & -1.3 & 12/0 & 4/0\\
+Robbins & -0.10 & -0.039 & -0.10 & -1.7 & 12/0 & 4/0\\
NPMLE & -0.29 & -0.061 & -0.42 & -2.1 & 14/0 & 6/0\\
NPMLE +Robbins & -0.50 & -0.075 & -0.69 & -3.0 & 14/0 & 5/0\\
freq.\ prior & -0.16 & -0.035 & -0.45 & -1.1 & 14/0 & 7/0\\
GLMM mode & +0.04 & +0.009 & +0.66 & +0.3 & 8/7 & 3/3\\
GLMM mean & -0.26 & -0.060 & -0.25 & -1.7 & 16/2 & 7/2\\
$m{=}20$ & +0.55 & +0.082 & +0.56 & +3.4 & 1/15 & 1/6\\
CV $m$ & -0.01 & -0.002 & -0.02 & -0.1 & 3/0 & 2/0\\
$m{=}1$ (CatBoost) & +4.85 & +1.649 & +6.86 & +22.4 & 0/19 & 0/8\\
\bottomrule
\end{tabular}

%% file: sections/08_mi_experiments.tex
\section{Experiments: information}\label{sec:mi-exp}

Section~\ref{sec:mi} makes three predictions about $I(X;Y)$. The credible interval of the beta model should under-cover when the prior is far from a beta, because its spread ignores the identification width. The likelihood interval should cover whenever Assumption~\ref{ass:exch} holds, and its width should shrink to the identified width as levels repeat. And it should fail under popularity dependence, where levels of different sizes no longer share one prior (\cref{sec:mi-pop}). We test them on the synthetic priors of \cref{tab:id} and on the real-data pools of \cref{sec:oracle}.

The plug-in estimate of $I(X;Y)$ is far off in this regime: on the MovieLens oracle it is off by $+0.03$ to $+0.32$ nats at $\bar n\ge1$, while the baseline's estimate is within $\pm0.006$ nats. Classical bias corrections (Miller--Madow, Chao--Shen) remove only part of that bias, and NSB, even given the number of levels, is several times further off than the baseline on the synthetic populations at $\bar n=1$ (Appendix~\ref{app:mi-point}).

\paragraph{Coverage of the intervals.} We check how often the intervals of \cref{sec:mi} contain the population value $I=h(\sum_xp_xq_x)-\sum_xp_xh(q_x)$. The \emph{likelihood interval} uses priors on a 101-point grid and the cutoff $\chi^2_{1,0.95}/2$, with levels above 200 rows thinned to 200 by a hypergeometric draw, which leaves the estimand unchanged; Appendix~\ref{app:mi-sens} checks both choices. The \emph{credible interval} of \citet{hernandez2019} is the posterior mean $\pm1.96$ posterior standard deviations under the beta prior, centred on the baseline's estimate. The \emph{identified interval} uses the population's own moments up to $L\le4$: the width that would remain if those moments were known. The synthetic populations draw $q_x$ from the five priors of \cref{tab:id}, with $10^4$ or $10^5$ levels of equal or lognormal popularity, $\bar n\in\{0.5,1,2,5\}$ and 100 data sets per cell; the oracle pools are those of \cref{sec:oracle}, real and shuffled, with 50 training sets per cell. \Cref{tab:mi-coverage} gives coverage and width at $\bar n=1$ and $5$; the moment interval, three cases in detail and the bias of the point estimates are in Appendices~\ref{app:mi-point} and~\ref{app:figs}.
\begin{itemize}
\item \emph{As data grow, the credible interval stays calibrated only when the prior is a beta} ($0.90$--$0.97$ for both beta priors). For the atomic priors and on Avazu and Book-Crossing rated the baseline's estimate carries the identification bias of \cref{tab:id}, while the interval shrinks like the inverse square root of the number of levels, so coverage falls as data grow, to $0.03$ for the three-atom prior already at $\bar n=1$.
\item \emph{Once levels repeat, little of the uncertainty is identification.} Once at least 100 levels are seen four times, the identified interval is $0.002$--$0.020$ nats wide (``ident.''), so a much wider interval, such as the moment interval (Appendix~\ref{app:mi-point}), is mostly sampling noise in the higher moments.
\item \emph{The likelihood interval removes most of that noise and covers under Assumption~\ref{ass:exch}.} It is a median $0.23$ times as wide as the moment interval and, at $\bar n=5$, close to the credible interval. Its coverage is $0.91$--$1.00$ in every synthetic cell (median $0.98$) and at least $0.90$ on every shuffled pool except Criteo at $\bar n=20$ ($0.64$; one level holds $37\%$ of the rows, so the finite pool is far from the many-level regime of \cref{prop:lik}). For the beta priors and the shuffled pools this is consistent with \cref{prop:lik}(iv), although the theorem fixes $L$ while the thinned levels here let it grow to 200; for the junk and three-atom priors, boundary points that (iv) does not cover, it is an empirical finding.
\item \emph{It tends to fail where \cref{sec:mi-pop} says it should.} Under popularity dependence the interval shrinks onto the value implied by the exchangeable model: on real MovieLens movies, where the popular levels are more polarized, coverage drops to $0.12$ at $\bar n=20$, every miss above $I$, and on real Avazu and Book-Crossing rated to $0.26$ and $0.30$, while the shuffled pools stay at $0.96$. The failure is a tendency, not a rule: Book-Crossing rated covers only $0.76$ already at $\bar n=1$, and MovieLens users, whose popularity dependence is weaker, still covers $0.92$ at $\bar n=20$. Criteo \texttt{C1} is degenerate for any exchangeable method, since one level holds $37\%$ of the rows.
\item \emph{The NPMLE gives the least biased point estimate once levels repeat} (\cref{tab:mi-point}): at $\bar n=5$ its bias is at most $0.8$ milli-nats on the synthetic priors, against up to $5.9$ for the baseline. At $\bar n=1$ no estimator dominates, because with $n_x\le2$ the NPMLE picks (to within $2\times10^{-4}$ nats) the maximum-entropy point of the identified interval (\cref{prop:npmle-maxent}, Appendix~\ref{app:npmle}), no less an assumption than the beta.
\end{itemize}

\paragraph{Testing exchangeability first.} Since the likelihood interval fails only when Assumption~\ref{ass:exch} fails, we ran the stratum moment test and the likelihood-ratio test of \cref{sec:mi-pop} on the same training sets and on the full data of the bakeoff tasks (Appendix~\ref{app:exch}). In nine of the ten pool cells where the likelihood interval covers less than $0.85$, one of the two tests rejects in $56$--$100\%$ of the training sets; on the shuffled pools they reject in $0$--$16\%$ (up to $70\%$ on Criteo), so their size is only roughly nominal. They also flag dependence too weak to matter (MovieLens users), and an effect size does not sort the cases: at $\bar n=20$ the largest deviation of a stratum's label rate from the pooled one is $0.10$ on MovieLens users and $0.03$ on Avazu, yet the likelihood interval covers $0.92$ and $0.26$. So we do not use the effect size as a rule. On the full data the stratum test rejects at $p<10^{-3}$ on 21 of the 23 tasks where it can run.

\paragraph{What can be said with weaker assumptions.} For these features we report two statements that do not need exchangeability across level sizes (\cref{tab:mi-bounds}, Appendix~\ref{app:exch}). The predictive lower bound of Remark~\ref{rem:mi-lower}, $h(\hat\mu)$ minus the held-out cross-entropy of the best bakeoff encoder, holds for any joint law; picking the best of the encoders on the same folds biases it slightly upwards. The stratified identified interval of \cref{sec:mi-pop} gives each group of level sizes its own prior, identified by up to four of its own moments at their plug-in values, so it assumes exchangeability within each group and leaves out sampling error; the singletons contribute only the trivial range of \cref{prop:nonid}. Their combination is a median $0.027$ nats wide, but $0.02$--$0.45$ nats (median $0.09$) on the eleven features where singletons hold more than $10\%$ of the rows, and the singletons make up a median $83\%$ of the width. On 8 of the 23 features the exchangeable estimates of the baseline and the NPMLE fall below the interval. On sparse real features $I(X;Y)$ is therefore known only to within several hundredths of a nat without an assumption that links levels of different sizes; the likelihood interval assumes one.

\begin{table}[t]
\centering
\small
\resizebox{\linewidth}{!}{\input{tables/tab_mi_coverage}}
\caption{Coverage of the true $I(X;Y)$ by the 95\% credible interval of \citet{hernandez2019} (``cred.'') and the likelihood interval of \cref{sec:mi} (``lik.''), with mean widths in nats; ``ident.'': mean width of the identified interval from the population's own moments up to the largest $L\le4$ seen by at least 100 levels (the likelihood interval also uses larger levels, so it can be narrower than ``ident.''). The moment interval is in \cref{tab:mi-coverage-full}. Synthetic: 100 data sets per cell; oracle pools: 50 training sets per cell (binomial standard error of a coverage near $0.95$: $0.02$ and $0.03$). $I$ is the population value.}\label{tab:mi-coverage}
\end{table}

%% file: tables/tab_mi_coverage.tex
\begin{tabular}{lr ccccc ccccc}
\toprule
 & & \multicolumn{5}{c}{$\bar n=1$} & \multicolumn{5}{c}{$\bar n=5$}\\
\cmidrule(lr){3-7}\cmidrule(lr){8-12}
population & $I$ & cred. & width & lik. & width & ident. & cred. & width & lik. & width & ident.\\
\midrule
\multicolumn{12}{l}{\emph{synthetic, $10^5$ equally popular levels}}\\
junk $f{=}0.5$ & 0.184 & 0.94 & 0.012 & 0.95 & 0.028 & 0.013 & 0.44 & 0.003 & 0.98 & 0.004 & 0.013\\
junk $f{=}0.8$ & 0.074 & 0.45 & 0.011 & 0.97 & 0.020 & 0.006 & 0.00 & 0.003 & 0.99 & 0.004 & 0.006\\
atoms $\{.2,.3,.85\}$ & 0.163 & 0.03 & 0.012 & 0.97 & 0.019 & 0.003 & 0.00 & 0.003 & 0.99 & 0.004 & 0.003\\
$\mathrm{Beta}(0.5,0.5)$ & 0.307 & 0.96 & 0.012 & 1.00 & 0.035 & 0.020 & 0.93 & 0.003 & 1.00 & 0.009 & 0.020\\
$\mathrm{Beta}(1,3)$ & 0.104 & 0.96 & 0.011 & 1.00 & 0.027 & 0.014 & 0.96 & 0.003 & 0.98 & 0.006 & 0.014\\
\midrule
\multicolumn{12}{l}{\emph{oracle pools, real}}\\
MovieLens movies & 0.087 & 0.94 & 0.065 & 0.98 & 0.077 & 0.005 & 0.94 & 0.019 & 0.94 & 0.027 & 0.004\\
MovieLens users & 0.064 & 0.94 & 0.053 & 0.92 & 0.062 & 0.003 & 0.90 & 0.014 & 0.96 & 0.017 & 0.003\\
BX users, rated & 0.141 & 0.68 & 0.070 & 0.76 & 0.092 & 0.019 & 0.52 & 0.022 & 1.00 & 0.035 & 0.015\\
BX users, $r\geq8$ & 0.144 & 0.94 & 0.138 & 0.94 & 0.172 & 0.270 & 0.92 & 0.043 & 0.94 & 0.062 & 0.012\\
Avazu device\_ip & 0.040 & 0.42 & 0.041 & 0.98 & 0.058 & 0.010 & 0.20 & 0.019 & 0.98 & 0.019 & 0.008\\
Criteo C1 & 0.004 & 0.70 & 0.023 & 0.98 & 0.054 & 0.007 & 1.00 & 0.017 & 0.94 & 0.024 & 0.002\\
\midrule
\multicolumn{12}{l}{\emph{oracle pools, shuffled $q$}}\\
MovieLens movies & 0.099 & 0.92 & 0.066 & 0.92 & 0.075 & 0.005 & 0.90 & 0.019 & 0.98 & 0.024 & 0.004\\
MovieLens users & 0.061 & 0.92 & 0.053 & 0.96 & 0.061 & 0.003 & 0.94 & 0.015 & 1.00 & 0.017 & 0.003\\
BX users, rated & 0.131 & 0.96 & 0.067 & 0.98 & 0.082 & 0.016 & 0.94 & 0.022 & 0.98 & 0.033 & 0.013\\
BX users, $r\geq8$ & 0.117 & 0.94 & 0.139 & 0.92 & 0.169 & 0.266 & 0.98 & 0.042 & 0.92 & 0.062 & 0.010\\
Avazu device\_ip & 0.046 & 0.82 & 0.048 & 1.00 & 0.054 & 0.011 & 0.70 & 0.019 & 0.94 & 0.019 & 0.008\\
Criteo C1 & 0.017 & 0.02 & 0.023 & 0.98 & 0.060 & 0.018 & 0.00 & 0.011 & 1.00 & 0.022 & 0.006\\
\bottomrule
\end{tabular}

%% file: sections/09_discussion.tex
\section{Discussion}\label{sec:discussion}

\paragraph{Prediction is easy, information is hard.} Both halves of the paper rest on one fact: a level seen $n$ times reveals $n$ moments of the prior on its conditional, and the prediction for it needs $n+1$. The levels that carry most of the test mass of a sparse feature are seen zero, one or two times, and under exchangeability a beta prior fitted by marginal likelihood supplies the two or three moments they need. Without exchangeability the oracle there is still read off the singleton and doubleton counts (\cref{cor:robbins}). For the information it is bad news: $I(X;Y)$ depends on the whole prior through $\E_\pi h(q)$, sparse levels pin it down only to an interval (\cref{prop:pi}), and without exchangeability the singletons' share is not identified at all (\cref{prop:nonid}). On the full real features, where exchangeability is rejected almost everywhere, the stratified identified interval is a few hundredths of a nat wide on the sparse features and up to $0.45$ nats on the sparsest (\cref{sec:mi-exp}).

\paragraph{When richer priors help.} As the theory predicts, \emph{prior shape} matters on large levels, where the NPMLE closes the gap the beta leaves once many levels identify higher moments, and \emph{popularity} matters on small levels, where the Robbins correction captures much of the same gain from the singleton and doubleton counts; on large sparse features the two add up. Elsewhere the richer priors gain little over the one-parameter baseline: a median of a few percent of its gain, and at most about $0.01$ nats per test row, in single-feature prediction. Downstream, boosting erased most of these differences on the one data set we tried (\cref{sec:downstream}).

\paragraph{Fit the smoothing; predict GLMMs by the posterior mean.} A badly chosen fixed $m$ is costly. How $m$ is fitted matters little (\cref{sec:bakeoff}). A fixed $m$ loses in proportion to its distance from the fitted value: a factor of two costs a median of $2$--$3\%$ of the gain over the bakeoff tasks and a factor of four $11$--$14\%$ (\cref{fig:fixed-m}). Because $\hat s$ ranges from about $1$ to over $100$ across real features no single default fits: the common $m=20$ is harmless on features with $\hat s$ near 20 but loses $13$--$23\%$ of the gain on sparse features with $\hat s\approx3$--$4$ and $31$--$41\%$ on the Avazu features (\cref{tab:fixed-m}). The GLMM's logit-normal prior is not the problem: predicted by the posterior mean it is close to the NPMLE, while at the usual conditional mode it gives back up to $31\%$ of the gain on singleton-heavy click data.

\paragraph{A practical recipe.} For a high-cardinality feature and a binary target:
\begin{enumerate}
\item Fit the smoothing strength instead of fixing it, by beta-binomial marginal likelihood (cheaper than cross-validation; consistent under exchangeability, \cref{prop:ml}).
\item Compute $(F_1+2F_2)/N$ from the training counts: the share of test rows on levels seen at most once. Under exchangeability the fitted encoder is asymptotically the oracle there; otherwise step 3 does.
\item Predict unseen levels by the label rate of the singletons and once-seen levels from the doubletons (the Robbins correction, with its likelihood-ratio gate). Use the frequency-aware prior when its likelihood-ratio test detects popularity dependence that reaches larger levels. In our bakeoff it was significantly better than the baseline on every feature with at least $2\times10^4$ levels and $(F_1+2F_2)/N>0.03$, and on none with fewer than $10^4$ levels (thresholds read off the same 24 tasks). It was within $1$ milli-nat of the baseline or better on every task, but on oracle pools without popularity dependence its gate let through losses of up to $14\%$ of the gap (\cref{tab:oracle}).
\item Use the NPMLE when many levels, say thousands, are seen three or more times. In the bakeoff it was better than the baseline on 14 of 24 features, and any loss stayed within the $1$ milli-nat margin on all but one; it was slightly worse on small oracle pools with one or two rows per level. Keep step 3 on top.
\item For $I(X;Y)$, report intervals, not a point: the likelihood interval with the NPMLE estimate (both assume exchangeability; the beta's credible interval is calibrated only for near-beta priors) next to the stratified identified interval of \cref{sec:mi-exp}, with the stratum test's effect size (\cref{sec:mi-pop}); the test rejects on almost every large feature, so it should not decide alone.
\end{enumerate}
Steps 1--4 rest on single-feature comparisons; all five are implemented in \texttt{\pkgname{}}.

\paragraph{Limitations.}
\begin{itemize}
\item \emph{Exchangeability is rejected on real data.} The exactness results and the information estimates assume it, the corrections of \cref{sec:popularity} do not; without it only the wide stratified interval is available.
\item \emph{What is proved.} Consistency of the type-II ML fit is proved only when every level is seen at most twice; with larger levels we show numerically that it stays close when levels are equally popular, and less so under heavy-tailed level sizes (\cref{rem:mixed}). The likelihood interval's $\chi^2_1$ cutoff is justified asymptotically and pointwise, at interior moment vectors where the ends of the identified set are differentiable (\cref{prop:lik}); uniform validity and a finite-sample guarantee are open, coverage at atomic priors is empirical, and the implementation uses a grid of priors with large levels thinned (checked in Appendix~\ref{app:mi-sens}). The exchangeability tests are only roughly of nominal size (up to $0.16$ on the shuffled pools, $0.70$ on Criteo). \Cref{prop:npmle-maxent} is proved for mirror ascent; the code uses EM.
\item \emph{The frequency-aware prior is parametric}; a nonparametric prior per level size is the natural extension.
\item \emph{The real-data oracle cannot know the tail.} The head + tail populations invent the tail's conditionals around their real mean, so they show the direction of popularity effects in the tail, not their size, and their tail shifts logits by popularity bucket, close to the frequency prior's form, which favours that prior. The bakeoff's random splits cannot check that the singletons' rate suits levels that first appear later in time.
\item \emph{Scope.} Fitted priors differ by at most about $0.01$ nats per test row, and the downstream study covers one data set with untuned models. With $k_y>2$ classes a Dirichlet has one concentration for a matrix-valued second moment, so exactness fails already at $n=1$; with several features the right prior for one may depend on the others.
\end{itemize}

%% file: sections/appendix.tex
\section{Proofs}\label{app:proofs}

\subsection{Section~\ref{sec:setup}}

\begin{proof}[Proof of Lemma~\ref{lem:decomp}]
By Assumption~\ref{ass:exch}, $n_x$ is independent of $q_x$, and given $n_x=n$ the pair $(q_x,k_x)$ has the law of $(q,K)$ in the definition of $R_n$. Grouping \eqref{eq:ece} by the value of $n_x$ gives the first identity. For the second, $\binom{N+1}{n+1}p^{n+1}(1-p)^{N-n}=\frac{N+1}{n+1}\,p\binom Nn p^n(1-p)^{N-n}$; sum over $x$.
\end{proof}

\subsection{Section~\ref{sec:exact}}\label{app:proofs-exact}

\begin{proof}[Proof of Lemma~\ref{lem:moments}]
$P(K=k\mid q,n)=\binom nk q^k(1-q)^{n-k}$; integrate against $\pi$. The predictive is $\E[q\mid K=k]=\E_\pi[q^{k+1}(1-q)^{n-k}]/\E_\pi[q^k(1-q)^{n-k}]$. The moment dependence follows from the expansion of $M_\pi(a,b)$.
\end{proof}

\begin{proof}[Proof of Lemma~\ref{lem:regret}]
Condition on $K=k$ and write $\bar p=p_\pi(k,n)=\E[q\mid K=k]$. Since $\kl(q\|r)=-h(q)-q\log r-(1-q)\log(1-r)$ is affine in $q$ apart from $-h(q)$,
\[
\E\big[\kl(q\|r)\mid K=k\big]=\E\big[\kl(q\|\bar p)\mid K=k\big]+\kl(\bar p\|r).
\]
Averaging over $K$ gives the first claim. The first term averaged over $K$ is $\E[\KL(P_{Y_{\rm new}\mid q}\|P_{Y_{\rm new}\mid K})]=I(q;Y_{\rm new}\mid K)$.
\end{proof}

\begin{proof}[Proof of Proposition~\ref{prop:exact}]
By Lemma~\ref{lem:moments} the predictive at $n\le1$ uses only $m_1,m_2$, and the law of $K$ at $n\le2$ uses only $m_1,m_2$; $B^\pi$ matches both. Lemma~\ref{lem:regret} then gives $\Delta_n=0$.
\end{proof}

\paragraph{Explicit form of Proposition~\ref{prop:n2}.} With $D_2=m_2$, $D_1=m_1-m_2$, $D_0=1-2m_1+m_2$,
\[
p_\pi(k,2)-p_{B^\pi}(k,2)=(-1)^k\,\frac{\delta_3}{D_k},\qquad
\Delta_2(p_{B^\pi})=\frac{\delta_3^2}{2}\sum_{k=0}^{2}\binom2k\frac{D_k}{M_\pi(k+1,2-k)\,M_\pi(k,3-k)}+O(\delta_3^3).
\]

\begin{proof}[Proof of Proposition~\ref{prop:n2}]
The denominators $D_k=M(k,2-k)$ involve only $m_1,m_2$ and agree. The numerators $M(3,0)=m_3$, $M(2,1)=m_2-m_3$ and $M(1,2)=m_1-2m_2+m_3$ shift by $+\delta_3,-\delta_3,+\delta_3$. Since $P(k\mid2)=\binom2kD_k>0$, the regret vanishes iff every predictive agrees, iff $\delta_3=0$. The expansion uses $\kl(p\|p+\epsilon)=\epsilon^2/(2p(1-p))+O(\epsilon^3)$ and $D_k\,p_k(1-p_k)=M(k+1,2-k)M(k,3-k)/D_k$. Equation~\eqref{eq:k3} is the beta's third central moment written with $s=a+b$ and $\mu=a/s$.
\end{proof}

\begin{proof}[Proof of Corollaries~\ref{cor:sym} and~\ref{cor:where}]
For a symmetric $\pi$, $\mu=1/2$ and both third central moments vanish, so $\delta_3=0$. Corollary~\ref{cor:where} combines Lemma~\ref{lem:decomp} with Propositions~\ref{prop:exact} and~\ref{prop:n2}.
\end{proof}

\begin{proof}[Proof of Proposition~\ref{prop:ml}]
(1) For $n=1$, $P(k=1)=\hat\mu$ for every $t$, as noted by \citet{hernandez2019}. (2) $P_t(2)=\mu(\mu s+1)/(s+1)=\mu(\mu+(1-\mu)t)$, and similarly for the other cells. A sum of logarithms of affine functions is concave, strictly so because $\partial_tP_t(1)\neq0$. (3) The population objective is $L(t;\mu)=\sum_kP_\pi(k\mid2)\log P_t(k)$. At $\mu=m_1$, Proposition~\ref{prop:exact} gives $P_{t_\pi}(\cdot)=P_\pi(\cdot\mid2)$, so by Gibbs' inequality $t_\pi$ maximises $L(\cdot;m_1)$, uniquely by strict concavity. The empirical objective $\ell/F_2$ converges pointwise to $L(\cdot;\mu)$ by the law of large numbers and continuity in $\hat\mu$, and pointwise convergence of concave functions to a concave limit with a unique maximiser implies convergence of the maximisers \citep[Thm.~2.7]{newey1994}.
\end{proof}

\paragraph{Explicit form of Proposition~\ref{prop:expansion}.} Write $\delta M(a,b)=\sum_i\binom bi(-1)^i\delta_{a+i}$ and, for each $k$, $A_k=M_\pi(k+1,n-k)$, $D_k=M_\pi(k,n-k)$ and $C_k=M_\pi(k,n+1-k)$. Then
\[
\Delta_n(p_{\tilde\pi})=\frac12\sum_{k=0}^{n}\binom nk\frac{\big(D_k\,\delta M(k+1,n-k)-A_k\,\delta M(k,n-k)\big)^2}{D_k\,A_k\,C_k}+O(\|\delta\|^3).
\]

\begin{proof}[Proof of Proposition~\ref{prop:expansion}]
$p_\pi(k,n)-p_{\tilde\pi}(k,n)=A_k/D_k-(A_k-\delta A_k)/(D_k-\delta D_k)=(D_k\,\delta A_k-A_k\,\delta D_k)/D_k^2+O(\delta^2)$. Insert in $\kl(p\|p+\epsilon)=\epsilon^2/(2p(1-p))+O(\epsilon^3)$ with $p(1-p)=A_kC_k/D_k^2$, and weight by $P_\pi(k\mid n)=\binom nkD_k$ (Lemma~\ref{lem:regret}).
\end{proof}

\begin{proof}[Proof of Proposition~\ref{prop:lower}]
By Lemma~\ref{lem:moments}, every level's count distribution, and hence the joint law of the training data, depends only on $m_1,\dots,m_{n_{\max}}$. So $\hat r$ has the same law under both priors, as does $K$ at $n_{\max}$. Bound the maximum by the average, apply Lemma~\ref{lem:regret} to each prior, and minimise the sum pointwise in $r$. The minimum is positive for every $k$ with $p_{\pi_1}(k)\neq p_{\pi_2}(k)$, and such a $k$ exists since $m_{n_{\max}+1}$ differs; e.g.\ $k=n_{\max}$, where $p_{\pi_i}=m_{n_{\max}+1}/m_{n_{\max}}$.
\end{proof}

\begin{proof}[Proof of the bound in Remark~\ref{rem:w1}]
Let $P$ and $\tilde P$ be the joint laws of $(K,Y_{\rm new})$ for a level of size $n$ under $\pi$ and $\tilde\pi$. By the chain rule, $\KL(P\|\tilde P)=\KL(P_K\|\tilde P_K)+\E_{P}\,\kl\big(p_\pi(K,n)\,\|\,p_{\tilde\pi}(K,n)\big)\ge\Delta^\pi_n(p_{\tilde\pi})$, the last step by Lemma~\ref{lem:regret}. For any coupling $\gamma$, $P=\int B_q\,d\gamma$ and $\tilde P=\int B_{\tilde q}\,d\gamma$, where $B_q$ is the law of $(K,Y_{\rm new})$ given the conditional $q$, the image of $n+1$ independent $\Bern(q)$ draws. Joint convexity of the KL divergence and data processing give $\KL(P\|\tilde P)\le\int\KL(B_q\|B_{\tilde q})\,d\gamma\le(n+1)\int\kl(q\|\tilde q)\,d\gamma$. Taking the infimum over $\gamma$ gives the first inequality. The second follows from $\kl(a\|b)\le(a-b)^2/(b(1-b))$, the $\chi^2$ bound, with $b(1-b)\ge\epsilon(1-\epsilon)$ on the support of $\tilde\pi$, and the third from $(q-\tilde q)^2\le|q-\tilde q|$ on $[0,1]$.
\end{proof}

\subsection{Section~\ref{sec:popularity}}

\begin{proof}[Proof of Lemma~\ref{lem:strat}]
Group the sum in \eqref{eq:ece} by the value of $n_x$. Given $n_x=n$ and $q_x$, the count $k_x\sim\Bin(n,q_x)$ does not depend on $p_x$, so $p_x$ enters only through the weight $p_x\Bin(n\mid N,p_x)$, which is absorbed into $\pi^*_n$.
\end{proof}

\begin{proof}[Proof of Proposition~\ref{prop:shift}]
$p\binom Nnp^n(1-p)^{N-n}=\frac{n+1}{N+1}\binom{N+1}{n+1}p^{n+1}(1-p)^{N-n}$. Insert in the definition of $\pi^*_n$; the constant $(n+1)/(N+1)$ cancels. For the Robbins form, $\E F_{n+1,j}\propto\binom{n+1}{j}M^*_n(j,n+1-j)$, so $(k+1)\E F_{n+1,k+1}\propto(n+1)\binom nkM^*_n(k+1,n-k)$ and $(n+1-k)\E F_{n+1,k}\propto(n+1)\binom nkM^*_n(k,n+1-k)$. Their ratio to the sum is $M^*_n(k+1,n-k)/M^*_n(k,n-k)=p^*(k,n)$, since $M(k+1,n-k)+M(k,n+1-k)=M(k,n-k)$.
\end{proof}

\begin{proof}[Proof of Lemma~\ref{lem:train-shift}]
By the identity above, $\pi^*_n$ is $\nu_{n+1}$ reweighted by $1-p$, so $\E_{\pi^*_n}g=\E_\nu[(1-p)g]/\E_\nu[1-p]$ with $\nu=\nu_{n+1}$, and
\[
\E_{\pi^*_n}g-\E_\nu g=\frac{-\mathrm{Cov}_\nu(p,g)}{1-\E_\nu p},\qquad |\mathrm{Cov}_\nu(p,g)|=\big|\E_\nu[p\,(g-\E_\nu g)]\big|\le\E_\nu p,
\]
since $p\ge0$ and $|g-\E_\nu g|\le1$. For $\varepsilon_n$, $p\,\Bin(n+1\mid N,p)=\frac{n+2}{N+1}\Bin(n+2\mid N+1,p)$, so $\E\sum_xp_x\Bin(n+1\mid N,p_x)=\frac{n+2}{N+1}\E F^{(N+1)}_{n+2}$, while the normaliser of $\nu$ is $\E F_{n+1}$. Finally $(n+2)F^{(N+1)}_{n+2}\le N+1$, the number of rows.
\end{proof}

\begin{proof}[Proof of Corollary~\ref{cor:unseen}]
At $n=0$ the oracle predictive is the mean of $\pi^*_0=\pi^{\lev}_1$ (\cref{prop:shift}), and by \cref{lem:regret} a constant $c$ has regret $\kl(m^*_{0,1}\|c)$ in that stratum, which has weight $w_0$. Given the training counts, the labels of the $F_1$ singletons are, to leading order, independent Bernoulli draws with means $q_x$ averaging $m^*_{0,1}$, so $\hat m^*_{0,1}$ has variance $\sum_xq_x(1-q_x)/F_1^2\le m^*_{0,1}(1-m^*_{0,1})/F_1$, with equality when the singletons' conditionals are equal. A second-order expansion of $\kl$, valid because the one pseudo-observation keeps $\hat m^*_{0,1}$ away from 0 and 1, gives $\E\,\kl(m^*_{0,1}\|\hat m^*_{0,1})\le1/(2F_1)+o(1/F_1)$, and $w_0\approx F_1/N$ by Good--Turing. For the baseline, $\hat\mu-\mu=O_p(N^{-1/2})$ adds $O(w_0/N)$.
\end{proof}

\begin{proof}[Proof of Proposition~\ref{prop:pop-n1}]
The beta $B_{\mu,s}$ predicts $\mu+(1-\mu)t$ after a one and $\mu(1-t)$ after a zero. The map $(\mu,t)\mapsto(\mu+(1-\mu)t,\mu(1-t))$ is injective on $(0,1)^2$ (indeed $t=p(1,1)-p(0,1)$), so both predictions match the oracle iff $(\mu,t)=(\mu^*_1,t^*_1)$. The estimates are \eqref{eq:robbins} with $n=1$.
\end{proof}

\begin{proof}[Proof of Remark~\ref{rem:freq-consistent}]
With levels of size 1 and 2 only, a singleton contributes a Bernoulli likelihood with mean $\mu(\log2)=\operatorname{expit}(a+b\log2)$, and a doubleton a beta-binomial likelihood for $n=2$, which depends only on the mean $\mu(\log3)=\operatorname{expit}(a+b\log3)$ and the correlation $t=1/(1+e^{c})$. The map $(a,b,c)\mapsto(\mu(\log2),\mu(\log3),t)$ is a bijection onto $(0,1)^3$, so the likelihood is that of a saturated model for the singletons' label rate and the doubletons' three-cell table, and its maximiser is the empirical singleton rate and the empirical doubleton mean and correlation (when these lie in the interior). These converge to the mean of $\pi^{\lev}_1$ and the mean and correlation of $\pi^{\lev}_2$ as $F_1,F_2\to\infty$, under the condition of \cref{cor:robbins}. The shifted encoder predicts unseen levels by $\mu(\log2)$, the mean of $\pi^{\lev}_1=\pi^*_0$ up to \cref{lem:train-shift}, and once-seen levels by the beta with mean $\mu(\log3)$ and correlation $t$, which is exact by Proposition~\ref{prop:pop-n1}.
\end{proof}

\begin{proof}[Proof of Corollary~\ref{cor:robbins}]
At $n=0$ the claim is \cref{cor:unseen}, with the singletons' mean within $O(1/\E F_1)$ of $m^*_{0,1}$ by \cref{lem:train-shift}. At $n=1$, by \eqref{eq:robbins} the oracle is a ratio of expected doubleton counts, $p^*(k,1)=(k+1)\E F_{2,k+1}/\{(k+1)\E F_{2,k+1}+(2-k)\E F_{2,k}\}$, and $F_{2,j}/F_2$ converges to $\E F_{2,j}/\E F_2$ as $F_2\to\infty$ (the occupancy counts $F_{2,j}$ concentrate around their means with variance of order $\E F_2$ when no level carries a non-vanishing share of the doubletons, the non-degeneracy condition of the corollary; \citealp{benhamou2017}), so $\hat p^*(k,1)$ converges to the corresponding rate of the training doubletons, which by \cref{lem:train-shift} (with $g(q)=q$ and $g(q)=q^2$, which fix the two rates) is within $O(1/\E F_2)$ of $p^*(k,1)$; the pseudo-pair moves each rate by $O(1/F_2)$ and keeps it inside $(0,1)$, so the per-stratum regret on once-seen levels vanishes by continuity of $\kl$. Levels seen twice or more are predicted as by the baseline, so they do not enter the claim. For the cost: when $q$ does not depend on popularity, the baseline is consistent for the same two predictions (\cref{prop:ml}). Each doubleton rate is a ratio of counts over about $F_2$ pairs, so its error is $O_p(F_2^{-1/2})$. By \cref{lem:regret} the excess risk on once-seen levels is second order in that error, $O(1/F_2)$ per level, with test weight $w_1\approx2F_2/N$ (Good--Turing), which gives $O(w_1/F_2)=O(1/N)$.
\end{proof}

\subsection{Section~\ref{sec:mi}}

\begin{proof}[Proof of Lemma~\ref{lem:rep}]
$\E[n_x\mid p_x]=Np_x$, and $n_x$ is independent of $q_x$ given $p_x$; group the sum by $n_x$.
\end{proof}

\begin{proof}[Proof of Proposition~\ref{prop:pi}]
Take $\pi_1,\pi_2$ attaining (or approaching) the two ends of $\mathcal I_L(m)$. The data have the same law under both, so $\E_1|\hat I-I_1|+\E_2|\hat I-I_2|\ge|I_1-I_2|=I_{\rm hi}-I_{\rm lo}$, and the maximum is at least the average. The second statement is the duality between moment matching and best polynomial approximation \citep[Appendix~E]{wu2016}: the maximum of $\E_{\pi_1}f-\E_{\pi_2}f$ over pairs of laws on $[0,1]$ sharing $L$ moments equals $2E_L(f)$ \citep{karlin1966,wu2016}. The $h(m_1)$ term is common to both priors.
\end{proof}

\begin{proof}[Proof of Proposition~\ref{prop:nonid}]
(i) Which levels are seen, and how often, does not depend on the $q_x$. Let $B$ be the event that some light level is seen at least twice; $P(B)\le M\binom N2(\rho/M)^2\le N^2\rho^2/(2M)$, the same under $G$ and $G'$. On the complement of $B$, each light level that is seen contributes one label, and since the light $q_x$ are i.i.d.\ and independent of the base levels, these labels are i.i.d.\ $\Bern(\mu_G)$ and independent of the base levels' counts. So given $B^{c}$ the fingerprint has the same law under $G$ and $G'$, and the total variation distance is at most $P(B)$.
(ii) $M\cdot N(\rho/M)(1-\rho/M)^{N-1}\to N\rho$.
(iii) $\mu=\E\sum_xp_xq_x$ is the same for all $G$ with mean $\mu_G$, and $H(Y\mid X)=\E\sum_xp_xh(q_x)$ is the base part plus $\rho\,\E_Gh(q)$. Over laws with mean $\mu_G$, $\E_Gh$ takes every value in $[0,h(\mu_G)]$: $h(\mu_G)$ at the point mass (Jensen), $0$ at the two-point law on $\{0,1\}$, and the values in between at their mixtures.
For the minimax statement, take $G$ and $G'$ at the two ends; as in the proof of \cref{prop:pi}, $\E_G|\hat I-I_G|+\E_{G'}|\hat I-I_{G'}|\ge|I_G-I_{G'}|-2\log2\cdot\mathrm{TV}$, and the total variation tends to $0$ by (i).
\end{proof}

\begin{proof}[Proof of Proposition~\ref{prop:lik}]
(i) Expanding $(1-q)^{n-k}$, $P_{\tilde\pi}(k\mid n)=\binom nk\sum_{j=0}^{n-k}\binom{n-k}j(-1)^jm_{k+j}(\tilde\pi)$ is affine in $m_{1:n}(\tilde\pi)$ (\cref{lem:moments}), so $\ell=\sum_{n,k}F_{n,k}\log P_{\tilde\pi}(k\mid n)$ depends on $\tilde\pi$ only through $m_{1:L}(\tilde\pi)$. The likelihood region is therefore the set of priors whose moment vector lies in $\hat M_c$, and the range of $I$ over it is the union over $m'\in\hat M_c$ of the range over priors with moments $m'$, which is $\mathcal I_L(m')$.

(ii) The map $m'\mapsto(P_{m'}(\cdot\mid n))_{n:\omega_n>0}$ is affine and injective, since $P(\cdot\mid L)$ alone gives $m_j=\sum_k\binom kj P(k\mid L)/\binom Lj$ for $j\le L$. Because $\pi((0,1))>0$, every $P_m(k\mid n)$ is positive, so the per-level log-likelihood is smooth near $m$ and the information $J=\sum_n\omega_n\sum_k\nabla P_m(k\mid n)\nabla P_m(k\mid n)^{\top}/P_m(k\mid n)$ is positive definite by injectivity. Given the sizes the levels are independent and $F_n=F\omega_n(1+o(1))$, so the model is locally asymptotically normal at $m$ with information $FJ$ (Lindeberg). The parameter set $\mathcal M_L$ is convex, hence regular in the sense of \citet{chernoff1954} at $m$, with a closed convex tangent cone $T$. In coordinates standardised by $(FJ)^{1/2}$ the likelihood-ratio statistic $2(\max\ell-\ell(m))$ converges in law to $\sup_{d\in T}(2d^{\top}Z-\|d\|^2)=\|\Pi_TZ\|^2$ with $Z\sim N(0,I_L)$ \citep{chernoff1954,self1987}. Projection onto a closed convex cone containing $0$ is non-expansive, so $\|\Pi_TZ\|^2\le\|Z\|^2\sim\chi^2_L$. Since $m\in\hat M_c$ exactly when this statistic is at most $2c$, (i) gives the claim.

(iii) The per-level limit $\bar\ell(m')=\sum_n\omega_n\sum_kP_m(k\mid n)\log P_{m'}(k\mid n)$ is concave and upper semicontinuous on the compact set $\mathcal M_L$ and, by Gibbs' inequality and injectivity, has the unique maximiser $m$. By the law of large numbers and Wald's consistency argument, for every neighbourhood $U$ of $m$ there is $\delta>0$ with $P\big(\sup_{\mathcal M_L\setminus U}\ell/F\le\ell(m)/F-\delta\big)\to1$; as $\ell(m)\le\max\ell$ and $c/F\to0$, $\hat M_c\subseteq U$ with probability tending to one. The set of priors with moments $m'$ is a compact-valued, upper hemicontinuous correspondence in the weak topology, and $\tilde\pi\mapsto h(m_1(\tilde\pi))-\E_{\tilde\pi}h$ is continuous, so by Berge's maximum theorem $I_{\rm hi}$ is upper and $I_{\rm lo}$ lower semicontinuous. Choosing $U$ on which $I_{\rm hi}<I_{\rm hi}(m)+\epsilon$ and $I_{\rm lo}>I_{\rm lo}(m)-\epsilon$ gives the first claim. For the rate, $v(m')=\min\{\E_{\tilde\pi}h:m_{1:L}(\tilde\pi)=m'\}$ is convex on $\mathcal M_L$ (a mixture of minimisers for $m'$ and $m''$ is feasible for the mixture of the moments), $w(m')=\max\{\cdots\}$ is concave, and $m'\mapsto h(m'_1)$ is concave. So $I_{\rm hi}=h(m_1)-v$ is concave and $I_{\rm lo}=(-w)-(-h(m_1))$ is a difference of convex functions; both are locally Lipschitz in the interior of $\mathcal M_L$ and differentiable almost everywhere. When $m$ is interior, local asymptotic normality puts $\hat M_c$ in a ball of radius $O_p(F^{-1/2})$ around $m$, and the Lipschitz property turns this into an excess width of $O_p(F^{-1/2})$.

(iv) $I\in\hat C_c$ exactly when $\sup_{m'\in A}\ell(m')\ge\max\ell-c$, with $A=\{m'\in\mathcal M_L:I_{\rm lo}(m')\le I\le I_{\rm hi}(m')\}\ni m$. If $I$ is strictly inside $\mathcal I_L(m)$, continuity at the interior point $m$ puts a neighbourhood of $m$ in $A$, and the maximiser of $\ell$ lies in it with probability tending to one by (iii) with $c=0$. At an end the argument needs only differentiability, not convexity. Suppose $I=I_{\rm hi}(m)>I_{\rm lo}(m)$. By continuity $I_{\rm lo}<I$ near $m$, so near $m$ the set $A$ is $\{I_{\rm hi}\ge I\}$. Let $g=\nabla I_{\rm hi}(m)\ne0$. Differentiability gives, for every $\epsilon>0$, a $\delta>0$ such that for $|d|<\delta$
\[
g^{\top}d\ge\epsilon|d|\ \Rightarrow\ m+d\in A\ \Rightarrow\ g^{\top}d\ge-\epsilon|d|.
\]
So the rescaled sets $\sqrt F(A-m)$ are squeezed between the cones $\{g^{\top}d\ge\pm\epsilon|d|\}$ on balls of growing radius, and they converge to the half-space $H=\{d:g^{\top}d\ge0\}$. This approximation of $A$ by a cone at $m$ is the regularity condition of \citet{chernoff1954}; convexity is not needed. As in (ii), with $T=\mathbb R^L$ since $m$ is interior, $2(\max\ell-\sup_A\ell)$ converges in law to $\|Z\|^2-\|\Pi_HZ\|^2=\min(0,u^{\top}Z)^2$, with $u$ the unit vector along $(FJ)^{-1/2}g$. This variable is $0$ with probability $\tfrac12$ and $\chi^2_1$ otherwise, so its probability of being at most $\chi^2_{1,1-\alpha}$ is $1-\alpha/2$. The lower end is the same argument applied to $-I_{\rm lo}$: if $I=I_{\rm lo}(m)<I_{\rm hi}(m)$ and $I_{\rm lo}$ is differentiable at $m$ with $g=\nabla I_{\rm lo}(m)\ne0$, then near $m$ the set $A$ is $\{I_{\rm lo}\le I\}$, which is squeezed in the same way between $\{g^{\top}d\le\pm\epsilon|d|\}$. It need not be convex, since $I_{\rm lo}$ is only a difference of convex functions, but its tangent cone is the half-space $\{g^{\top}d\le0\}$ and the limit is the same.

At a kink the cone changes. For the concave $I_{\rm hi}$ the set $\{I_{\rm hi}\ge I\}$ is convex and its tangent cone $K$ is a closed convex cone strictly inside a half-space, so the limit is $\|Z\|^2-\|\Pi_KZ\|^2=\operatorname{dist}(Z,K)^2$, which can exceed the half-space limit. For a quadrant in two dimensions, $P(\operatorname{dist}(Z,K)^2\le\chi^2_{1,0.95})=\tfrac14+\tfrac12(0.95)+\tfrac14P(\chi^2_2\le3.84)=0.938$, below the nominal $0.95$. At a kink of $I_{\rm lo}$ the set $\{I_{\rm lo}\le I\}$ can contain more than a half-space near $m$, and the coverage is then at least $1-\alpha/2$.
\end{proof}

\section{Further results on moment exactness}\label{app:exact-more}

This appendix collects remarks on \cref{sec:exact} that the main text only points to.

\begin{remark}[More parameters, more exact levels]\label{rem:more}
Nothing here is specific to the beta. Any family that can match $m_1,\dots,m_J$ of $\pi$ yields an encoder exact for $n\le J-1$. A $J$-atom discrete prior can match $2J-1$ moments (Gauss quadrature), so a nonparametric prior such as the NPMLE can extend exactness to larger $n$. This is where its gains appear empirically (\cref{sec:synthetic,sec:bakeoff}).
\end{remark}

\begin{remark}[Rates and the low-signal boundary]\label{rem:rates}
Standard M-estimation gives $\hat t-t_\pi=O_p(F_2^{-1/2})$, so the mismatch in $m_2$ is $O_p(F_2^{-1/2})$ and, by Proposition~\ref{prop:expansion} with $\delta_2\neq0$, the regret on $n=1$ levels is $O_p(1/F_2)$. The fitted encoder also centres the beta on $\hat\mu$, whose error is $O_p(N^{-1/2})$, so $\delta_1\neq0$ as well and the regret on levels seen at most once is $O_p(1/F_2+1/N)$. When $t_\pi$ is close to $0$ the constrained maximiser often sits at $\hat t=0$, i.e.\ $\hat s=\infty$ and the encoder returns $\hat\mu$ (with probability tending to $1/2$ in the boundary case $t_\pi=0$, which the standing assumption $v>0$ excludes; \citealp{chernoff1954,self1987}). Near $t=0$ the marginal predictor's regret is about $2I(X;Y)^2$ against $O(1/F_2)$ for the fitted beta, so the baseline beats the marginal only once $I(X;Y)$ is large compared with $F_2^{-1/2}$. This is visible at $f=0.9$, $N=500$ in the junk-fraction sweep and at $\bar n\le0.5$ in the MovieLens oracle.
\end{remark}

\begin{remark}[Large $n$]\label{rem:largen}
The expansion is informative only while the mismatches are small relative to the moments. For large $n$ the behaviour depends on $\pi$. If $\pi$ has a density, both predictives are within $O(1/n)$ of $k/n$ and the regret is governed by the density's exponents at $0$ and $1$: $\Delta_n=O(n^{-2}\log n)$ for a density positive at both ends, $O(n^{-2})$ if it vanishes there, and $O(n^{-1-\alpha})$ for a U-shaped density $\propto q^{\alpha-1}$ near $0$ with $\alpha<1$ (\cref{prop:largen-smooth}). If $\pi$ is atomic, the oracle risk decays exponentially while any beta predictor keeps the parametric rate $R_n\approx1/(2n)$, so $n\Delta_n\to1/2$ (\cref{prop:largen-atomic}). This is the regime where the NPMLE pays off (Appendix~\ref{app:gap}).
\end{remark}

\begin{remark}[Regret and transport distance]\label{rem:w1}
Learning a population of parameters is usually measured in Wasserstein distance \citep{tian2017,vinayak2019}, which controls the regret but not conversely. Write $\Delta^\pi_n(p_{\tilde\pi})$ for the regret on levels of size $n$ of the oracle built for $\tilde\pi$ when the prior is $\pi$. Then for every $n$
\[
\Delta^\pi_n(p_{\tilde\pi})\ \le\ (n+1)\inf_{\gamma\in\Gamma(\pi,\tilde\pi)}\E_\gamma\,\kl(q\,\|\,\tilde q)\ \le\ \frac{(n+1)\,W_2^2(\pi,\tilde\pi)}{\epsilon(1-\epsilon)}\ \le\ \frac{(n+1)\,W_1(\pi,\tilde\pi)}{\epsilon(1-\epsilon)},
\]
where $\Gamma(\pi,\tilde\pi)$ is the set of couplings of $q\sim\pi$ and $\tilde q\sim\tilde\pi$ \citep{villani2009}, and the last two bounds assume that $\tilde\pi$ is supported in $[\epsilon,1-\epsilon]$ (proof in Appendix~\ref{app:proofs}). So a prior within $W_1$-distance $\delta$ of the truth loses at most $O(n\delta)$ nats on levels of size $n$. The converse fails: by \cref{lem:moments} the regret on levels of size $n$ depends only on $m_1,\dots,m_{n+1}$, so a prior can be far from $\pi$ in $W_1$ and still be exact there, as the moment-matched beta is at $n\le1$ for a three-atom $\pi$. So the $W_1$ floor of order $1/T$ that \citet{tian2017} prove for $T$ observations per unit sits in moments the regret on levels seen fewer than $T$ times never sees.
\end{remark}

\paragraph{Numerical check.}\label{sec:numcheck}

All quantities below are exact sums over binomial outcomes, with no Monte Carlo. The vanishing of the regret at $n\le1$ ($n\le2$ for the symmetric prior) holds to machine precision, and the second-order formula of Proposition~\ref{prop:expansion} matches the exact regret closely when the mismatch is small (Table~\ref{tab:regret} in Appendix~\ref{app:numcheck}). Table~\ref{tab:avg} averages over level sizes, with all $p_x$ equal so that $n_x$ is Poisson with mean $\bar n$ and $w_n$ is the Poisson pmf. ``Pseudo-true'' is the population limit of the type-II ML fit with $\mu=m_1$ fixed, over the same design. A Monte Carlo check of Proposition~\ref{prop:ml} on the first prior ($t_\pi=0.32$), with all levels of size 2, gives $\hat t=0.220,0.335,0.311,0.320$ for $F_2=2\cdot10^2,2\cdot10^3,2\cdot10^4,2\cdot10^5$. On the true distributions of $q$ of the six real-data pools of \cref{sec:oracle} (row-weighted, $\mu$ from $0.03$ to $0.63$, none symmetric), the moment-matched beta's regret is at most $0.9$ milli-nats at $n=2$, under $1\%$ of $I(X;Y)$ on every pool, against up to $1.6$ at $n=3$ and $2.2$ at $n=5$: the third-moment mismatch is not zero, but it enters squared.

\section{Symmetric conditionals under popularity dependence}\label{app:sym}

This appendix completes \cref{sec:popularity} for conditionals symmetric about $1/2$, the case of label-randomised designs (\cref{cor:sym}).

\begin{table}[h]
\centering\small
\begin{tabular}{llll}
\toprule
Test level seen & Oracle needs & Read off from & Beta encoder exact iff \\
\midrule
0 times & mean of $\pi^*_0$ & singletons & $\mu_0=\mu^*_0$\\
1 time & mean, variance of $\pi^*_1$ & doubletons & $(\mu_1,t_1)=(\mu^*_1,t^*_1)$\\
2 times & 3 moments of $\pi^*_2$ & tripletons & $(\mu_2,t_2)=(\mu^*_2,t^*_2)$ and $\kappa_3$ matches\\
\bottomrule
\end{tabular}
\caption{The popularity-aware oracle for levels seen at most twice.}\label{tab:pop-n2}
\end{table}

If every $\pi^*_n$ is symmetric about $1/2$ (for example, $q_x\mid p_x$ symmetric for every $p_x$, as in a junk-fraction prior whose junk share depends on popularity), then $\mu^*_n=1/2$, $\kappa_3(\pi^*_n)=0$, and each stratum is described by one number, $t^*_n=4\,\mathrm{Var}_{\pi^*_n}(q)$.

\begin{proposition}[Symmetric case]\label{prop:pop-sym}
If every $\pi^*_n$ is symmetric about $1/2$, the oracle for $n\le2$ is
\begin{center}\small
\begin{tabular}{lccc}
\toprule
& $k=0$ & $k=1$ & $k=2$\\
\midrule
$n=0$ & $1/2$ & & \\
$n=1$ & $(1-t^*_1)/2$ & $(1+t^*_1)/2$ & \\
$n=2$ & $\dfrac{1-t^*_2}{2(1+t^*_2)}$ & $1/2$ & $\dfrac{1+3t^*_2}{2(1+t^*_2)}$\\
\bottomrule
\end{tabular}
\end{center}
and a beta with mean $1/2$ is exact on all levels seen at most twice if and only if its correlation is $t^*_1$ on once-seen and $t^*_2$ on twice-seen levels. If moreover every training level is seen at most twice, the blind baseline converges to $t^*_1$ and is exact on unseen and once-seen levels, despite the popularity dependence.
\end{proposition}

Under symmetry the correlations are ``agreeing minus disagreeing pairs'': $\hat t^*_1=(F_{2,0}+F_{2,2}-F_{2,1})/F_2$ from doubletons and $\hat t^*_2=(F_{3,0}+F_{3,3}-\tfrac13(F_{3,1}+F_{3,2}))/F_3$ from tripletons. In a simulated junk mixture whose junk share falls with popularity ($q_0=0.15$, $3\times10^5$ levels and rows, 10 repetitions), these estimates give $t^*_1=0.259$ and $t^*_2=0.314$ against the test-weighted truths $0.260$ and $0.320$. The correlation among singletons ($0.184$, the wrong stratum for $n=1$) and the blind pooled fit ($0.347$) are both off. Per test row on once-seen levels, the blind baseline's regret is $4.5\times10^{-3}$ nats and the shifted per-stratum encoder's is $1.4\times10^{-5}$.

\begin{proof}[Proof of Proposition~\ref{prop:pop-sym}]
$n=0$: the mean is $1/2$. $n=1$: Proposition~\ref{prop:pop-n1} with $\mu^*_1=1/2$. $n=2$: symmetry gives $\E(q-\tfrac12)^3=0$, i.e.\ $m_3=\tfrac32m_2-\tfrac14$; with $m_2=(1+t)/4$ this is $m_3=(1+3t)/8$, so $p^*(2,2)=m_3/m_2=(1+3t)/(2(1+t))$. For $k=1$, $M(2,1)=M(1,2)$ by symmetry and $M(1,1)=M(2,1)+M(1,2)$, so $p^*(1,2)=1/2$. The beta $\BetaD(s/2,s/2)$ gives $(k+s/2)/(2+s)$, which with $s=1/t-1$ reproduces the table; both third central moments vanish, so the condition of Proposition~\ref{prop:n2} holds automatically. For the last claim: when all training levels have size at most 2, the blind likelihood is driven by doubletons only (Proposition~\ref{prop:ml}, part 1), which by Proposition~\ref{prop:shift} are draws from $\pi^*_1$; part 3 of Proposition~\ref{prop:ml} then gives $\hat t\to t^*_1$.
\end{proof}

\section{Numerical check of Section~\ref{sec:exact}}\label{app:numcheck}

\Cref{tab:regret} gives the per-level regret of the moment-matched beta for three discrete priors, computed exactly, next to the second-order formula of Appendix~\ref{app:gap}; \cref{tab:avg} averages it over Poisson level sizes and compares the moment-matched beta with the pseudo-true limit of the type-II ML fit (Remark~\ref{rem:mixed}).

\begin{table}[!ht]
\centering\small
\begin{tabular}{lcccccc}
\toprule
Prior $\pi$ (atoms; weights) & $\kappa_3(\pi)$ & $\kappa_3(B^\pi)$ & $\Delta_1$ & $\Delta_2$ & $\Delta_3$ & $\Delta_5$ \\
\midrule
$\{.1,.5,.9\}$; $\{\frac14,\frac12,\frac14\}$ & 0 & 0 & $<10^{-16}$ & $<10^{-16}$ & $2.0\!\times\!10^{-4}$ & $9.8\!\times\!10^{-4}$\\
\quad second-order formula & & & 0 & 0 & $2.0\!\times\!10^{-4}$ & $9.3\!\times\!10^{-4}$\\
$\{.2,.3,.85\}$; $\{.3,.4,.3\}$ & 0.0166 & 0.0046 & $<10^{-16}$ & $5.7\!\times\!10^{-3}$ & $1.0\!\times\!10^{-2}$ & $1.8\!\times\!10^{-2}$\\
\quad second-order formula & & & 0 & $5.9\!\times\!10^{-3}$ & $9.2\!\times\!10^{-3}$ & $2.1\!\times\!10^{-2}$\\
$\{.4,.45,.6,.7\}$; $\{.2,.3,.3,.2\}$ & 0.0003 & $-0.0001$ & 0 & $4.9\!\times\!10^{-6}$ & $1.7\!\times\!10^{-5}$ & $6.0\!\times\!10^{-5}$\\
\quad second-order formula & & & 0 & $4.9\!\times\!10^{-6}$ & $1.7\!\times\!10^{-5}$ & $6.2\!\times\!10^{-5}$\\
\bottomrule
\end{tabular}
\caption{Per-level regret $\Delta_n$ (nats) of the moment-matched beta against the true-prior oracle.}\label{tab:regret}
\end{table}

\begin{table}[!ht]
\centering\small
\begin{tabular}{lccccccc}
\toprule
Prior $\pi$ & $\bar n$ & $s_\pi$ & pseudo-true $s$ & oracle eCE & gap, matched & gap, pseudo-true & of which $n\le1$\\
\midrule
$\{.1,.5,.9\}$; $\{\frac14,\frac12,\frac14\}$ & 1 & 2.125 & 2.123 & 0.1417 & 0.00002 & 0.00002 & $<10^{-6}$\\
($I=0.184$) & 2 & & 2.119 & 0.1131 & 0.00015 & 0.00015 & $<10^{-6}$\\
 & 5 & & 2.102 & 0.0660 & 0.00169 & 0.00168 & $<10^{-6}$\\
 & 10 & & 2.088 & 0.0325 & 0.00672 & 0.00672 & $<10^{-6}$\\
$\{.1,.5,.9\}$; $\{.4,.2,.4\}$ & 1 & 0.953 & 0.956 & 0.1911 & 0.00034 & 0.00034 & $<10^{-6}$\\
($I=0.295$) & 5 & & 1.014 & 0.0591 & 0.00696 & 0.00661 & $5\times10^{-6}$\\
 & 10 & & 1.114 & 0.0256 & 0.01323 & 0.01193 & $<10^{-6}$\\
$\{.05,.95\}$; $\{\frac12,\frac12\}$ & 1 & 0.235 & 0.237 & 0.2354 & 0.00226 & 0.00225 & $2\times10^{-6}$\\
($I=0.495$) & 5 & & 0.288 & 0.0162 & 0.03078 & 0.02821 & $4.9\times10^{-5}$\\
 & 10 & & 0.375 & 0.0007 & 0.03750 & 0.02962 & $4\times10^{-6}$\\
$\{.2,.3,.85\}$; $\{.3,.4,.3\}$ & 1 & 2.254 & 2.271 & 0.1229 & 0.00196 & 0.00196 & $1\times10^{-6}$\\
($I=0.163$) & 5 & & 2.389 & 0.0455 & 0.01554 & 0.01534 & $3\times10^{-6}$\\
 & 10 & & 2.532 & 0.0171 & 0.02079 & 0.02029 & $<10^{-6}$\\
\bottomrule
\end{tabular}
\caption{Total excess cross-entropy (nats) of the oracle and gap of the beta encoder, averaged over Poisson($\bar n$) level sizes. ``Of which $n\le1$'': the part of the pseudo-true fit's gap on levels seen at most once, where the moment-matched beta has none. The type-II ML fit with larger levels present loses at most $5\times10^{-5}$ nats there, under $0.3\%$ of its total gap (also for an asymmetric two-atom prior $\{.1,.9\}$, $\{.8,.2\}$; script \texttt{pseudo\_true.py}). Levels are equally popular, so level sizes are Poisson; \cref{tab:heavy} has heavy-tailed sizes.}\label{tab:avg}
\end{table}

\begin{table}[!ht]
\centering\small
\begin{tabular}{lcccccc}
\toprule
Prior $\pi$ & $s_\pi$ & design & pseudo-true $s$ & $\Delta_1$ & loss on $n\le1$ & share of $I$\\
\midrule
$\{.1,.5,.9\}$; $\{\frac14,\frac12,\frac14\}$ & 2.125 & 1/2/200 & 2.281 & $1.3\times10^{-4}$ & -- & --\\
 & & lognormal, $\bar n=1$ / 2 / 5 & 2.113 / 2.115 / 2.124 & $\le9\times10^{-7}$ & $\le1.3\times10^{-7}$ & $<0.01\%$\\
$\{.1,.5,.9\}$; $\{.4,.2,.4\}$ & 0.953 & 1/2/200 & 1.371 & $5.1\times10^{-3}$ & -- & --\\
 & & lognormal, $\bar n=1$ / 2 / 5 & 1.028 / 1.061 / 1.118 & $2.4$--$10\times10^{-4}$ & $3.6$--$4.9\times10^{-5}$ & $0.02\%$\\
$\{.05,.95\}$; $\{\frac12,\frac12\}$ & 0.235 & 1/2/200 & 0.619 & $3.5\times10^{-2}$ & -- & --\\
 & & lognormal, $\bar n=1$ / 2 / 5 & 0.294 / 0.322 / 0.373 & $1.8$--$7.8\times10^{-3}$ & $2.7$--$3.7\times10^{-4}$ & $0.07\%$\\
$\{.2,.3,.85\}$; $\{.3,.4,.3\}$ & 2.254 & 1/2/200 & 2.811 & $1.1\times10^{-3}$ & -- & --\\
 & & lognormal, $\bar n=1$ / 2 / 5 & 2.407 / 2.455 / 2.529 & $1.0$--$3.1\times10^{-4}$ & $1.5$--$1.8\times10^{-5}$ & $0.01\%$\\
\bottomrule
\end{tabular}
\caption{Pseudo-true type-II ML fit under heavy-tailed level sizes. ``1/2/200'': $60\%$, $30\%$ and $10\%$ of the levels seen 1, 2 and 200 times. ``Lognormal'': popularities lognormal with $\sigma=1.5$ and Poisson counts. $\Delta_1$: per-level regret (nats) of the pseudo-true beta at $n=1$, which is zero for the moment-matched beta ($\Delta_0=0$ for both). ``Loss on $n\le1$'': $w_0\Delta_0+w_1\Delta_1$ per test row, where $w_0+w_1$ is $0.34$, $0.21$ and $0.09$ at $\bar n=1,2,5$. Script \texttt{paper/theory/pseudo\_true\_heavy.py}.}\label{tab:heavy}
\end{table}

\begin{table}[!ht]
\centering\small
\begin{tabular}{lcccc}
\toprule
pool & $I(X;Y)$ & identified interval, $L=2$ & half-width / $I$ & beta regret at $n=2$ / $I$\\
\midrule
MovieLens movies & 0.0871 & $[0.0831, 0.1164]$ & $19\%$ & $0.26\%$\\
Book-Crossing users, rated & 0.1408 & $[0.1199, 0.1819]$ & $22\%$ & $0.65\%$\\
Avazu \texttt{device\_ip} & 0.0400 & $[0.0228, 0.0523]$ & $37\%$ & $0.22\%$\\
\bottomrule
\end{tabular}
\caption{The numbers behind \cref{fig:teaser}(b) at $L=n=2$, on each pool's row-weighted (popularity-blind) law of $q$ with infinitely many levels (nats). The intervals are not centred on $I$. Script \texttt{paper/fig\_teaser.py}.}\label{tab:teaser}
\end{table}

\section{More on the likelihood interval}\label{app:lik-more}

So the cutoff that guarantees coverage by projection is $\chi^2_L/2$, with $L$ the number of moments the table identifies. After thinning $L$ can be 200, and $\chi^2_{200,0.95}/2\approx117$ against $\chi^2_{1,0.95}/2\approx1.9$, so the projection interval is far wider. The $\chi^2_1$ cutoff that we use is the profile-likelihood cutoff for the scalar $I$. At an interior moment vector where the ends are differentiable, (iv) says it is asymptotically valid, even conservative: when $I$ is set only partially, the truth is either inside the identified set, where the profile likelihood is flat, or at one end, where only one side of the cutoff binds. At a kink of an end it can under-cover (a quadrant-shaped cone gives $0.938$ at nominal $0.95$; Appendix~\ref{app:proofs}). This is the observation of \citet{imbens2004}, and the likelihood interval is an instance of the profile quasi-likelihood-ratio set for a scalar function of a partially identified parameter (Procedure~3 of \citealp{chen2018}), which covers the whole identified set with asymptotic probability at least $1-\alpha$ under their conditions. What \cref{prop:lik} adds is that in this model the conditions can be checked directly: the likelihood depends on the prior only through $m_{1:L}$, is concave in it, and is locally asymptotically normal at an interior $m$. Two limits remain. First, (iv) is pointwise. Intervals aimed at the ends of the identified set can under-cover uniformly when the set is short compared with the sampling error \citep{stoye2009}. For the two-sided $\chi^2_1$ cutoff this does not happen when the gradients of $I_{\rm lo}$ and $I_{\rm hi}$ become parallel, since the acceptance region then contains a slab through $m$ and the profile statistic is dominated by a $\chi^2_1$ variable; we have not proved uniform validity in general. Second, (iv) excludes moment vectors on the boundary of $\mathcal M_L$, which include every prior with $r$ atoms in $(0,1)$ once $L\ge2r$, and so the atomic priors of our experiments. There the identified set is a point, and the profile statistic converges to a chi-bar-squared variable whose weights depend on the local cones of $\mathcal M_L$ \citep{chernoff1954,self1987}; it need not be dominated by $\chi^2_1$. \Cref{sec:mi-exp} checks the coverage of all the intervals at finite $F$; for the atomic priors that check is the only evidence.

\section{Mutual information and estimating the unseen}\label{sec:mi-vv}

\paragraph{What is the same.} \citet{valiant2017} estimate a symmetric property $\sum_xf(p_x)$ of the distribution of $X$ from its fingerprint $(F_n)$. Each $\E F_n$ is linear in the histogram of the $p_x$ through Poisson kernels; they solve an LP for the histograms consistent with the observed fingerprint, and prove optimality by moment matching tied to polynomial approximation. Our problem has this structure with the binomial kernel in place of the Poisson kernel and the table $F_{n,k}$ in place of $F_n$: the unknown is the mixing law $\pi$ of the $q_x$, the target is the linear functional $\E_\pi h$, the identified set is a moment LP, and the lower bound is the duality in Proposition~\ref{prop:pi}. The closest work is \citet{tian2017}, binomial observations of many units with unknown parameters, and its NPMLE counterpart \citep{vinayak2019}.

\paragraph{What is different.}
\begin{itemize}
\item \textbf{A different layer.} Valiant and Valiant work on the popularity layer, and $H(X)$ depends on the unseen support. Under Assumption~\ref{ass:exch}, $I(X;Y)$ depends only on the $q$ layer, and unseen levels drop out (Lemma~\ref{lem:rep}). The support size of $X$ never enters.
\item \textbf{Generic MI estimators pay for the wrong layer.} Writing $I=H(X)+H(Y)-H(X,Y)$ and using minimax entropy estimators \citep{jiao2015} requires $N\gtrsim\nlev/\log\nlev$ for both $H(X)$ and $H(X,Y)$. When most levels are unseen this fails, even though $I$ may be nearly identified from doubletons.
\item \textbf{The per-unit sample size is random and mostly tiny.} In \citet{tian2017} every unit has the same number of observations; here the profile $(F_n)$ decides which moments are available and how precisely.
\item \textbf{An exchangeability assumption is essential.} The unseen estimators are assumption-free on the histogram. Here, without Assumption~\ref{ass:exch} or a model linking strata, the singleton share of $H(Y\mid X)$ is unidentified (\cref{sec:mi-pop}).
\end{itemize}

\begin{table}[h]
\centering\small
\begin{tabular}{lccccccc}
\toprule
$L$ & 1 & 2, 3 & 4, 5 & 6, 7 & 8, 9 & 10, 11 & 12\\
$E_L(h)$ (nats) & 0.3466 & 0.0378 & 0.0123 & 0.0059 & 0.0034 & 0.0022 & 0.0015\\
\bottomrule
\end{tabular}
\caption{Best uniform polynomial approximation error of the binary entropy on $[0,1]$: the worst-case half-width of the identified interval for $I(X;Y)$ when levels have at most $L$ observations.}\label{tab:EL}
\end{table}

A heuristic sample-size requirement follows. With $\nlev$ equally popular levels and $\lambda=N/\nlev$, $F_{\ge L}\approx\nlev\lambda^L/L!$; if $N=\nlev^{1-\delta}$, the largest usable $L$ is about $1/\delta$ and the worst-case floor about $0.22\,\delta^2$ nats. So worst-case consistency needs $N=\nlev^{1-o(1)}$, nearly linear in $\nlev$ like the $\nlev/\log\nlev$ threshold for $H(X)$, for a different reason. For favourable priors the floor is far below the worst case (Table~\ref{tab:id}), and heavy-tailed popularity supplies large levels at small $N$. A finite-sample minimax rate in terms of $(F_n)$ is open.

\section{The moment interval and point estimates for the mutual information}\label{app:mi-point}

\paragraph{The moment interval.} \texttt{\pkgname{}.mi\_interval} takes the confidence region for $\hat m_{1:L}$ from the joint covariance of the pooled U-statistics, which is far from diagonal because every level with $n_x\ge L$ contributes to all of them. The region is a box in the principal axes of that covariance, $|v_i^\top(m-\hat m)|\le c\sqrt{\lambda_i}$, with $c$ the \v{S}id\'ak quantile that gives joint $95\%$ coverage for Gaussian estimates ($c=2.49$ at $L=4$). It is a polytope, so both programmes stay linear. The implementation scans $m_1$ on a grid, so that $h(m_1)$ and $\E h$ come from the same prior. The useful $L$ is the largest level size with many levels: the coefficients of high-degree approximations grow quickly and $\hat m_L$ rests on only $F_{\ge L}$ levels, while the bias $E_L(h)$ falls like $L^{-2}$. In the coverage study we use the largest $L\le4$ for which at least 100 levels are seen $L$ times. A per-moment box of $\pm2$ standard errors, which ignores the correlation between the moments, has the same coverage and is up to twice as wide (median ratio $0.50$ at $\bar n=20$).

\paragraph{Coverage and width.} In our studies the moment interval covers when Assumption~\ref{ass:exch} holds: at least $0.99$ in every synthetic cell and at least $0.90$ on every shuffled pool except Criteo. It is wide: $0.025$--$0.32$ nats in the synthetic cells and up to $0.69$ nats on the 495-level Book-Crossing $r\ge8$ pool at $\bar n=0.5$, often more than $I$ itself. Where few levels repeat, identification dominates that width: on the Book-Crossing $r\ge8$ pool $L$ is 1 or 2 at $\bar n\le1$, and the identified interval is $0.66$ of the $0.69$-nat width at $\bar n=0.5$ and $0.27$ of $0.57$ at $\bar n=1$. It is no more robust to popularity dependence than the likelihood interval: on real MovieLens movies its coverage is $0.60$ at $\bar n=1$ and $2$ although it is four times wider than the likelihood interval, which covers there, and every miss is an interval entirely above $I$. This is consistent with its higher moments resting only on the levels seen at least $j$ times, the popular and more polarized movies, while the likelihood fits one prior to all cells, singletons included; the likelihood interval meets the same problem only once its width has shrunk, at $\bar n\ge10$ (\cref{sec:mi-exp}). It covered the large-$\bar n$ cells where the likelihood interval fails ($0.92$--$1.00$) only because it is several times wider. Its midpoint is a poor point estimate (\cref{tab:mi-point}).

\Cref{tab:mi-coverage-full} repeats \cref{tab:mi-coverage} with the moment interval added. At $\bar n=5$ the likelihood interval is $5$--$9$ times narrower than the moment interval on the synthetic priors and $4$--$6$ times on the oracle pools; at $\bar n=1$ the ratios are $3$--$5$ and $2$--$4$. The exception is Criteo, where the moment interval is narrow and under-covers.

\begin{table}[h]
\centering\small
\resizebox{\linewidth}{!}{\input{tables/tab_mi_coverage_full}}
\caption{As \cref{tab:mi-coverage}, with the moment interval (``mom.'').}\label{tab:mi-coverage-full}
\end{table}

\Cref{tab:mi-point} gives the bias and RMSE of five point estimates of $I(X;Y)$ in the coverage study of \cref{sec:mi-exp}, on the populations of \cref{tab:mi-coverage}: the plug-in $h(\hat\mu)-\sum_x(n_x/N)h(k_x/n_x)$, the baseline (the beta estimator of \citet{hernandez2019}), the NPMLE, the frequency-aware prior and the midpoint of the moment interval. The plug-in is off by $0.2$--$0.4$ nats at $\bar n=1$ and still by about $0.1$ nats at $\bar n=5$, except on Criteo, where $I$ is small. Among the others, the NPMLE has the smallest bias at $\bar n=5$; at $\bar n=1$ it removes the baseline's bias on atomic priors and Avazu but not elsewhere, and its RMSE on the real pools is no smaller.

\begin{table}[h]
\centering\small
\resizebox{\linewidth}{!}{\input{tables/tab_mi_point}}
\caption{Bias of the point estimates of $I(X;Y)$, in milli-nats, with the RMSE in brackets. Same populations and data sets as \cref{tab:mi-coverage}: 100 data sets per synthetic cell and 50 per oracle pool. ``freq.'': frequency-aware prior; ``midpoint'': centre of the moment interval.}\label{tab:mi-point}
\end{table}

\paragraph{Classical estimators.} \Cref{tab:mi-classical} gives, on the same data sets, three classical bias-corrected estimators applied to $I=H(X)+H(Y)-H(X,Y)$: Miller--Madow \citep{miller1955,panzeri1996}, Chao--Shen \citep{chao2003} and NSB \citep{nemenman2002} (\texttt{\pkgname{}.mi\_baselines}). NSB is given the true number of levels $|\mathcal X|$, unseen ones included, which the other estimators do not know. None of them models the conditionals $q_x$. Miller--Madow removes only a small part of the plug-in's bias ($+0.16$ to $+0.33$ nats at $\bar n=1$, except on Criteo). Chao--Shen is off by $0.08$--$0.16$ nats at $\bar n=1$ and by up to $0.14$ nats at $\bar n=5$. NSB is the best of the three: $-63$ to $-13$ milli-nats on the synthetic populations and $+9$ to $+56$ on the oracle pools at $\bar n=1$, and within $\pm18$ at $\bar n=5$. On the synthetic populations the baseline and the NPMLE are within $\pm11$ milli-nats at $\bar n=1$ (\cref{tab:mi-point}), several times closer than NSB; on the real pools at $\bar n=1$ the three are of similar size (baseline $-20$ to $+27$, NPMLE $-1$ to $+33$), and at $\bar n=5$ NSB is as accurate as the NPMLE on most pools.

\begin{table}[h]
\centering\small
\resizebox{\linewidth}{!}{\input{tables/tab_mi_classical}}
\caption{Bias of classical estimators of $I(X;Y)$, in milli-nats, with the RMSE in brackets, on the data sets of \cref{tab:mi-point}. M--M: Miller--Madow; C--S: Chao--Shen; NSB: posterior mean of \citet{nemenman2002} given the true number of levels.}\label{tab:mi-classical}
\end{table}

\section{Implementation checks of the likelihood interval}\label{app:mi-sens}

The likelihood interval of \cref{sec:mi-exp} restricts the prior to a 101-point grid on $[0,1]$ and thins every level above 200 rows to 200 by a hypergeometric draw. \Cref{tab:mi-sens} checks both choices on the same data sets.
\begin{itemize}
\item \emph{Grid.} With 401 grid points the ends of the interval move by at most $3\times10^{-5}$ nats, and coverage does not change, on the three-atom prior and on $\mathrm{Beta}(1,3)$ at $\bar n=1$ and $5$. The finer grid takes three to six times longer.
\item \emph{Atoms off the grid.} The atoms of the three-atom prior ($0.2$, $0.3$, $0.85$) lie on the grid. Moving them to $0.2037$, $0.3061$, $0.8529$ leaves coverage at $0.98$ for $\bar n=1$ and changes it from $0.94$ to $0.96$ for $\bar n=5$, within one binomial standard error, with the same widths.
\item \emph{Thinning.} On shuffled Avazu device IPs, where the largest level has a median of 357 rows at $\bar n=5$ and 1{,}415 at $\bar n=20$, a cap of 1000 rows instead of 200 moves the ends by at most $4\times10^{-4}$ nats and the mean width not at all; coverage at $\bar n=20$ goes from $0.93$ to $0.90$, a difference of one data set in 30. On MovieLens movies no level exceeds 200 rows at $\bar n\le20$, so the cap never acts.
\item \emph{Lognormal popularity.} The likelihood interval is not reported for the $10^5$ lognormal-popularity levels at $\bar n\ge2$: a single data set took more than ten minutes, against seconds for the other cells, so these cells were not run.
\end{itemize}

\begin{table}[h]
\centering\small
\resizebox{\linewidth}{!}{\input{tables/tab_mi_sens}}
\caption{Implementation checks of the 95\% likelihood interval: coverage of $I$ and mean width under setting A / setting B, and the largest change of either end over the data sets (paired: the same data sets under both settings). The off-grid row uses different populations, so no change of ends is given.}\label{tab:mi-sens}
\end{table}

\section{Exchangeability tests}\label{app:exch}

\paragraph{Construction of the stratum moment test.} The test groups levels into strata of sizes $1,2,3,4,5$--$6,7$--$10,11$--$20,21$--$50,51$--$100$ and above $100$, drops strata with fewer than 30 levels, and computes the two averages in each (only $m_1$ for singletons). Within a stratum the per-level estimates are independent with a common mean under the null, given the level sizes, so their sample covariance over the number of levels estimates the covariance of the stratum's averages, and strata are independent. The statistic is the generalised-least-squares residual $\sum_s(\hat\theta_s-\bar\theta)^{\top}\hat\Sigma_s^{-1}(\hat\theta_s-\bar\theta)$ around the pooled estimate $\bar\theta$, referred to $\chi^2$ with the number of stratum moments minus two degrees of freedom. 

\Cref{tab:exch-pools} gives the rejection rates of the stratum moment test and of the frequency-aware prior's likelihood-ratio test (\cref{sec:mi-pop}) on the training sets of the oracle pools, next to the coverage of the likelihood interval from \cref{tab:mi-coverage}; \cref{tab:exch-tasks} gives both tests on the full data of the 24 bakeoff tasks. For the four-class \texttt{employee\_salaries} target, each test is run one class against the rest and the smallest $p$-value is Bonferroni-corrected. The stratum test cannot run on \texttt{churn/state}, where only one stratum has 30 levels. All values of $\bar n$ are in the supplementary CSV files.

\begin{itemize}
\item \emph{Together the two tests flag most coverage failures.} In nine of the ten pool cells where the likelihood interval covers less than $0.85$, at least one test rejects at level $0.05$ in $56$--$100\%$ of the training sets. The likelihood-ratio test catches the MovieLens and Book-Crossing rated failures, and the stratum test the Avazu and Criteo ones, where the dependence on popularity is not a log-linear drift of the mean. The one miss is Book-Crossing $r\ge8$ at $\bar n=20$ (coverage $0.78$, no rejections); its shuffled pool also covers only $0.90$, so popularity is not the cause there.
\item \emph{They also raise alarms that do not matter.} On MovieLens users both tests reject from $\bar n=2$ on in almost every training set, while the likelihood interval keeps covering ($0.92$--$0.96$), and on movies they reject from $\bar n=1$, while coverage holds up to $\bar n=5$. The tests detect popularity dependence, and that dependence does not always move $I$ by more than the interval's width.
\item \emph{Their size on the shuffled pools is only roughly nominal}: $0.00$--$0.16$ for both tests at level $0.05$, where $0.16$ is several standard errors above $0.05$ with 50 training sets, except the stratum test on Criteo at $\bar n\ge10$ ($0.68$--$0.70$). A shuffled pool is one fixed permutation of a few thousand rates, and on Criteo, where one level holds $37\%$ of the rows, the few popular levels' rates differ from the rest by chance; the likelihood interval under-covers in the same cells ($0.64$ at $\bar n=20$), so the flag is not wrong.
\item \emph{On full real data, exchangeability is rejected almost everywhere.} With the whole long tail, the stratum test rejects at $p<10^{-3}$ on 21 of the 23 tasks where it can run, and the likelihood-ratio test on 15 of 24. The effect sizes differ widely: the largest deviation of a stratum's mean label rate from the pooled rate is $0.002$ on the Criteo fields, $0.01$--$0.10$ on most others, and $0.18$--$0.23$ on MovieLens movies, Book-Crossing rated users and \texttt{employee\_salaries}. With millions of rows part of this is power, but the strata do differ by percentage points.
\end{itemize}

\begin{table}[h]
\centering\small
\input{tables/tab_exch_pools}
\caption{Exchangeability tests on the oracle pools: share of the 50 training sets in which the stratum moment test (``stratum'') and the likelihood-ratio test of the frequency-aware prior (``LR'') reject at level $0.05$, and coverage of $I(X;Y)$ by the likelihood interval (``lik.\ cov.'', from \cref{tab:mi-coverage}).}\label{tab:exch-pools}
\end{table}

\begin{table}[h]
\centering\small
\input{tables/tab_exch_tasks}
\caption{Exchangeability tests on the full data of the 24 bakeoff tasks: $p$-values of the stratum moment test and of the likelihood-ratio test of the frequency-aware prior, and the largest absolute deviation of a stratum's mean label rate $\hat m_1$ from the pooled one (effect size).}\label{tab:exch-tasks}
\end{table}

\begin{table}[t]
\centering\small
\resizebox{\linewidth}{!}{\input{tables/tab_mi_bounds}}
\caption{What the full real features say about $I(X;Y)$ (nats) without exchangeability across level sizes. ``pred.\ bound'': $h(\hat\mu)$ minus the held-out cross-entropy of the best bakeoff encoder (Remark~\ref{rem:mi-lower}), valid for any joint law. ``stratified'': identified interval at the plug-in moments when levels in each size group (1, 2, 3, 4, 5--6, 7--10, 11--20, 21--50, 51--100, $>100$) share a prior, identified by its first $\min(\text{smallest size},4)$ moments, with no link between groups; sampling error is not included. ``combined'': the two together. ``singletons'': the singleton group's part of the width, $(\text{its row share})\cdot h(\text{its label rate})$ (\cref{prop:nonid}). ``baseline'', ``NPMLE'': point estimates under exchangeability. ``lik.\ (exch.)'': likelihood interval under exchangeability, computed for tasks up to $10^5$ rows. Script \texttt{experiments/mi\_realdata\_bounds.py}.}\label{tab:mi-bounds}
\end{table}

\section{The NPMLE when levels are small}\label{app:npmle}

The NPMLE of \cref{sec:experiments} is fitted on a grid $q_1<\dots<q_G$ in $[0,1]$ with $G>L+1$ points; a prior is a weight vector $w$ in the simplex, with moments $m_j(w)=\sum_iw_iq_i^j$. With $F_{n,k}$ the number of levels seen $n$ times with $k$ ones and $M=\sum_{n,k}F_{n,k}$ the number of seen levels, the log-likelihood is
\[
\ell(w)=\sum_{n,k}F_{n,k}\log P_w(n,k),\qquad P_w(n,k)=\sum_iw_i\,b_{n,k}(q_i),\qquad b_{n,k}(q)=\tbinom nk q^k(1-q)^{n-k},
\]
and its directional derivative towards a point mass at $q$ is $M\{D_w(q)-1\}$ with
\[
D_w(q)=\frac1M\sum_{n,k}F_{n,k}\,\frac{b_{n,k}(q)}{P_w(n,k)}.
\]
A weight vector $w$ is an NPMLE if and only if $D_w\le1$ on the grid, with equality on the support of $w$ \citep{lindsay1983}. The plug-in information of a prior is $I(w)=h(m_1(w))-\sum_iw_ih(q_i)$.

Parts (i) and (ii) of the next proposition are classical for binomial mixtures \citep{lindsay1983,wood1999}; we state them in the form used here.

\begin{proposition}[The NPMLE at small level sizes]\label{prop:npmle-maxent}
Suppose every level has $n_x\le L$.
\begin{enumerate}
\item[(i)] $\ell(w)$ depends on $w$ only through $m_{1:L}(w)$. If $\hat m$ maximises $\ell$ over the moment space, every $w$ with $m_{1:L}(w)=\hat m$ is an NPMLE. $\hat m$ is unique when every $F_{L,k}$, $k=0,\dots,L$, is positive.
\item[(ii)] All NPMLEs give the same posterior-mean encoding to every level with $n_x\le L-1$, but $\{I(w):w\text{ an NPMLE}\}$ is the whole identified interval $\mathcal I_L(\hat m)$ of \cref{sec:mi}, on the grid. Both ends are attained by NPMLEs with at most $L+1$ atoms.
\item[(iii)] If $\hat m$ lies in the interior of the moment space, then $D_w\equiv1$ on the grid for every NPMLE $w$.
\item[(iv)] Let $w^{(0)}$ have full support and let $w^{(t+1)}_i\propto w^{(t)}_i\exp\{\eta\,D_{w^{(t)}}(q_i)\}$ for some step $\eta>0$ (mirror ascent with the entropy mirror map). Then every iterate has the form $w^{(t)}_i\propto w^{(0)}_i\exp\{\sum_{j=1}^L\lambda^{(t)}_jq_i^j\}$. If the iterates converge to an NPMLE whose moments lie in the interior of the moment space, the limit is
\[
w^\star=\arg\min\big\{\KL(w\,\|\,w^{(0)}):\ m_{1:L}(w)=\hat m\big\},
\]
the I-projection of $w^{(0)}$ onto the NPMLE set \citep{csiszar1975}. For uniform $w^{(0)}$ it is the maximum-entropy law on the grid with moments $\hat m$.
\end{enumerate}
\end{proposition}

\begin{proof}
(i) By Lemma~\ref{lem:moments}, $P_w(n,k)$ is a linear function of $m_0(w)=1,m_1(w),\dots,m_n(w)$, so $\ell(w)=\tilde\ell(m_{1:L}(w))$ and the first statement follows. For the second, the probabilities $P(L,0),\dots,P(L,L)$ are an invertible linear image of $(1,m_1,\dots,m_L)$ (Bernstein against power basis), and $\sum_kF_{L,k}\log P(L,k)$ is strictly concave in them when every $F_{L,k}>0$. So $\tilde\ell$ is strictly concave on the moment space.

(ii) The posterior mean for a level with $(n,k)$ is $P_w(n+1,k+1)/P_w(n,k)\cdot\binom nk/\binom{n+1}{k+1}$, a function of $m_{1:n+1}$, which is fixed on the NPMLE set when $n+1\le L$. $I(w)$ is a linear function of $w$ plus $h(\hat m_1)$, and the NPMLE set is the polytope $\{w:m_{1:L}(w)=\hat m\}$, so the image is the grid version of $\mathcal I_L(\hat m)$. The ends are attained at vertices of the polytope, which are basic solutions of $L+1$ equality constraints and so have at most $L+1$ nonzero weights.

(iii) Write $\partial_j=\partial\tilde\ell/\partial m_j$ at $\hat m$, treating $m_0$ as a free coordinate as well. Then $\partial\ell/\partial w_i=\sum_{j=0}^L\partial_jq_i^j$, and also $\partial\ell/\partial w_i=M\,D_w(q_i)$. Each $P_w(n,k)$ is homogeneous of degree one in $w$, so $\sum_iw_i\,\partial\ell/\partial w_i=M$, that is, $\partial_0+\sum_{j\ge1}\partial_j\hat m_j=M$. At an interior maximiser $\partial_j=0$ for $j\ge1$, so $\partial_0=M$ and $D_w\equiv1$.

(iv) $D_w(q)$ is a linear combination of $b_{n,k}(q)$ with $n\le L$, a polynomial of degree at most $L$ in $q$. Each step adds $\eta D_{w^{(t)}}(q_i)$ to $\log w_i$ up to a constant, so $\log w^{(t)}_i-\log w^{(0)}_i$ is a polynomial of degree at most $L$ in $q_i$, and so is the limit. Laws of the form $w^{(0)}\exp\{\sum_j\lambda_jq^j\}$ form an exponential family with sufficient statistic $(q,\dots,q^L)$; since $G>L+1$ its log-partition function is strictly convex, and each interior moment vector is attained by exactly one member. For any $w$ with $m_{1:L}(w)=\hat m$, $\KL(w\|w^{(0)})=\KL(w\|w^\star)+\KL(w^\star\|w^{(0)})$, because $\log(w^\star/w^{(0)})$ is a polynomial whose expectation is the same under $w$ and $w^\star$. So $w^\star$ is the unique minimiser.
\end{proof}

\begin{remark}[What the implementation returns, and what it means for $I$]\label{rem:npmle-maxent}
\texttt{\pkgname{}} fits the NPMLE by the classical EM update $w_i\leftarrow w_iD_w(q_i)$ \citep{laird1978} from uniform weights. This multiplies by the polynomial $D_w$ rather than by its exponential, so it leaves the exponential family of (iv), but only slightly: with all levels of size at most two and the five priors of \cref{tab:id}, its $\log w$ differs from a quadratic in $q$ by at most $0.06$, and its information is within $2\times10^{-4}$ nats of the maximum-entropy value. The mirror-ascent update stays in the family to machine precision. The NPMLE information at $n_x\le2$ is therefore, for practical purposes, $h(\hat m_1)-\E_{\pi^\star}h(q)$ with $\pi^\star\propto\exp(\lambda_1q+\lambda_2q^2)$ on $[0,1]$. The posterior-average form used in the code, $h(\hat\mu)-\sum_x(n_x/N)\,\E[h(q_x)\mid n_x,k_x]$, equals $I(\hat w)$ when the fitted cell probabilities equal the observed frequencies within each level size, and differs from it by sampling noise otherwise.

The choice is not forced by the data. The two-atom law with the same moments has the same likelihood and gives exactly $I_{\rm lo}$ for all five priors (junk $f=0.5$: $0.170$, against $0.185$ for maximum entropy), and EM started from weights concentrated near $1/2$ or near the edges stops at $0.210$ and $0.195$. For the symmetric case of \citet{hernandez2019}, with $f=F_{2,1}/F_2$ the share of doubletons with one label of each kind, $4\,\mathrm{Var}(q)=1-2f$, the identified interval is $\big[\log2-h\big(\tfrac12+\tfrac12\sqrt{1-2f}\big),\,(1-2f)\log2\big]$, and the moment-matched beta has $a=b=f/(1-2f)$. The maximum-entropy point is close to the beta point for $0.3\le f<0.5$ and below it under strong overdispersion ($f=0.05$: interval $[0.574,0.624]$, beta $0.612$, NPMLE $0.584$), because $\exp(\text{quadratic})$ puts less mass near $0$ and $1$ than a U-shaped beta. At $f\ge1/2$ the only consistent prior is the point mass at $1/2$ and $I=0$; EM approaches it slowly and stops at $0.006$ for $f=1/2$.

So at $n_x\le2$ the NPMLE is a maximum-entropy rule, which is as much an assumption about the unidentified moments as the beta family. Its advantage as an MI estimator in \cref{sec:mi-exp} comes from levels seen three or more times, which feed $m_3,m_4,\dots$ into the likelihood.
\end{remark}

\section{Second-order gap and large-$n$ behaviour}\label{app:gap}

Proposition~\ref{prop:expansion} (proved in Appendix~\ref{app:proofs}) describes the regret of a beta when its moments are close to those of $\pi$. For large $n$ the mismatches $\delta M(k,n-k)$ are not small relative to $M_\pi(k,n-k)$, which decays exponentially in $n$, and a different argument is needed. This appendix makes Remark~\ref{rem:largen} precise. Throughout, $r(k,n)=(k+a)/(n+a+b)$ is the predictive of a fixed $\BetaD(a,b)$ with $a,b>0$, for instance the moment-matched beta or the limit of the type-II ML fit, and $\Delta_n=\Delta_n(r)$.

\subsection{Priors with a density}

\begin{proposition}[Regret at large $n$, priors with a density]\label{prop:largen-smooth}
Let $\pi$ have density $g(q)\propto q^{\alpha-1}(1-q)^{\beta-1}\phi(q)$ with $\alpha,\beta>0$ and $\phi$ Lipschitz on $[0,1]$ with $0<\inf\phi\le\sup\phi<\infty$. Define, for $c>0$,
\[
\rho_n(c)=\begin{cases}n^{-1-c}, & c<1,\\ n^{-2}\log n, & c=1,\\ n^{-2}, & c>1.\end{cases}
\]
Then there is a constant $C$, depending on $\pi$, $a$ and $b$ but not on $n$, such that for $n\ge2$
\[
\Delta_n\ \le\ C\Big(n^{-2}+\mathbb 1[a\neq\alpha]\,\rho_n(\alpha)+\mathbb 1[b\neq\beta]\,\rho_n(\beta)\Big).
\]
In particular $\Delta_n=O(n^{-2})$ when the beta has the boundary exponents of $\pi$ ($a=\alpha$, $b=\beta$) or when $\alpha,\beta>1$; $\Delta_n=O(n^{-2}\log n)$ for a density that is positive and finite at both ends; and $\Delta_n=O(n^{-1-\min(\alpha,\beta)})$ for a U-shaped prior with $\alpha$ or $\beta$ below one.
\end{proposition}

\begin{proof}
Fix $k$ and let $\bar m=(k+\alpha)/(n+\alpha+\beta)$ be the mean of $B'=\BetaD(k+\alpha,n-k+\beta)$. The posterior of $q$ given $K=k$ is $B'$ tilted by $\phi$, so
\[
p_\pi(k,n)-\bar m=\frac{\mathrm{Cov}_{B'}(q,\phi(q))}{\E_{B'}\phi(q)},\qquad
\big|p_\pi(k,n)-\bar m\big|\le\kappa\,\mathrm{Var}_{B'}(q)=\kappa\,\frac{\bar m(1-\bar m)}{n+\alpha+\beta+1},
\]
with $\kappa=\mathrm{Lip}(\phi)/\inf\phi$, since $|\mathrm{Cov}(q,\phi(q))|=\tfrac12|\E(q-q')(\phi(q)-\phi(q'))|\le\mathrm{Lip}(\phi)\,\mathrm{Var}(q)$ for an independent copy $q'$. Next, write $\nu=n+\alpha+\beta$ and $\nu'=n+a+b$; a direct computation gives
\[
\bar m-r(k,n)=\frac{(\alpha-a)(n-k+b)-(\beta-b)(k+a)}{\nu\,\nu'},\qquad r(1-r)=\frac{(k+a)(n-k+b)}{\nu'^2}.
\]
Use $\kl(p\|r)\le(p-r)^2/(r(1-r))$ and $(x+y)^2\le2x^2+2y^2$. Because $r/\bar m$ and $(1-r)/(1-\bar m)$ are bounded below by a positive constant uniformly in $k\le n$, the first part contributes at most $C\,\bar m(1-\bar m)/n^2\le C/n^2$. The second contributes at most
\[
\frac{4}{\nu^2}\Big[(\alpha-a)^2\,\frac{n-k+b}{k+a}+(\beta-b)^2\,\frac{k+a}{n-k+b}\Big].
\]
Averaging over $K$ (Lemma~\ref{lem:regret}), it remains to bound $\E\big[(n-K+b)/(K+a)\big]\le C\,n\,\E[1/(K+1)]$ and its mirror image. Given $q$, $\E[1/(K+1)\mid q]=\{1-(1-q)^{n+1}\}/((n+1)q)\le\min\{1,1/(nq)\}$, and $g(q)\le Cq^{\alpha-1}$ near $0$, so
\[
n\,\E\big[1/(K+1)\big]\le C\,n\Big(\int_0^{1/n}q^{\alpha-1}\,dq+\frac1n\int_{1/n}^1q^{\alpha-2}\,dq\Big)+C
=\begin{cases}O(n^{1-\alpha}),&\alpha<1,\\ O(\log n),&\alpha=1,\\ O(1),&\alpha>1.\end{cases}
\]
Dividing by $\nu^2\asymp n^2$ gives $\rho_n(\alpha)$; the term with $(\beta-b)^2$ is symmetric.
\end{proof}

The following heuristic indicates that the rates are attained. Near $q=0$ both predictives are of order $1/n$: $p_\pi(k,n)\approx(k+\alpha)/n$ and $r(k,n)\approx(k+a)/n$, so a level with small $k$ costs $\kl\approx\frac1n\{(k+\alpha)\log\frac{k+\alpha}{k+a}-(\alpha-a)\}\approx(\alpha-a)^2/(2nk)$ for large $k$, and $P(K=k)\asymp k^{\alpha-1}n^{-\alpha}$. Summing over $k$ up to a small multiple of $n$ gives exactly $\rho_n(\alpha)$: the sum converges for $\alpha<1$, is harmonic for $\alpha=1$, and is dominated by $k\asymp n$ for $\alpha>1$. In words, at large $n$ the beta's regret is governed by how its density behaves at $0$ and $1$, where a mismatch in the exponent shifts predictions of order $1/n$ by an amount of the same order.

\subsection{Atomic priors}

\begin{proposition}[Regret at large $n$, atomic priors]\label{prop:largen-atomic}
Let $\pi=\sum_{j=1}^Jw_j\delta_{\theta_j}$ with $J\ge2$ distinct atoms and minimum spacing $\gamma=\min_{i\ne j}|\theta_i-\theta_j|$. Then
\begin{enumerate}
\item the oracle risk decays exponentially: $R_n(p_\pi)\le h(\varepsilon_n)+\varepsilon_n\log(J-1)$ with $\varepsilon_n=2e^{-n\gamma^2/2}$ (for $\varepsilon_n\le1/2$);
\item for every atom $\theta\in(0,1)$, $n\,\E\big[\kl(\theta\|r(K,n))\mid q=\theta\big]\to\frac12$; an atom at $0$ (or $1$) instead gives $n\,\E[\kl]\to a$ (or $b$).
\end{enumerate}
Hence, if all atoms lie in $(0,1)$, $n\,\Delta_n\to\frac12$ for every beta, whatever its parameters.
\end{proposition}

\begin{proof}
(1) By Lemma~\ref{lem:regret}, $R_n(p_\pi)=I(q;Y_{\rm new}\mid K)\le H(q\mid K)$. Let $\hat\theta(K)$ be the atom closest to $K/n$. It errs only if $|K/n-q|\ge\gamma/2$, which by Hoeffding's inequality has probability at most $2e^{-n\gamma^2/2}$ whatever the true atom, so $P(\hat\theta\ne q)\le\varepsilon_n$. Fano's inequality gives $H(q\mid K)\le h(\varepsilon_n)+\varepsilon_n\log(J-1)$.

(2) Fix $\theta\in(0,1)$, put $s=a+b$, $r=(K+a)/(n+s)$ and $A=\{|r-\theta|\le\theta(1-\theta)/2\}$. On $A$, Taylor's theorem gives $\kl(\theta\|r)=(r-\theta)^2/(2\theta(1-\theta))+O(|r-\theta|^3)$ with a constant depending only on $\theta$. Off $A$, $\kl(\theta\|r)\le\log\{(n+s)/\min(a,b)\}$ because $r\in[a/(n+s),1-b/(n+s)]$, and $P(A^c)$ is exponentially small by Hoeffding. Since $\E(r-\theta)^2=\{n\theta(1-\theta)+(a-s\theta)^2\}/(n+s)^2=\theta(1-\theta)/n+O(n^{-2})$ and $\E|r-\theta|^3=O(n^{-3/2})$, we get $n\,\E\kl(\theta\|r)\to\frac12$. For $\theta=0$, $K=0$ and $\kl(0\|r)=-\log(1-a/(n+s))=a/n+O(n^{-2})$.

The last statement follows from $\Delta_n=R_n(r)-R_n(p_\pi)$, $R_n(r)=\sum_jw_j\E[\kl(\theta_j\|r)\mid q=\theta_j]$, and (1).
\end{proof}

So for an atomic prior the beta keeps the parametric rate $1/(2n)$ of estimating a single Bernoulli parameter, while the oracle, which only has to tell the atoms apart, pays an exponentially small price. The asymptotic regime starts around $n\approx1/\gamma^2$. Before it the regret is below $1/(2n)$ but largest in absolute terms: for the three-atom prior of \cref{tab:regret} ($\gamma=0.1$) it peaks near $n=8$ at $0.022$ nats. This is the regime of the dense cell $\bar n=20$ of \cref{sec:synthetic}, and the regret that the NPMLE recovers there. Since the posterior of $q$ concentrates on the true atom exponentially fast, any prior estimate that puts mass near each atom, such as the NPMLE on a fine grid, inherits the exponential rate up to its own estimation error.

\paragraph{Numerical check.} \Cref{tab:largen} gives exact values of $\Delta_n$ (sums over all binomial outcomes) for the moment-matched beta, divided by the rate predicted above. The ratios level off, at a constant for the first, third and last two priors and slowly for the second, whose rate carries a logarithm (its ratio is $0.0087$ at $n=2^{14}$ and $0.0091$ at $n=2^{16}$). For the atomic priors the ratio reaches $1/2$ once $n\gg1/\gamma^2$.

\begin{table}[h]
\centering\small
\begin{tabular}{llcccccc}
\toprule
Prior $\pi$ & rate & $n=4$ & $16$ & $64$ & $256$ & $1024$ & $4096$\\
\midrule
$\tfrac12\BetaD(2,4)+\tfrac12\BetaD(4,2)$ & $n^{-2}$ & 0.000 & 0.013 & 0.047 & 0.074 & 0.086 & 0.091\\
density $\tfrac23(1+q)$ & $n^{-2}\log n$ & 0.001 & 0.002 & 0.005 & 0.006 & 0.007 & 0.008\\
$\tfrac12\BetaD(0.3,2)+\tfrac12\BetaD(2,0.3)$ & $n^{-1.3}$ & 0.001 & 0.006 & 0.010 & 0.011 & 0.012 & 0.012\\
$\{.2,.3,.85\}$; $\{.3,.4,.3\}$ & $n^{-1}$ & 0.058 & 0.304 & 0.328 & 0.381 & 0.497 & 0.500\\
$\{.1,.5,.9\}$; $\{\frac14,\frac12,\frac14\}$ & $n^{-1}$ & 0.002 & 0.190 & 0.471 & 0.493 & 0.498 & 0.500\\
\bottomrule
\end{tabular}
\caption{Regret $\Delta_n$ of the moment-matched beta at large $n$, divided by the rate of \cref{prop:largen-smooth,prop:largen-atomic}. Moment-matched betas: $\BetaD(1.6,1.6)$, $\BetaD(1.15,0.92)$, $\BetaD(0.23,0.23)$, $\BetaD(0.98,1.27)$ and $\BetaD(1.06,1.06)$, none with the boundary exponents of its prior. Absolute values at $n=4096$: $5.4\times10^{-9}$, $4.1\times10^{-9}$, $2.5\times10^{-7}$, $1.2\times10^{-4}$ and $1.2\times10^{-4}$ nats.}\label{tab:largen}
\end{table}

\section{Downstream models}\label{app:downstream}

This appendix gives the design and full results of \cref{sec:downstream}.

\paragraph{Design.} The single-feature results are about the encoder; in practice the encodings feed a downstream model alongside other columns. We use the Amazon employee-access data (32,769 rows, $P(y=1)=0.942$), whose nine columns are all categorical, and two feature sets: the eight raw columns (\texttt{ROLE\_CODE} relabels \texttt{ROLE\_TITLE} and is dropped), and these plus all 28 pairwise concatenations, which are large and under-sampled. Every encoder replaces every column by its encoding. Rows used to fit the downstream model get out-of-fold encodings from five inner folds, and the test part is encoded by the encoder fitted on the whole training part, so there is no target leakage. Downstream models are logistic regression ($C=1$) on the standardized logits of the encodings and gradient boosting (scikit-learn \texttt{HistGradientBoosting}: learning rate $0.05$, 300 iterations, 31 leaves, $L_2$ penalty 1, no early stopping) on the encodings; the references are one-hot logistic regression ($C=1$) and native CatBoost (500 iterations, default settings). None of these was tuned. Two further encoders add, for every column, the count feature $\log(1+n_x)$, with $n_x$ the level's count in the training part (0 for unseen levels; for a training row, the count of the other training rows, so that training and test rows see the same distribution of $n_x$), next to the baseline or the NPMLE encoding: the practitioner's usual answer to popularity dependence. Outer cross-validation is stratified $5$-fold repeated 3 times, with the same 15 splits for every encoder; differences from the baseline are tested with the corrected resampled $t$-test of \citet{nadeau2003}.

\paragraph{Results} (\cref{tab:downstream}).
\begin{itemize}
\item \emph{With a linear model the encoder matters a little, and the ranking follows the single-feature results.} The NPMLE beats the baseline on all 15 folds, by $0.0012$ nats on the raw columns and $0.0019$ on the pairwise set; the frequency prior beats it by $0.0002$ and $0.0013$ nats (14 and 15 of 15 folds). The GLMM at its conditional mode and fixed $m=20$ lose $0.0045$ and $0.0035$ nats on the sparse pairwise columns, never beating the baseline on a fold; the GLMM loss matches what we saw on the click logs.
\item \emph{With gradient boosting the encoders are indistinguishable}, all within $0.0009$ nats of the baseline and none significant, except GLMM on the pairwise set ($+0.0052$ nats, 0 of 15 folds better). A flexible downstream model re-learns most of the calibration that a better prior provides.
\item \emph{The GLMM's loss is again its prediction rule.} Predicting by the posterior mean instead of the conditional mode turns its $+4.5$ and $+5.2$ milli-nats on the pairwise set into $-1.3$ (logistic, 15 of 15 folds better) and $-0.8$ (boosting, not significant), close to the NPMLE.
\item \emph{A count feature helps more than most changes of prior.} With boosting, adding $\log(1+n_x)$ to the baseline gains $4.9$ milli-nats on the raw columns and $1.8$ on the pairwise set, more than the spread between encoders, and on top of the NPMLE about the same. With logistic regression it gains $3.3$ milli-nats on the pairwise set ($2.6$ on top of the NPMLE) and $0.7$ ($p=0.05$) on the raw columns. Popularity information is valuable downstream, and a separate feature lets the model use it in ways a prior cannot.
\item \emph{Features matter more than the encoder.} Adding the pairwise columns gains $0.016$--$0.019$ nats, about ten times the spread between encoders. Native CatBoost on the raw columns, which builds its own feature combinations, is best overall ($0.1366$), ahead of every encoder-plus-model pipeline (best $0.1396$), and one-hot logistic regression beats every target-encoded logistic regression on the raw columns.
\end{itemize}
The downstream test therefore supports a modest claim: no encoder improves on the type-II ML baseline by more than $2$ milli-nats in any setting, the NPMLE gives a small but consistent gain with a linear model, the GLMM at its conditional mode or fixed smoothing can hurt on sparse columns, and a count feature helps both models. It does not support the claim that the choice of encoder drives downstream accuracy.

\begin{table}[h]
\centering
\small
\resizebox{\linewidth}{!}{\input{tables/tab_downstream}}
\caption{Downstream log-loss on Amazon employee access (mean over 15 paired folds, nats). $\Delta$: encoder minus baseline in milli-nats (negative is better); $^{*}$: $p<0.05$ in the corrected resampled $t$-test. With boosting on the raw columns all encoders without the count feature are within $\pm0.0008$ nats of the baseline ($p>0.38$), so those cells are left blank. ``+ count'': the encoding plus $\log(1+n_x)$ for every column.}\label{tab:downstream}
\end{table}

\section{GLMM fitting: adaptive quadrature vs Laplace}\label{app:glmm}

\paragraph{Model and fit.} The GLMM encoder of \citet{pargent2022} is the random-intercept logistic model $\operatorname{logit}q_x=b_0+u_x$, $u_x\sim\mathcal N(0,\sigma^2)$, with $(b_0,\sigma)$ fitted by maximising the marginal likelihood
\[
\ell(b_0,\sigma)=\sum_x\log\int\binom{n_x}{k_x}\operatorname{expit}(b_0+u)^{k_x}\{1-\operatorname{expit}(b_0+u)\}^{n_x-k_x}\,\varphi_\sigma(u)\,du ,
\]
and each level encoded by $\operatorname{expit}(b_0+\hat u_x)$ at the conditional mode $\hat u_x$; unseen levels get $\operatorname{expit}(b_0)$. In the beta-binomial language of \cref{sec:setup} this is a logit-normal prior for $q$, fitted by type-II ML, with a plug-in (mode, not mean) prediction. The integral depends on a level only through $(n_x,k_x)$, so it is computed once per distinct pair. Our implementation (\texttt{\pkgname{}.fit\_glmm}) evaluates it with 15-node adaptive Gauss--Hermite quadrature (AGQ) centred at the conditional mode and scaled by the curvature there. For levels whose posterior is wide compared with the prior (standard deviation above $0.15\sigma$; in practice small all-zero or all-one levels at large $\sigma$), where the posterior is one-sided and the Gaussian nodes misplace the mass, it integrates instead on a 1,601-point grid of the prior over $\pm8\sigma$. $(b_0,\log\sigma)$ are found by L-BFGS-B with $\log\sigma\in[-7,4]$, started at the logit of the label rate and $\sigma=1$. With one node and no grid, the same code is the Laplace approximation, which is what \texttt{lme4::glmer} uses by default (\texttt{nAGQ = 1}).

\paragraph{Why the Laplace approximation fails on singletons.} If every level is a singleton, the data identify only $\E_\pi q=\E\operatorname{expit}(b_0+u)$ (\cref{prop:ml}, part 1 is the beta version): the exact likelihood is constant along a ridge of $(b_0,\sigma)$ and only the larger levels choose a point on it. The Laplace approximation of a singleton's probabilities is not a probability distribution, and its error grows with $\sigma$. At $b_0=-10$, $\sigma=10$ it gives $0.998$ for $y=0$ and $0.126$ for $y=1$ (sum $1.12$), against the exact $0.838$ and $0.162$; at $b_0=-2$, $\sigma=1$ it is accurate to $0.002$. When most labels are zero, moving along the ridge towards large $\sigma$ and very negative $b_0$ makes the many zero-singletons nearly certain while the few one-singletons keep a sizeable probability, so the approximate likelihood increases without bound. On the churn pseudo-identifier the profile Laplace log-likelihood (maximised over $b_0$) rises by $475$ nats between $\sigma=0.57$ and $\sigma=10$ and keeps rising, while the AGQ profile peaks near $\sigma=0.57$ and is $2.0$ and $157$ nats lower at $\sigma=1.1$ and $\sigma=10$. The Laplace fit therefore stops wherever the optimiser or a bound stops it, with an intercept that predicts almost surely $0$ for every unseen or new level and an encoding that is far too extreme for the levels that were seen.

\paragraph{Held-out comparison.} \Cref{tab:glmm} compares the two fits on three singleton-heavy features with stratified 5-fold cross-validation, and on synthetic beta data with exact excess cross-entropy. On the two features where nearly all seen levels are singletons the Laplace fit runs to $\sigma=24$ and $45$ and loses more than $1$ nat per row; AGQ fixes this. On the 500,000-row Avazu sample, where $62\%$ of seen levels are singletons but many levels are large, the Laplace fit is finite and only moderately worse. AGQ is not enough to make the plug-in GLMM competitive on these features: it is still $0.009$--$0.014$ nats worse than the beta baseline on Avazu and on the click user identifier, and on the latter worse than the marginal predictor. The reason is the prediction rule, not the logit-normal prior: the same AGQ fit encoded by the posterior mean is better than the baseline on the click features of the bakeoff (\cref{sec:bakeoff}, ``GLMM mean''). When the conditionals are drawn from a beta, the two fits are within $0.0002$ nats of each other and $0.002$ nats behind the baseline, whose prior is then correct.

\begin{table}[h]
\centering\small
\resizebox{\linewidth}{!}{%
\begin{tabular}{lcccccc}
\toprule
& seen levels & singletons & baseline & GLMM, AGQ-15 & GLMM, Laplace & marginal\\
\midrule
churn pseudo-id & 3,603 & 90\% & 0.408 & 0.409 ($\sigma=0.42$) & 1.484 ($\sigma=23.6$) & 0.408\\
\texttt{click\_small/user\_id} & 24,121 & 99\% & 0.451 & 0.465 ($\sigma=1.44$) & 1.721 ($\sigma=44.7$) & 0.453\\
Avazu \texttt{device\_ip}, 500k rows & 149,623 & 62\% & 0.409 & 0.418 ($\sigma=1.60$) & 0.424 ($\sigma=1.75$) & 0.446\\
\midrule
beta, $\bar n=1$ (eCE) & 12,642 & 58\% & 0.0761 & 0.0786 ($\sigma=1.03$) & 0.0784 ($\sigma=0.76$) & \\
beta, $\bar n=2$ (eCE) & 17,293 & 31\% & 0.0667 & 0.0682 ($\sigma=1.04$) & 0.0680 ($\sigma=0.89$) & \\
\bottomrule
\end{tabular}}
\caption{GLMM encoder fitted by 15-node adaptive quadrature and by the Laplace approximation. Real features: held-out log-loss (nats), mean over 5 folds, with the mean fitted $\sigma$; ``singletons'': share of the levels seen in a training fold that are seen once. Synthetic: $\nlev=2\times10^4$ equally popular levels with $q_x\sim\BetaD(1.5,3.5)$ and Poisson$(\bar n)$ level sizes, exact excess cross-entropy, mean over 10 data sets (standard deviations $\le0.001$); seen levels and singleton shares are expected values. The Laplace failure needs singleton-heavy data, not signal: on two of the three real features the baseline barely beats the marginal predictor.}\label{tab:glmm}
\end{table}

\section{Real-data results in detail}\label{app:details}

This appendix gives the full account of the synthetic sweep (\cref{sec:synthetic}), the oracle pools (\cref{sec:oracle}) and the bakeoff (\cref{sec:bakeoff}) that the main text summarises.

\begin{figure}[h]
\centering
\includegraphics[width=\linewidth]{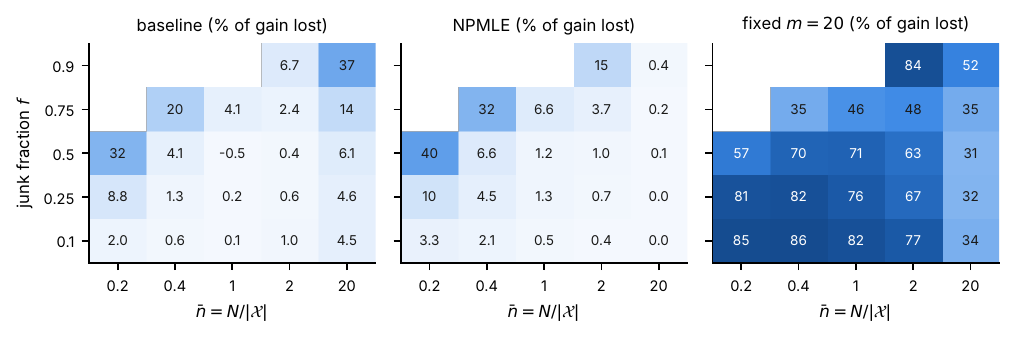}
\caption{Junk-fraction family with unknown $(f,q_0)$: share of the oracle's gain lost by each method, as a function of the junk fraction $f$ and the mean level size $\bar n=N/\nlev$. Blank cells: the oracle gains less than $0.001$ nats over the marginal predictor. 40 data sets per cell; standard errors of the gaps are below $0.0003$ nats at $\bar n\le2$ and $0.0012$ nats at $\bar n=20$.}\label{fig:junk}
\end{figure}

\paragraph{Synthetic junk-fraction sweep.} \Cref{fig:junk} gives every cell; \cref{sec:synthetic} states the conclusions. Three details are not there. At $\bar n\le2$ the baseline's gap is below $0.0004$ nats at $\bar n\ge0.4$ except in one cell, and large shares of gain lost appear only where there is almost nothing to gain (at $f\ge0.75$ and $\bar n\le0.4$ the oracle beats the marginal predictor by under $0.002$ nats). The NPMLE is worse than the baseline in 18 of 19 cells at $\bar n\le2$ if the three cells left blank in \cref{fig:junk} (family oracle gain under $0.001$ nats) are counted; its losses reach $0.0008$ nats ($f=0.25$, $\bar n=0.4$), and at $f=0.9$, $\bar n=1$ it gives up $35\%$ of the gain against $13\%$ for the baseline. At $\bar n=20$ the baseline's gap is $0.005$--$0.008$ nats, the gap of \cref{prop:expansion,rem:largen} for an atomic prior, and the NPMLE's is $0.0$--$0.4\%$ of the gain.

\begin{figure}[h]
\centering
\includegraphics[width=0.6\linewidth]{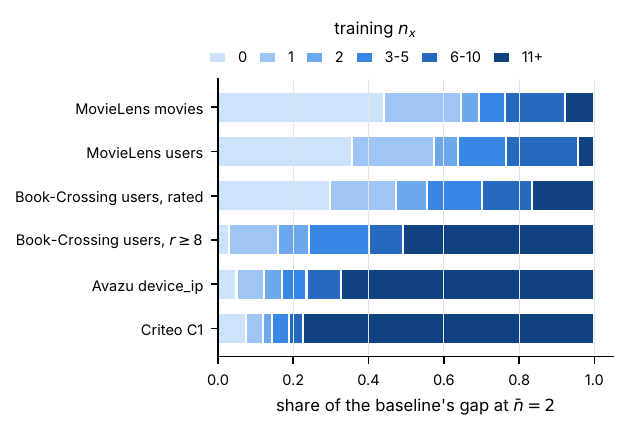}
\caption{Where the baseline's gap lives at $\bar n=2$: its split by the training size $n_x$ of the test row's level. With popularity dependence (top three pools) much of it is on levels seen at most once, where the frequency prior acts; without it (bottom three), most of it is on levels seen 11 or more times, where only a richer prior can help.}\label{fig:gap-by-n}
\end{figure}

\paragraph{Oracle pools at $\bar n=2$.} \Cref{sec:oracle} states the conclusions; these are the details behind them, for the head-only pools.
\begin{itemize}
\item \emph{Prior shape.} On Avazu the moment-matched beta loses $2.0$ milli-nats against $2.2$ for the fitted baseline. The NPMLE's losses on the two smaller pools are $-72\pm18\%$ of the gap (Book-Crossing $r\ge8$) and $-17\pm9\%$ (Criteo). Adding the Robbins correction does not rescue it ($24\pm5\%$ on Avazu, $-79\pm19\%$ and $-34\pm11\%$ on the other two), because its encoding on levels seen twice or more is still the noisy NPMLE.
\item \emph{Popularity dependence.} The popularity-blind oracle loses $7.3$ milli-nats on movies against $7.9$ for the baseline. The frequency prior closes $81\%$ (movies), $81\%$ (users) and $50\%$ (Book-Crossing rated) of the gap, the unseen correction $43\%$, $32\%$ and $28\%$ ($\pm1$--$2\%$, $p<10^{-4}$), and the Robbins correction $59\%$, $36\%$ and $34\%$ ($\pm1$--$3\%$); on movies the Robbins correction cuts the part of the gap on once-seen levels from $1.6$ to $0.4$ milli-nats, and the frequency prior stays ahead because it also corrects levels seen two to ten times. On top of the NPMLE, the Robbins correction lifts it from $14\%$ to $41\%$ on movies, from $-9\%$ to $9\%$ on users and from $21\%$ to $35\%$ on Book-Crossing rated. On the three pools without popularity dependence the frequency prior closes $-1$ to $0\%$, the unseen correction $0$ to $-11\%$ and the Robbins correction $-1$ to $-14\%$; without its doubleton test the Robbins correction would cost $36$--$58\%$ of the gap on Book-Crossing $r\ge8$ and Criteo.
\end{itemize}

\paragraph{Other populations.} \Cref{tab:oracle-sens} gives the same summary for head-only pools with thresholds of 50 and 100 rows and for the head + tail populations of \cref{sec:oracle}, 20 training sets per cell (10 on the two click pools with the tail, which have about a million levels; 50 for the head-only pools). In the head + tail population every level of the source keeps its real row share; a level below 100 rows gets $q_x=\operatorname{expit}(\ell+c_b)$, with $\ell$ drawn from the head's values of $\operatorname{logit}q$ (clipped at $0.5/n$ from 0 and 1) and $c_b$ chosen so that the mean of $q$ over the level's size bucket $b$ equals that bucket's real label rate (\texttt{realdata\_oracle.py --tail shifted-head}). The tail thus has the real link between popularity and the mean rate, but within a bucket the head's shape, and no link within a bucket. The shuffled controls of the head + tail populations (supplementary CSVs) permute these $q$ across all levels; there, as for the head-only pools, the LR test keeps the frequency prior at the baseline ($-4$ to $+1\%$ of the gap), and the unseen and Robbins corrections close between $-38\%$ and $+2\%$ of gaps below $0.8$ milli-nats on MovieLens and Book-Crossing. The exception is Criteo, where they close $31$--$45\%$ of a $0.1$-milli-nat gap: one level holds $37\%$ of the rows, so the row-weighted mean $\hat\mu$ is pulled towards that one level's rate and away from the rate of the small levels, even though the permutation removes any systematic link to popularity. On the shuffled Avazu head + tail population the baseline's gap ($1.1$ milli-nats at $\bar n=2$) has under $1\%$ on levels seen at most once, the corrections and the frequency prior close $0$--$1\%$ of it, and the NPMLE closes $98\%$: with the link to popularity removed, what remains is prior shape. The $23\%$ of the real Avazu gap at $n_x\le1$, and what the corrections recover there, therefore come from the tail's dependence on popularity. Thresholds of 500 rows leave $29$--$618$ levels per pool ($29$ on Book-Crossing $r\ge8$, $268$ on Avazu), too few to fit the NPMLE or the frequency prior reliably; they are in the supplementary CSVs. With the 50-row threshold only the methods of \cref{tab:oracle} were run.

\begin{table}[!ht]
\centering\small
\resizebox{\linewidth}{!}{\input{tables/tab_oracle_sensitivity}}
\caption{Real-data oracle at $\bar n=2$ under three ways of building the population: head-only pools with at least 100 rows per level (as in \cref{tab:oracle}) or at least 50, and the semi-synthetic head + tail population. ``baseline gap'': gap to the full oracle in milli-nats; ``at $n_x\le1$'': its share on levels seen at most once; ``closed'': share of the baseline's gap closed, $\pm$ paired standard error. ``--'': not run. Script \texttt{experiments/oracle\_sensitivity.py}.}\label{tab:oracle-sens}
\end{table}

\paragraph{Bakeoff.} \Cref{sec:bakeoff} gives the results against the baseline. \Cref{tab:wtl} compares the fitted encoders with each other: for each pair, the number of the 24 tasks on which one is significantly better or worse than the other, with Holm's correction over all 45 pairs of the ten encoders shown and the other two (66 pairs) within each task, so the family is fixed in advance; the rest are ties. Under this stricter correction the NPMLE beats the baseline on 13 tasks instead of 14. The table supports three statements of the main text. The Robbins correction and the NPMLE tie (4 wins, 5 losses). Putting the Robbins correction on top of the NPMLE never loses to any other encoder and beats the NPMLE on 7 tasks. The posterior-mean GLMM beats its plug-in version on 9 tasks and never loses to it, while it loses to the NPMLE on 4 and never beats it.

\begin{table}[!ht]
\centering\small
\resizebox{\linewidth}{!}{\input{tables/tab_wtl}}
\caption{Paired comparisons of fitted encoders over the 24 bakeoff tasks. Cell (row A, column B): number of tasks on which A is significantly better than B / significantly worse (corrected resampled $t$-test over 25 splits, Holm over all 66 pairs of fitted encoders within a task, $p_{\rm Holm}<0.01$); the remaining tasks are ties. Script \texttt{experiments/bakeoff\_summary.py}.}\label{tab:wtl}
\end{table}

Against the baseline, fixed $m=20$ is also significantly better on one task (Book-Crossing rated ISBN, by $0.6\%$) and loses $23\%$ of the gain on Amazon \texttt{MGR\_ID}, $16\%$ on \texttt{employee\_salaries} and $10$--$13\%$ on the Book-Crossing users. CatBoost's prediction-time statistic ($m=1$) loses up to $18\%$ on the features of \cref{tab:bakeoff} with a gain above $0.03$ nats, and more than the whole gain on the weak ones (Criteo, Book-Crossing ISBN, Amazon \texttt{RESOURCE}). Scikit-learn's automatic smoothing, which gives singletons encodings near 0 or 1, has log-loss above the marginal predictor's on every sparse feature (supplementary CSVs).

\begin{table}[t]
\centering
\small
\resizebox{\linewidth}{!}{\input{tables/tab_bakeoff}}
\caption{Single-feature bakeoff on nine sparse features, 25 paired splits per task. The nine illustrate one to three sparse features from each of the three largest sources (click logs, Book-Crossing, Amazon) and the multi-class feature; they were picked for illustration, and \cref{tab:bakeoff-summary} and the full \cref{tab:bakeoff-full} use all 24 tasks. ``gain'': held-out log-loss of the marginal predictor minus that of the baseline (nats). Method columns: the method's log-loss minus the baseline's, as a percentage of that gain (negative is better); ``+unseen'': baseline + unseen correction; ``+Robbins'': baseline + Robbins correction for $n\le1$ (\cref{cor:robbins}); ``NPMLE +Robbins'': the same correction on top of the NPMLE; ``GLMM mode'': the GLMM encoded at the conditional mode (the usual plug-in), ``GLMM mean'': the same fit encoded by the posterior mean; $^{*}$: $p_{\rm Holm}<0.01$; bold: best significant improvement. Last two columns: share of that improvement coming from test rows on levels unseen in training, shown where the improvement is significant; it exceeds $100\%$ when the method is slightly worse than the baseline on seen levels, and is negative when the gain on seen levels outweighs a loss on unseen ones. Sample sizes, numbers of levels, the cross-validated $m$ and CatBoost's target statistic are in \cref{tab:bakeoff-full}. $^\dagger$Four classes, fitted one-vs-rest.}\label{tab:bakeoff}
\end{table}

\begin{figure}[t]
\centering
\includegraphics[width=\linewidth]{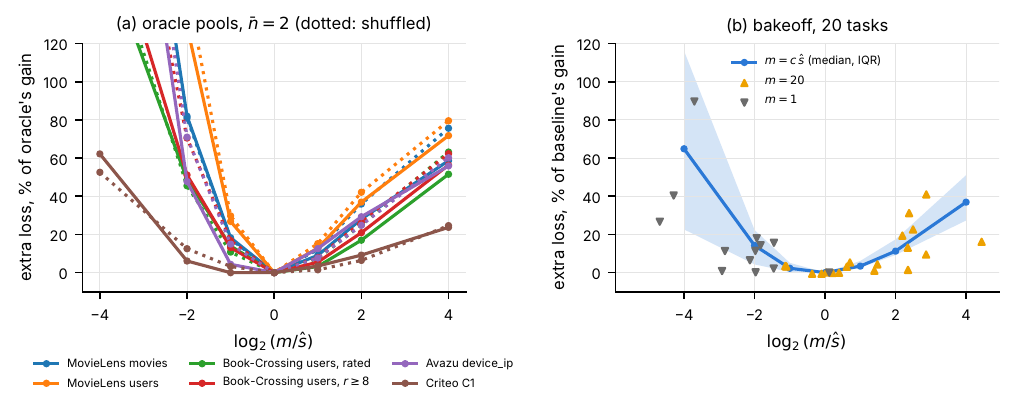}
\caption{Fixed smoothing against the fitted value $\hat s$. (a) Oracle pools at $\bar n=2$: excess cross-entropy of the $m$-estimate with $m=c\,\hat s$ minus the baseline's ($c=1$), as a share of the full oracle's gain, mean over 20 training sets; solid real, dotted shuffled. (b) Bakeoff: held-out log-loss of the same encoders minus the baseline's, as a share of the baseline's gain, median and interquartile range over the fixed set of 20 tasks, with fixed $m=20$ and $m=1$ placed at $\log_2(m/\hat s)$ for each task (median $\hat s$ over the 25 training folds). Script \texttt{experiments/paper/fig\_fixed\_m.py}.}\label{fig:fixed-m}
\end{figure}

\begin{table}[t]
\centering\small
\resizebox{0.8\linewidth}{!}{\input{tables/tab_fixed_m}}
\caption{Fitted and fixed smoothing on the 24 bakeoff tasks: median over the 25 training folds of the marginal-likelihood $\hat s$ and of the leave-one-out CV choice of $m$, and the held-out log-loss of fixed $m=20$ and $m=1$ minus the baseline's, in milli-nats per test row and as a share of the baseline's gain (``--'' where that gain is below $0.5\%$ of $H(Y)$). For \texttt{employee\_salaries} $\hat s$ is the Dirichlet concentration. Script \texttt{experiments/fixed\_m\_sweep.py}.}\label{tab:fixed-m}
\end{table}

\section{Further figures}\label{app:figs}

\begin{figure}[t]
\centering
\includegraphics[width=\linewidth]{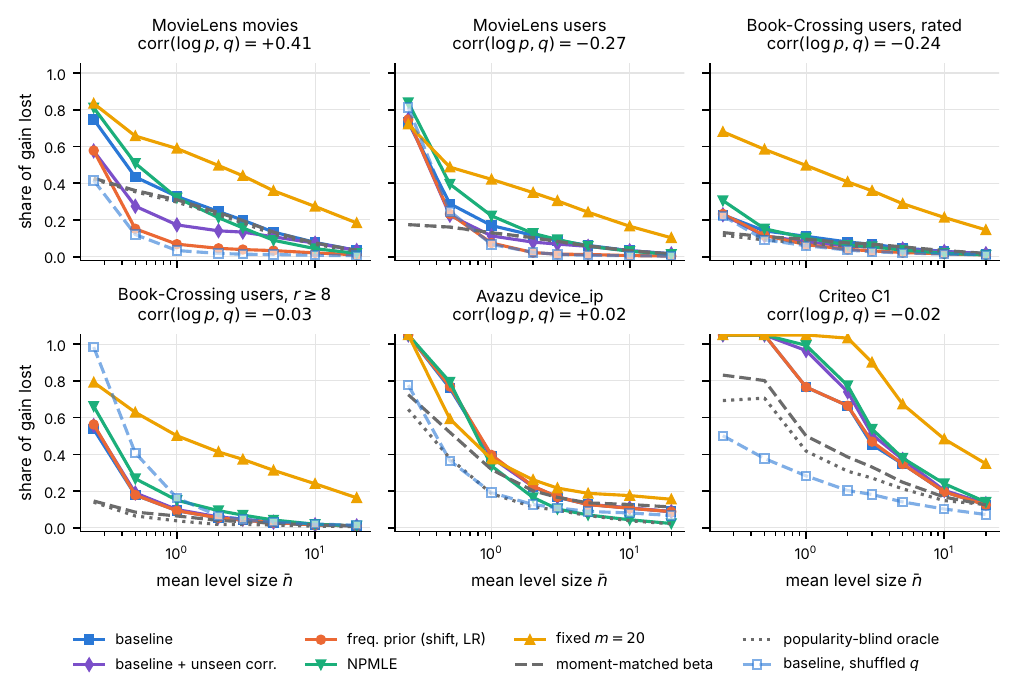}
\caption{Real-data subsampling oracle: share of the full oracle's gain lost against the mean level size $\bar n$, one panel per pool. Solid lines: real data. Open squares: baseline on the shuffled control, where $q_x$ is permuted across levels. Grey lines: the moment-matched beta (dashed) and the popularity-blind oracle (dotted), both built from the true $q$. 50 training sets per point. Values above 1 are clipped.}\label{fig:oracle}
\end{figure}

\begin{figure}[t]
\centering
\includegraphics[width=\linewidth]{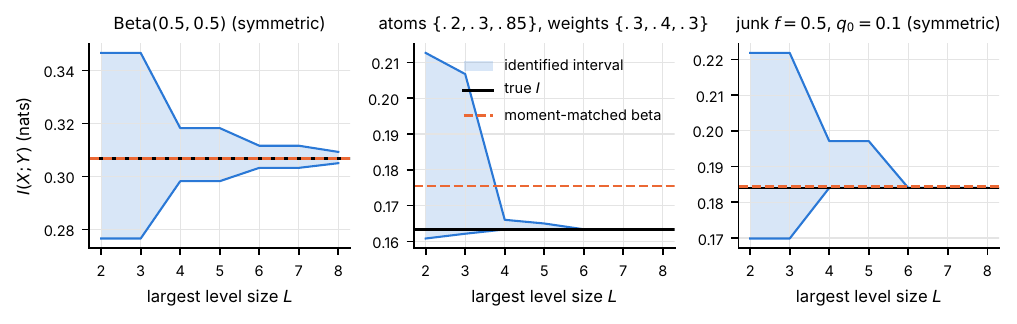}
\caption{The identified interval for $I(X;Y)$ shrinks with the largest usable level size $L$ (shaded; $L=1$ gives $[0,h(\mu)]$ and is not drawn). Solid: the true $I$. Dashed: the value implied by the moment-matched beta, the limit of the beta estimator when all levels have $n\le2$. For the symmetric priors the interval only shrinks at even $L$, because $m_3$ is fixed by $m_1,m_2$. The beta point is exact for the beta prior, $0.0005$ nats off for the junk prior, and $7\%$ off for the asymmetric three-atom prior, where it leaves the interval from $L=4$ on.}\label{fig:interval}
\end{figure}

\begin{figure}[t]
\centering
\includegraphics[width=0.6\linewidth]{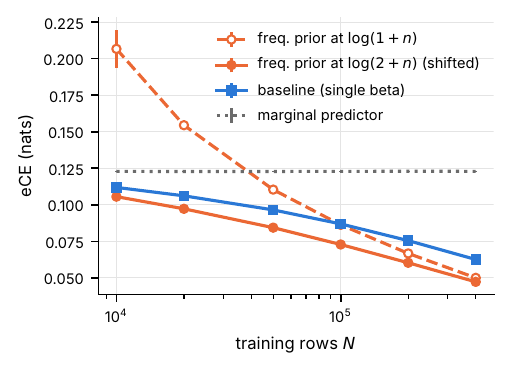}
\caption{The $n+1$ shift on a synthetic popularity-linked prior: $2\times10^5$ levels with $\log p_x\propto1.5z_x$ and $\operatorname{logit}q_x=-0.3+0.8z_x+\varepsilon_x$, $z_x,\varepsilon_x\sim\mathcal N(0,1)$. The frequency prior (log-linear spread, no LR test) read at $\log(1+n)$ and at $\log(2+n)$, the baseline, and the marginal predictor. Mean $\pm$ s.e.\ over 20 draws.}\label{fig:shift}
\end{figure}

\begin{figure}[t]
\centering
\includegraphics[width=\linewidth]{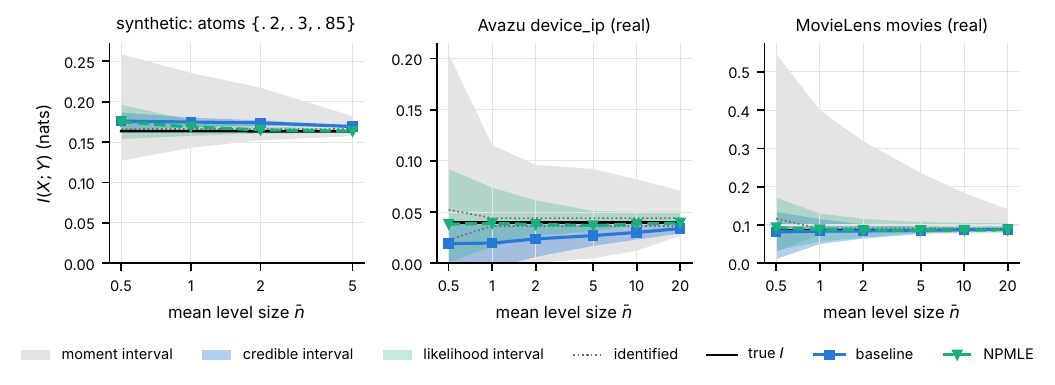}
\caption{Intervals for $I(X;Y)$ against the mean level size $\bar n$: median lower and upper ends of the 95\% moment interval (grey), the likelihood interval (green) and the credible interval of \citet{hernandez2019} (blue), the identified interval at the population's moments (dotted, nearly on top of the true $I$), the true $I$ (black), and the mean baseline and NPMLE estimates. Left: three-atom prior, $10^5$ equally popular levels, where the baseline's estimate, and hence its credible interval, sits $0.011$ nats above $I$. Middle: Avazu \texttt{device\_ip}, where the baseline underestimates $I$ and its interval misses. Right: MovieLens movies, where popularity dependence pushes the moment interval above $I$ at small $\bar n$ and the likelihood interval above it at large $\bar n$.}\label{fig:mi-coverage}
\end{figure}

\clearpage
\section{Datasets and the subsampling design}\label{app:data}

\begin{table}[!ht]
\centering\small
\resizebox{\linewidth}{!}{%
\begin{tabular}{llll}
\toprule
Method & \texttt{method=} & Prior and prediction rule & Tuning and gate\\
\midrule
baseline & \texttt{baseline} & beta centred on $\hat\mu$, posterior mean & $s$ by type-II ML\\
+unseen & \texttt{baseline\_unseen} & baseline; unseen levels get the singletons' rate & none\\
+Robbins & \texttt{baseline\_robbins} & +unseen; once-seen levels from doubleton rates (\cref{prop:pop-n1}) & doubleton LR test, level $0.05$\\
frequency prior & \texttt{freq\_prior} & beta-binomial regression on $\log(2+n_x)$ (\cref{sec:popularity}) & LR test of $b=d=0$, else baseline\\
NPMLE & \texttt{npmle} & grid prior (400 points), posterior mean & EM to convergence\\
NPMLE +Robbins & \texttt{npmle\_robbins} & NPMLE; unseen and once-seen as for +Robbins & $G$-test of doubleton pairs, level $0.05$\\
GLMM mode & \texttt{glmm} & logit-normal; $\operatorname{expit}(b_0+\hat u_x)$ at the mode, $\operatorname{expit}(b_0)$ if unseen & $(b_0,\sigma)$ by 15-node AGQ\\
GLMM mean & \texttt{glmm\_mean} & same fit; posterior mean of $q_x$, prior mean if unseen & as GLMM mode\\
CV $m$ & \texttt{cv\_m} & $m$-estimate & $m$ by leave-one-out log-loss\\
fixed $m=20$ & \texttt{fixed\_m} & $m$-estimate, $m=20$ & none\\
$m=1$ (CatBoost) & \texttt{catboost\_ts} & $m$-estimate, $m=1$ (CatBoost's prediction-time statistic) & none\\
\midrule
family oracle (\cref{sec:synthetic}) & -- & Bayes rule in the junk family & $(f,q_0)$ by marginal likelihood\\
full oracle (\cref{sec:oracle}) & -- & Bayes rule under the pool's true joint law of $(p_x,q_x)$ & none (knows the pool)\\
popularity-blind oracle & -- & Bayes rule under the pool's row-weighted law of $q$, ignoring $n_x$ & none (knows the pool)\\
moment-matched beta & -- & beta with the first two moments of that law & none (knows the pool)\\
\bottomrule
\end{tabular}}
\caption{Methods of \cref{sec:experiments} and their \texttt{method=} strings in \texttt{\pkgname{}.fit\_counts}, with the reference predictors that use the truth. In the bakeoff, Holm's correction on each task is over the 11 comparisons with the baseline: the 10 other encoders above and the Robbins correction without its doubleton test (\texttt{baseline\_robbins\_raw}).}\label{tab:methods}
\end{table}

\paragraph{Sources and targets.} \Cref{tab:data} lists the data. Every task is one categorical feature and a binary target (four salary quartiles for \texttt{employee\_salaries}); levels are the raw values of the feature as strings, with missing values forming one level for the Criteo fields and dropped elsewhere. No level is merged, filtered or hashed beyond what the providers did, since rare levels are the object of study. This is why we use raw rows and not the preprocessed Avazu and Criteo files distributed with click-through-rate benchmark suites, which fold rare identifiers into shared buckets.
\begin{itemize}
\item \emph{MovieLens 1M} \citep{harper2015}: 1,000,209 ratings of 3,706 movies by 6,040 users; $y=\mathbb 1[\text{rating}\ge4]$ ($\bar y=0.575$); features \texttt{movie\_id} and \texttt{user\_id}.
\item \emph{Book-Crossing} \citep{ziegler2005}: 1,149,780 ratings, of which 433,671 are explicit (1--10) and the rest implicit (0). Main target $y=\mathbb 1[\text{rating}\ge8]$ among explicit ratings ($\bar y=0.576$; 8 is the median); features ISBN, user, and, joined from the book and user tables, author, publisher and location (rows whose ISBN is not in the book table are dropped for author and publisher). The \emph{rated} target is explicit rating versus implicit interaction on all rows ($\bar y=0.377$), with features ISBN and user. Book and user tables are read as Latin-1, skipping malformed lines (19 of the 271,379 lines of the book table).
\item \emph{Amazon employee access} \citep{kaggleamazon2013}: the 32,769 labelled rows of the training file, $y=$ \texttt{ACTION} ($\bar y=0.942$); features \texttt{RESOURCE}, \texttt{MGR\_ID} and \texttt{ROLE\_FAMILY\_DESC} for the bakeoff and all nine columns for \cref{app:downstream} (\texttt{ROLE\_CODE} is a relabelling of \texttt{ROLE\_TITLE} and is dropped there).
\item \emph{Avazu} \citep{kaggleavazu2014}: the first 4,000,000 rows of the raw training file, one day (21 October 2014, 23 hours), $y=$ click ($\bar y=0.174$); features \texttt{device\_ip} and the ``user'' \texttt{device\_ip}$\times$\texttt{device\_model}.
\item \emph{Criteo 1TB click logs} \citep{criteo1tb}: six of the parquet parts of day 0 (parts 79, 104, 115, 119, 124 and 141 of the Hugging Face copy), 4,591,543 rows, $y=$ click ($\bar y=0.032$); hashed categorical fields \texttt{C1}, \texttt{C20}, \texttt{C22}.
\item \emph{OpenML} \citep{vanschoren2014}: the ARFF exports of four OpenML data sets: \texttt{employee\_salaries} (ID 42125; \texttt{employee\_position\_title}, $y=$ quartile of \texttt{current\_annual\_salary}), \texttt{churn} (ID 40701; \texttt{state}, and the pseudo-identifier \texttt{state}$\times$\texttt{area\_code}$\times$\texttt{account\_length}), a 5\% random sample (26,969 rows) of \texttt{airlines} (ID 1169, 539,383 rows; \texttt{AirportFrom}, \texttt{Flight}; $y=$ \texttt{Delay}), used as stored in the ARFF export of the original bakeoff, whose sampling seed is not recorded and \texttt{Click\_prediction\_small} (ID 1220; \texttt{advertiser\_id}, \texttt{user\_id}).
\end{itemize}
Per-task sizes, numbers of levels and $(F_1+2F_2)/N$ are in \ifpreprint\texttt{bakeoff/tables/table\_tasks.csv} of the results archive (\cref{app:repro})\else the supplementary table \texttt{table\_tasks.csv}\fi; the sparse features have $(F_1+2F_2)/N$ between $0.14$ (Book-Crossing author) and $0.42$ (Book-Crossing ISBN) on the full data, and up to $0.96$ for the churn pseudo-identifier.

\begin{table}[h]
\centering\small
\resizebox{\linewidth}{!}{%
\begin{tabular}{llrrl}
\toprule
Source & Tasks & Rows & Levels (largest feature) & Licence / terms\\
\midrule
MovieLens 1M & 2 bakeoff, 2 pools & 1,000,209 & 6,040 & GroupLens research licence; no redistribution\\
Book-Crossing & 7 bakeoff, 2 pools & 1,149,780 & 340,556 & free for research use, with citation\\
Amazon employee access & 3 bakeoff, downstream & 32,769 & 7,518 & Kaggle competition rules\\
Avazu & 2 bakeoff, 1 pool & 4,000,000 & 1,549,515 & Kaggle competition rules\\
Criteo 1TB, day 0 (6 parts) & 3 bakeoff, 1 pool & 4,591,543 & 965,565 & CC BY-NC-SA 4.0\\
OpenML (4 data sets) & 7 bakeoff & 5,000--39,948 & 30,114 & OpenML terms of each data set\\
\bottomrule
\end{tabular}}
\caption{Data sources. None of the data is redistributed with the code; the loaders in \texttt{experiments/realdata\_tasks.py} read the files as obtained from the providers.}\label{tab:data}
\end{table}

\paragraph{Subsampling oracle pools.} A pool is built from a feature and a target by keeping the levels with at least 100 rows in the full data and all their rows. The population is the empirical distribution of those rows: a row falls in level $x$ with probability $p_x=M_x/M$, where $M_x$ is the level's number of rows and $M=\sum_xM_x$, and $y\sim\Bern(q_x)$ with $q_x$ the level's label rate. \Cref{tab:pools} describes the six pools. The threshold keeps each $q_x$ within a standard error of about $0.05$ or less of the level's long-run rate, so that the population is a realistic one; it also removes the real long tail, so sparsity in the experiment comes only from subsampling (\cref{sec:oracle}).

For a pool with $\nlev$ levels and $\bar n\in\{0.25,0.5,1,2,3,5,10,20\}$, a training set is $N=\bar n\,\nlev$ rows (rounded) drawn i.i.d.\ from the population (levels with replacement, labels fresh from $\Bern(q_x)$, not the original labels). Every method is fitted to the training counts and scored by its exact excess cross-entropy $\sum_xp_x\kl(q_x\|r_x)$, with unseen levels receiving the method's prediction for $n_x=0$. The full oracle's prediction for a level with $(n,k)$ is the posterior mean of $q$ over pool levels with weights $p_x\,\mathrm{Poi}(n;Np_x)\,q_x^k(1-q_x)^{n-k}$; the popularity-blind oracle uses the weights $p_x\,q_x^k(1-q_x)^{n-k}$, i.e.\ the row-weighted law of $q$; and the moment-matched beta matches the first two moments of that law. The shuffled control applies one random permutation of the $q_x$ across levels per pool, kept fixed over training sets. All methods are fitted to the same 50 training sets per cell, drawn from one seeded random stream per pool. The level split of the gap (\cref{fig:gap-by-n}) assigns each level's term in the excess cross-entropy to its training count $n_x$.

\begin{table}[h]
\centering\small
\begin{tabular}{lrrrrrc}
\toprule
Pool & source rows & levels kept & rows kept & median rows & $\mu$ & $\mathrm{corr}(\log p,q)$\\
\midrule
MovieLens \texttt{movie\_id} & 1,000,209 & 2,019 & 942,225 & 314 & 0.585 & $+0.41$\\
MovieLens \texttt{user\_id} & 1,000,209 & 2,945 & 847,302 & 212 & 0.562 & $-0.27$\\
Book-Crossing user, $r\ge8$ & 433,671 & 495 & 117,645 & 152 & 0.625 & $-0.03$\\
Book-Crossing user, rated & 1,149,780 & 1,847 & 658,805 & 197 & 0.276 & $-0.24$\\
Avazu \texttt{device\_ip} & 4,000,000 & 2,307 & 710,240 & 152 & 0.178 & $+0.02$\\
Criteo \texttt{C1} & 4,412,639 & 1,982 & 2,886,620 & 210 & 0.030 & $-0.02$\\
\bottomrule
\end{tabular}
\caption{Subsampling-oracle pools: levels with at least 100 rows. $\mu=\sum_xp_xq_x$; the correlation is across levels. The largest level holds $0.3$--$7\%$ of a pool's rows, except Criteo \texttt{C1}, where one level holds $37\%$ (Criteo source rows exclude the rows with \texttt{C1} missing).}\label{tab:pools}
\end{table}

\begin{table}[h]
\centering
\small
\resizebox{\linewidth}{!}{\input{tables/tab_oracle_gaps}}
\caption{Real-data oracle at $\bar n=2$: gap to the full oracle of every method of \cref{tab:oracle}, of fixed $m=20$ and the cross-validated $m$ (``CV $m$''), and of the two references built from the true $q$, in milli-nats (share of the oracle's gain).}\label{tab:oracle-gaps}
\end{table}

\begin{table}[h]
\centering
\small
\resizebox{\linewidth}{!}{\input{tables/tab_bakeoff_full}}
\caption{All 24 single-feature tasks of \cref{sec:bakeoff}. Columns as in \cref{tab:bakeoff}, plus $N$, the number of levels, ``CV $m$'', the $m$-estimate with $m$ tuned by leave-one-out log-loss, and ``$m{=}1$ (CatBoost)'', the $m$-estimate with $m=1$ to which CatBoost's ordered target statistic reduces at prediction time. Holm's correction is over the 11 comparisons with the baseline on each task (\cref{tab:methods}), including the Robbins correction without its doubleton test, which is not shown. scikit-learn's \texttt{TargetEncoder} was also run but is left out: its near-0/1 singleton encodings make its held-out loss a matter of clipping (up to $3.5$ nats per row at a $10^{-12}$ clip); leaving it out changes no significance call. Shares are meaningless where the baseline gains almost nothing over the prior (\texttt{churn}, \texttt{click\_small/user\_id}, \texttt{airlines/Flight}), and are shown as ``n/a'' where that gain is negative (\texttt{churn/pseudo\_id}), since dividing by it flips their sign; \cref{tab:bakeoff-abs} gives these tasks in milli-nats.}\label{tab:bakeoff-full}
\end{table}

\Cref{tab:bakeoff-abs} gives the same comparisons in absolute terms: the difference in held-out log-loss per test row to the baseline, in milli-nats, with a 95\% interval from the corrected resampled $t$-test, Bonferroni-corrected over the 11 comparisons with the baseline on each task, so the intervals of a task hold simultaneously. Where the shares of \cref{tab:bakeoff-full} explode because the baseline gains almost nothing, the absolute differences are small: on \texttt{airlines/Flight} CatBoost's $+3171\%$ is $+70$ milli-nats, and every other method is within $0.3$ milli-nats of the baseline.

\begin{table}[!ht]
\centering\small
\resizebox{\linewidth}{!}{\input{tables/tab_bakeoff_abs}}
\caption{All 24 bakeoff tasks in absolute terms: held-out log-loss of each method minus that of the baseline, in milli-nats per test row (negative is better), $\pm$ the half-width of a 95\% interval from the corrected resampled $t$-test over 25 splits, Bonferroni-corrected over the 11 comparisons with the baseline on each task. Script \texttt{experiments/paper/tab\_bakeoff\_abs.py}.}\label{tab:bakeoff-abs}
\end{table}

\section{Reproducibility}\label{app:repro}

\paragraph{Code.} All methods are implemented in the Python package \texttt{\pkgname{}} (BSD-3-Clause; NumPy, SciPy, pandas and scikit-learn; CatBoost and pyarrow optional). \ifpreprint The code is available at \pkgurl.\else The code is provided as supplementary material and will be released with a tagged version on publication; the link is withheld for review.\fi{} The main entry points are \texttt{fit\_counts(n, method)} on a level-by-class count matrix, with \texttt{method} one of \texttt{"baseline"}, \texttt{"baseline\_unseen"}, \texttt{"baseline\_robbins"}, \texttt{"freq\_prior"}, \texttt{"npmle"}, \texttt{"npmle\_robbins"}, \texttt{"glmm"}, \texttt{"glmm\_mean"}, \texttt{"cv\_m"} and \texttt{"fixed\_m"}; \texttt{mi\_interval}, \texttt{moment\_covariance}, \texttt{identified\_interval} and \texttt{worst\_case\_halfwidth} for \cref{sec:mi}; \texttt{credible\_interval} for the interval of \citet{hernandez2019}; and \texttt{stratum\_moment\_test} and \texttt{fit\_freq\_prior} for the exchangeability tests. A scikit-learn-style \texttt{BayesianTargetEncoder} wraps the encoders with out-of-fold encoding for downstream use.

\paragraph{Figures and tables.} Every figure and table is produced by a script in \texttt{experiments/} of the package; \cref{tab:repro} maps them. Data are not redistributed (\cref{tab:data}); the scripts take their paths as arguments, and \texttt{experiments/README.md} gives the expected layout and the exact commands. The intermediate CSV files (per-split and per-training-set results) are \ifpreprint in the archive \texttt{results.zip} attached to the tagged release of the code\else in the folder \texttt{results/} of the supplementary material and will be released with the code\fi, so that the tables and figures can be rebuilt without rerunning the experiments.

\begin{table}[h]
\centering\small
\resizebox{\linewidth}{!}{%
\begin{tabular}{lll}
\toprule
Output & Script (in \texttt{experiments/}) & Input\\
\midrule
\cref{fig:teaser}, \cref{tab:teaser} & \texttt{paper/fig\_teaser.py} & \texttt{table\_by\_n.csv} of the bakeoff, oracle pools\\
\cref{tab:regret,tab:avg}, MC check of \cref{prop:ml} & \texttt{paper/theory/verify.py}, \texttt{paper/theory/pseudo\_true.py} & none (exact sums)\\
\cref{tab:heavy} & \texttt{paper/theory/pseudo\_true\_heavy.py} & none (exact sums)\\
Real-pool regrets in \cref{sec:numcheck} & \texttt{paper/real\_pool\_regret.py} & oracle pools\\
\cref{fig:shift} & \texttt{paper/fig\_shift.py} & synthetic\\
\cref{tab:id}, \cref{fig:interval} & \texttt{paper/theory/unseen\_mi\_bounds.py}, \texttt{paper/fig\_interval.py} & none (linear programs)\\
\cref{fig:junk} & \texttt{paper/run\_junk\_fraction.py}, \texttt{paper/fig\_junk.py} & synthetic\\
\cref{fig:oracle,fig:gap-by-n}, \cref{tab:oracle,tab:oracle-gaps} & \texttt{realdata\_oracle.py}, \texttt{realdata\_oracle\_tables.py}, \texttt{paper/fig\_oracle.py} & oracle pools\\
\cref{tab:bakeoff,tab:bakeoff-full} & \texttt{realdata\_paired.py}, \texttt{realdata\_tables.py}, \texttt{paper/tab\_bakeoff.py} & 24 tasks\\
\cref{tab:downstream} & \texttt{downstream\_amazon.py} & Amazon\\
\cref{fig:mi-coverage}, \cref{tab:mi-coverage,tab:mi-coverage-full,tab:mi-point} & \texttt{mi\_coverage.py} (and \texttt{-{}-lik-only}), \texttt{mi\_coverage\_tables.py}, \texttt{paper/fig\_mi\_coverage.py}, \texttt{paper/tab\_mi\_point.py} & synthetic, oracle pools\\
\cref{tab:exch-pools,tab:exch-tasks} & \texttt{exchangeability.py}, \texttt{paper/tab\_exch.py} & oracle pools, 24 tasks\\
\cref{tab:largen} & \texttt{paper/large\_n\_regret.py} & none (exact sums)\\
\cref{tab:glmm} & \texttt{paper/glmm\_laplace.py} & OpenML, Avazu\\
\cref{tab:wtl,tab:bakeoff-abs}, fixed-set medians & \texttt{bakeoff\_summary.py}, \texttt{paper/tab\_wtl.py}, \texttt{paper/tab\_bakeoff\_abs.py} & 24 tasks\\
\cref{tab:oracle-sens} & \texttt{realdata\_oracle.py} (\texttt{-{}-min-rows}, \texttt{-{}-tail shifted-head}), \texttt{oracle\_sensitivity.py}, \texttt{paper/tab\_oracle\_sensitivity.py} & oracle pools with tails\\
\cref{tab:mi-classical} & \texttt{mi\_coverage.py -{}-baselines-only}, \texttt{paper/tab\_mi\_classical.py} & synthetic, oracle pools\\
\cref{tab:bakeoff-summary} & \texttt{paper/tab\_bakeoff\_summary.py} & 24 tasks\\
\cref{fig:fixed-m}, \cref{tab:fixed-m} & \texttt{fixed\_m\_sweep.py}, \texttt{paper/fig\_fixed\_m.py} & 24 tasks, oracle pools\\
\cref{tab:mi-bounds} & \texttt{mi\_realdata\_bounds.py}, \texttt{paper/tab\_mi\_bounds.py} & 23 binary tasks\\
\cref{tab:mi-sens} & \texttt{mi\_coverage.py -{}-lik-only -{}-lik-grid -{}-n-cap} & synthetic, oracle pools\\
\bottomrule
\end{tabular}}
\caption{Scripts behind each figure and table.}\label{tab:repro}
\end{table}

\paragraph{Randomness and paired designs.} Every experiment fixes its seeds, and every comparison is paired: in the bakeoff, all methods share the 25 splits of a 5-fold cross-validation repeated with seeds $0$--$4$; in the oracle pools and the synthetic sweeps, all methods are fitted to the same training sets. Standard errors are therefore those of paired differences, as reported in the tables.

\paragraph{Compute.} Everything runs on a CPU; nothing needs a GPU. On a 4-core machine, the synthetic junk-fraction sweep takes about 15 minutes, one oracle pool (8 values of $\bar n$, 50 training sets, real or shuffled) a few minutes, the MI coverage study about 1.5 hours for the synthetic part with 4 workers (one BLAS thread each), and the bakeoff about 1.5 hours on 3 cores, dominated by the GLMM and NPMLE fits on the click logs. The baseline itself fits a feature with $10^6$ levels in about 5 seconds. The exact-sum checks of \cref{sec:numcheck} and \cref{app:gap} take seconds.

%% file: tables/tab_mi_coverage_full.tex
\begin{tabular}{lr ccccccc ccccccc}
\toprule
 & & \multicolumn{7}{c}{$\bar n=1$} & \multicolumn{7}{c}{$\bar n=5$}\\
\cmidrule(lr){3-9}\cmidrule(lr){10-16}
population & $I$ & cred. & width & mom. & width & lik. & width & ident. & cred. & width & mom. & width & lik. & width & ident.\\
\midrule
\multicolumn{16}{l}{\emph{synthetic, $10^5$ equally popular levels}}\\
junk $f{=}0.5$ & 0.184 & 0.94 & 0.012 & 1.00 & 0.098 & 0.95 & 0.028 & 0.013 & 0.44 & 0.003 & 1.00 & 0.038 & 0.98 & 0.004 & 0.013\\
junk $f{=}0.8$ & 0.074 & 0.45 & 0.011 & 1.00 & 0.067 & 0.97 & 0.020 & 0.006 & 0.00 & 0.003 & 1.00 & 0.025 & 0.99 & 0.004 & 0.006\\
atoms $\{.2,.3,.85\}$ & 0.163 & 0.03 & 0.012 & 1.00 & 0.091 & 0.97 & 0.019 & 0.003 & 0.00 & 0.003 & 1.00 & 0.025 & 0.99 & 0.004 & 0.003\\
$\mathrm{Beta}(0.5,0.5)$ & 0.307 & 0.96 & 0.012 & 1.00 & 0.120 & 1.00 & 0.035 & 0.020 & 0.93 & 0.003 & 1.00 & 0.050 & 1.00 & 0.009 & 0.020\\
$\mathrm{Beta}(1,3)$ & 0.104 & 0.96 & 0.011 & 1.00 & 0.080 & 1.00 & 0.027 & 0.014 & 0.96 & 0.003 & 1.00 & 0.031 & 0.98 & 0.006 & 0.014\\
\midrule
\multicolumn{16}{l}{\emph{oracle pools, real}}\\
MovieLens movies & 0.087 & 0.94 & 0.065 & 0.60 & 0.329 & 0.98 & 0.077 & 0.005 & 0.94 & 0.019 & 0.90 & 0.153 & 0.94 & 0.027 & 0.004\\
MovieLens users & 0.064 & 0.94 & 0.053 & 0.94 & 0.127 & 0.92 & 0.062 & 0.003 & 0.90 & 0.014 & 1.00 & 0.087 & 0.96 & 0.017 & 0.003\\
BX users, rated & 0.141 & 0.68 & 0.070 & 0.92 & 0.220 & 0.76 & 0.092 & 0.019 & 0.52 & 0.022 & 0.98 & 0.163 & 1.00 & 0.035 & 0.015\\
BX users, $r\geq8$ & 0.144 & 0.94 & 0.138 & 1.00 & 0.567 & 0.94 & 0.172 & 0.270 & 0.92 & 0.043 & 1.00 & 0.235 & 0.94 & 0.062 & 0.012\\
Avazu device\_ip & 0.040 & 0.42 & 0.041 & 0.90 & 0.114 & 0.98 & 0.058 & 0.010 & 0.20 & 0.019 & 1.00 & 0.086 & 0.98 & 0.019 & 0.008\\
Criteo C1 & 0.004 & 0.70 & 0.023 & 0.66 & 0.040 & 0.98 & 0.054 & 0.007 & 1.00 & 0.017 & 0.52 & 0.012 & 0.94 & 0.024 & 0.002\\
\midrule
\multicolumn{16}{l}{\emph{oracle pools, shuffled $q$}}\\
MovieLens movies & 0.099 & 0.92 & 0.066 & 0.98 & 0.259 & 0.92 & 0.075 & 0.005 & 0.90 & 0.019 & 0.98 & 0.141 & 0.98 & 0.024 & 0.004\\
MovieLens users & 0.061 & 0.92 & 0.053 & 0.98 & 0.205 & 0.96 & 0.061 & 0.003 & 0.94 & 0.015 & 1.00 & 0.093 & 1.00 & 0.017 & 0.003\\
BX users, rated & 0.131 & 0.96 & 0.067 & 0.98 & 0.269 & 0.98 & 0.082 & 0.016 & 0.94 & 0.022 & 1.00 & 0.160 & 0.98 & 0.033 & 0.013\\
BX users, $r\geq8$ & 0.117 & 0.94 & 0.139 & 1.00 & 0.572 & 0.92 & 0.169 & 0.266 & 0.98 & 0.042 & 1.00 & 0.246 & 0.92 & 0.062 & 0.010\\
Avazu device\_ip & 0.046 & 0.82 & 0.048 & 0.90 & 0.106 & 1.00 & 0.054 & 0.011 & 0.70 & 0.019 & 1.00 & 0.087 & 0.94 & 0.019 & 0.008\\
Criteo C1 & 0.017 & 0.02 & 0.023 & 0.82 & 0.057 & 0.98 & 0.060 & 0.018 & 0.00 & 0.011 & 0.46 & 0.020 & 1.00 & 0.022 & 0.006\\
\bottomrule
\end{tabular}

%% file: tables/tab_mi_point.tex
\begin{tabular}{lr rrrrr rrrrr}
\toprule
 & & \multicolumn{5}{c}{$\bar n=1$} & \multicolumn{5}{c}{$\bar n=5$}\\
\cmidrule(lr){3-7}\cmidrule(lr){8-12}
population & $I$ & plug-in & baseline & NPMLE & freq. & midpoint & plug-in & baseline & NPMLE & freq. & midpoint\\
\midrule
\multicolumn{12}{l}{\emph{synthetic, $10^5$ equally popular levels}}\\
junk $f{=}0.5$ & 0.184 & $+332$ \scriptsize(332) & $+0.8$ \scriptsize(3.2) & $+4.1$ \scriptsize(5.9) & $+0.8$ \scriptsize(3.2) & $+13.8$ \scriptsize(15.0) & $+110$ \scriptsize(110) & $+1.8$ \scriptsize(2.0) & $+0.8$ \scriptsize(1.2) & $+1.8$ \scriptsize(2.0) & $+8.6$ \scriptsize(9.2)\\
junk $f{=}0.8$ & 0.074 & $+393$ \scriptsize(393) & $-5.2$ \scriptsize(5.7) & $-0.1$ \scriptsize(3.3) & $-5.2$ \scriptsize(5.7) & $+6.9$ \scriptsize(8.9) & $+115$ \scriptsize(115) & $-4.3$ \scriptsize(4.3) & $+0.5$ \scriptsize(0.8) & $-4.3$ \scriptsize(4.3) & $+4.5$ \scriptsize(4.9)\\
atoms $\{.2,.3,.85\}$ & 0.163 & $+344$ \scriptsize(344) & $+11.3$ \scriptsize(11.7) & $+5.2$ \scriptsize(6.0) & $+11.4$ \scriptsize(11.7) & $+25.4$ \scriptsize(26.1) & $+116$ \scriptsize(116) & $+5.9$ \scriptsize(6.0) & $+0.1$ \scriptsize(0.9) & $+5.8$ \scriptsize(5.8) & $+6.2$ \scriptsize(6.6)\\
$\mathrm{Beta}(0.5,0.5)$ & 0.307 & $+256$ \scriptsize(256) & $-0.3$ \scriptsize(3.1) & $-2.1$ \scriptsize(4.9) & $-0.3$ \scriptsize(3.1) & $+6.3$ \scriptsize(8.5) & $+89.9$ \scriptsize(89.9) & $-0.1$ \scriptsize(0.9) & $-0.5$ \scriptsize(1.1) & $-0.1$ \scriptsize(0.9) & $+2.8$ \scriptsize(4.3)\\
$\mathrm{Beta}(1,3)$ & 0.104 & $+302$ \scriptsize(302) & $+0.2$ \scriptsize(2.7) & $+0.5$ \scriptsize(3.5) & $+0.2$ \scriptsize(2.7) & $+10.7$ \scriptsize(11.8) & $+104$ \scriptsize(104) & $+0.2$ \scriptsize(0.8) & $+0.3$ \scriptsize(0.9) & $+0.2$ \scriptsize(0.8) & $+3.6$ \scriptsize(4.0)\\
\midrule
\multicolumn{12}{l}{\emph{oracle pools, real}}\\
MovieLens movies & 0.087 & $+319$ \scriptsize(319) & $-3.3$ \scriptsize(16.8) & $+1.0$ \scriptsize(17.6) & $-13.1$ \scriptsize(20.9) & $+152$ \scriptsize(157) & $+107$ \scriptsize(107) & $-1.4$ \scriptsize(5.0) & $-1.3$ \scriptsize(5.2) & $-7.0$ \scriptsize(8.5) & $+68.3$ \scriptsize(69.3)\\
MovieLens users & 0.064 & $+348$ \scriptsize(348) & $+4.0$ \scriptsize(13.9) & $+8.5$ \scriptsize(16.4) & $+4.8$ \scriptsize(14.5) & $-0.0$ \scriptsize(18.7) & $+111$ \scriptsize(111) & $-2.6$ \scriptsize(4.5) & $-0.2$ \scriptsize(3.8) & $-0.4$ \scriptsize(3.7) & $-5.5$ \scriptsize(10.8)\\
BX users, rated & 0.141 & $+254$ \scriptsize(254) & $+27.2$ \scriptsize(32.1) & $+32.1$ \scriptsize(36.6) & $+16.6$ \scriptsize(23.4) & $-30.2$ \scriptsize(40.2) & $+102$ \scriptsize(102) & $+11.0$ \scriptsize(12.0) & $+7.8$ \scriptsize(9.3) & $+2.0$ \scriptsize(5.4) & $-15.6$ \scriptsize(20.4)\\
BX users, $r\geq8$ & 0.144 & $+289$ \scriptsize(289) & $+20.2$ \scriptsize(38.2) & $+33.1$ \scriptsize(46.4) & $+20.2$ \scriptsize(38.2) & $+140$ \scriptsize(153) & $+106$ \scriptsize(107) & $+2.1$ \scriptsize(11.2) & $+3.5$ \scriptsize(11.9) & $+2.1$ \scriptsize(11.2) & $+6.6$ \scriptsize(24.9)\\
Avazu device\_ip & 0.040 & $+201$ \scriptsize(201) & $-20.2$ \scriptsize(22.4) & $-1.0$ \scriptsize(11.3) & $-17.7$ \scriptsize(20.8) & $+17.1$ \scriptsize(31.3) & $+84.6$ \scriptsize(84.6) & $-12.9$ \scriptsize(13.4) & $-3.1$ \scriptsize(4.4) & $-9.8$ \scriptsize(11.0) & $+9.0$ \scriptsize(12.0)\\
Criteo C1 & 0.004 & $+39.1$ \scriptsize(39.6) & $+2.0$ \scriptsize(5.8) & $+0.6$ \scriptsize(3.6) & $+2.0$ \scriptsize(5.8) & $+16.0$ \scriptsize(24.6) & $+24.6$ \scriptsize(24.7) & $+2.8$ \scriptsize(4.0) & $-0.2$ \scriptsize(1.6) & $+2.8$ \scriptsize(4.0) & $+1.9$ \scriptsize(4.8)\\
\midrule
\multicolumn{12}{l}{\emph{oracle pools, shuffled $q$}}\\
MovieLens movies & 0.099 & $+319$ \scriptsize(319) & $+3.6$ \scriptsize(17.8) & $+7.6$ \scriptsize(19.2) & $+3.6$ \scriptsize(17.7) & $+34.4$ \scriptsize(42.2) & $+107$ \scriptsize(107) & $+0.2$ \scriptsize(5.0) & $+0.3$ \scriptsize(5.1) & $+0.2$ \scriptsize(5.0) & $+20.7$ \scriptsize(24.5)\\
MovieLens users & 0.061 & $+345$ \scriptsize(346) & $+1.3$ \scriptsize(15.4) & $+5.1$ \scriptsize(16.8) & $+1.2$ \scriptsize(15.2) & $+44.7$ \scriptsize(51.6) & $+113$ \scriptsize(113) & $-0.3$ \scriptsize(3.8) & $+0.5$ \scriptsize(3.7) & $-0.3$ \scriptsize(3.8) & $+19.3$ \scriptsize(21.1)\\
BX users, rated & 0.131 & $+254$ \scriptsize(254) & $+2.3$ \scriptsize(14.9) & $+7.5$ \scriptsize(18.3) & $+2.2$ \scriptsize(14.9) & $+13.0$ \scriptsize(32.8) & $+98.1$ \scriptsize(98.2) & $+1.6$ \scriptsize(5.0) & $+1.9$ \scriptsize(5.1) & $+1.5$ \scriptsize(4.9) & $+6.9$ \scriptsize(14.7)\\
BX users, $r\geq8$ & 0.117 & $+287$ \scriptsize(287) & $-2.4$ \scriptsize(35.2) & $+12.8$ \scriptsize(38.1) & $-2.5$ \scriptsize(34.7) & $+172$ \scriptsize(181) & $+108$ \scriptsize(108) & $-0.4$ \scriptsize(8.9) & $+2.8$ \scriptsize(9.4) & $-0.4$ \scriptsize(8.9) & $+50.6$ \scriptsize(55.6)\\
Avazu device\_ip & 0.046 & $+198$ \scriptsize(198) & $-10.3$ \scriptsize(13.0) & $+1.0$ \scriptsize(8.7) & $-10.1$ \scriptsize(13.1) & $+7.0$ \scriptsize(25.4) & $+84.0$ \scriptsize(84.1) & $-7.2$ \scriptsize(8.0) & $+0.9$ \scriptsize(3.5) & $-7.1$ \scriptsize(7.9) & $+3.5$ \scriptsize(9.5)\\
Criteo C1 & 0.017 & $+39.2$ \scriptsize(39.9) & $+23.0$ \scriptsize(23.6) & $+5.3$ \scriptsize(7.0) & $+18.5$ \scriptsize(21.5) & $+11.5$ \scriptsize(22.3) & $+23.7$ \scriptsize(23.9) & $+18.1$ \scriptsize(18.2) & $+1.2$ \scriptsize(2.6) & $+15.6$ \scriptsize(16.9) & $-7.2$ \scriptsize(9.5)\\
\bottomrule
\end{tabular}

%% file: tables/tab_mi_classical.tex
\begin{tabular}{l rrr rrr}
\toprule
 & \multicolumn{3}{c}{$\bar n=1$} & \multicolumn{3}{c}{$\bar n=5$}\\
\cmidrule(lr){2-4}\cmidrule(lr){5-7}
population & M--M & C--S & NSB & M--M & C--S & NSB\\
\midrule
\multicolumn{7}{l}{\emph{synthetic, $10^5$ equally popular levels}}\\
junk $f{=}0.5$ & $+280$ \scriptsize(280) & $+151$ \scriptsize(151) & $-28.6$ \scriptsize(28.8) & $+48.5$ \scriptsize(48.5) & $-33.8$ \scriptsize(33.9) & $-9.3$ \scriptsize(9.5)\\
junk $f{=}0.8$ & $+326$ \scriptsize(326) & $+116$ \scriptsize(116) & $-12.6$ \scriptsize(12.8) & $+39.9$ \scriptsize(39.9) & $-136$ \scriptsize(136) & $-8.4$ \scriptsize(8.5)\\
atoms $\{.2,.3,.85\}$ & $+291$ \scriptsize(291) & $+163$ \scriptsize(163) & $-17.2$ \scriptsize(17.4) & $+51.5$ \scriptsize(51.5) & $-0.2$ \scriptsize(1.9) & $-3.9$ \scriptsize(3.9)\\
$\mathrm{Beta}(0.5,0.5)$ & $+217$ \scriptsize(217) & $+145$ \scriptsize(145) & $-62.7$ \scriptsize(62.7) & $+43.3$ \scriptsize(43.3) & $+8.4$ \scriptsize(8.6) & $-18.2$ \scriptsize(18.2)\\
$\mathrm{Beta}(1,3)$ & $+256$ \scriptsize(256) & $+155$ \scriptsize(155) & $-39.7$ \scriptsize(39.9) & $+47.9$ \scriptsize(47.9) & $-4.1$ \scriptsize(4.5) & $-13.8$ \scriptsize(14.2)\\
\midrule
\multicolumn{7}{l}{\emph{oracle pools, real}}\\
MovieLens movies & $+252$ \scriptsize(253) & $+95.4$ \scriptsize(97.6) & $+17.7$ \scriptsize(24.1) & $+49.6$ \scriptsize(49.9) & $+14.9$ \scriptsize(16.4) & $+2.3$ \scriptsize(5.2)\\
MovieLens users & $+278$ \scriptsize(279) & $+84.4$ \scriptsize(86.8) & $+9.5$ \scriptsize(16.5) & $+48.2$ \scriptsize(48.4) & $+1.6$ \scriptsize(8.4) & $+2.6$ \scriptsize(4.5)\\
BX users, rated & $+212$ \scriptsize(212) & $+159$ \scriptsize(160) & $+50.1$ \scriptsize(52.3) & $+57.6$ \scriptsize(57.8) & $+31.3$ \scriptsize(32.3) & $+5.0$ \scriptsize(6.9)\\
BX users, $r\geq8$ & $+243$ \scriptsize(244) & $+149$ \scriptsize(153) & $+51.8$ \scriptsize(58.7) & $+54.0$ \scriptsize(55.2) & $-10.4$ \scriptsize(19.8) & $+0.2$ \scriptsize(11.3)\\
Avazu device\_ip & $+167$ \scriptsize(167) & $+154$ \scriptsize(154) & $+38.6$ \scriptsize(39.6) & $+50.5$ \scriptsize(50.6) & $+35.6$ \scriptsize(35.9) & $+5.5$ \scriptsize(6.5)\\
Criteo C1 & $+34.5$ \scriptsize(35.0) & $+27.7$ \scriptsize(28.5) & $+13.9$ \scriptsize(15.0) & $+19.9$ \scriptsize(20.0) & $+2.7$ \scriptsize(3.7) & $+4.8$ \scriptsize(5.2)\\
\midrule
\multicolumn{7}{l}{\emph{oracle pools, shuffled $q$}}\\
MovieLens movies & $+253$ \scriptsize(254) & $+100$ \scriptsize(103) & $+18.6$ \scriptsize(25.1) & $+51.0$ \scriptsize(51.3) & $+21.8$ \scriptsize(23.0) & $+5.4$ \scriptsize(7.3)\\
MovieLens users & $+277$ \scriptsize(277) & $+89.0$ \scriptsize(91.7) & $+11.1$ \scriptsize(19.7) & $+50.4$ \scriptsize(50.5) & $+7.4$ \scriptsize(11.4) & $+4.6$ \scriptsize(5.8)\\
BX users, rated & $+205$ \scriptsize(206) & $+127$ \scriptsize(129) & $+30.8$ \scriptsize(34.4) & $+53.3$ \scriptsize(53.6) & $+32.8$ \scriptsize(33.6) & $+3.6$ \scriptsize(5.9)\\
BX users, $r\geq8$ & $+240$ \scriptsize(241) & $+157$ \scriptsize(161) & $+55.6$ \scriptsize(62.7) & $+56.3$ \scriptsize(57.0) & $-2.7$ \scriptsize(15.5) & $+7.3$ \scriptsize(11.4)\\
Avazu device\_ip & $+164$ \scriptsize(165) & $+145$ \scriptsize(145) & $+33.0$ \scriptsize(34.5) & $+50.3$ \scriptsize(50.4) & $+35.5$ \scriptsize(35.9) & $+3.8$ \scriptsize(5.0)\\
Criteo C1 & $+34.5$ \scriptsize(35.1) & $+16.4$ \scriptsize(17.5) & $+8.8$ \scriptsize(10.5) & $+19.0$ \scriptsize(19.2) & $-1.1$ \scriptsize(3.1) & $+1.7$ \scriptsize(3.1)\\
\bottomrule
\end{tabular}

%% file: tables/tab_mi_sens.tex
\begin{tabular}{l l r r c c r}
\toprule
check (A / B) & cell & $\bar n$ & sets & coverage & width (nats) & max.\ $|\Delta|$ end\\
\midrule
grid 101 / 401 & three atoms & 1 & 25 & 1.00 / 1.00 & 0.049 / 0.049 & 1.8e-05\\
grid 101 / 401 & three atoms & 5 & 25 & 0.88 / 0.88 & 0.012 / 0.013 & 3.0e-05\\
grid 101 / 401 & $\mathrm{Beta}(1,3)$ & 1 & 25 & 0.96 / 0.96 & 0.059 / 0.059 & 1.9e-05\\
grid 101 / 401 & $\mathrm{Beta}(1,3)$ & 5 & 25 & 1.00 / 1.00 & 0.014 / 0.014 & 2.8e-05\\
\addlinespace
atoms on / off grid & three atoms & 1 & 50 & 0.98 / 0.98 & 0.052 / 0.052 & --\\
atoms on / off grid & three atoms & 5 & 50 & 0.94 / 0.96 & 0.013 / 0.012 & --\\
\addlinespace
thinning cap 200 / 1000 & Avazu device IP, shuffled & 5 & 30 & 0.97 / 0.97 & 0.019 / 0.019 & 1.9e-04\\
thinning cap 200 / 1000 & Avazu device IP, shuffled & 20 & 30 & 0.93 / 0.90 & 0.009 / 0.009 & 4.2e-04\\
\bottomrule
\end{tabular}

%% file: tables/tab_exch_pools.tex
\begin{tabular}{l ccc ccc ccc }
\toprule
 & \multicolumn{3}{c}{$\bar n=1$} & \multicolumn{3}{c}{$\bar n=5$} & \multicolumn{3}{c}{$\bar n=20$}\\
\cmidrule(lr){2-4}\cmidrule(lr){5-7}\cmidrule(lr){8-10}
pool & stratum & LR & lik.\ cov. & stratum & LR & lik.\ cov. & stratum & LR & lik.\ cov.\\
\midrule
\multicolumn{10}{l}{\emph{oracle pools, real}}\\
MovieLens movies & 0.96 & 1.00 & 0.98 & 1.00 & 1.00 & 0.94 & 1.00 & 1.00 & 0.12\\
MovieLens users & 0.52 & 0.84 & 0.92 & 1.00 & 1.00 & 0.96 & 1.00 & 1.00 & 0.92\\
BX users, rated & 0.72 & 0.98 & 0.76 & 1.00 & 1.00 & 1.00 & 1.00 & 1.00 & 0.30\\
BX users, $r\geq8$ & 0.04 & 0.02 & 0.94 & 0.08 & 0.00 & 0.94 & 0.00 & 0.00 & 0.78\\
Avazu device\_ip & 0.08 & 0.24 & 0.98 & 0.30 & 0.24 & 0.98 & 0.50 & 0.10 & 0.26\\
Criteo C1 & 0.06 & 0.00 & 0.98 & 0.12 & 0.00 & 0.94 & 0.66 & 0.04 & 0.58\\
\midrule
\multicolumn{10}{l}{\emph{oracle pools, shuffled $q$}}\\
MovieLens movies & 0.16 & 0.16 & 0.92 & 0.10 & 0.00 & 0.98 & 0.06 & 0.00 & 1.00\\
MovieLens users & 0.04 & 0.02 & 0.96 & 0.02 & 0.06 & 1.00 & 0.04 & 0.02 & 0.98\\
BX users, rated & 0.08 & 0.04 & 0.98 & 0.00 & 0.00 & 0.98 & 0.02 & 0.00 & 0.96\\
BX users, $r\geq8$ & 0.02 & 0.00 & 0.92 & 0.02 & 0.08 & 0.92 & 0.04 & 0.10 & 0.90\\
Avazu device\_ip & 0.10 & 0.00 & 1.00 & 0.08 & 0.00 & 0.94 & 0.16 & 0.00 & 0.96\\
Criteo C1 & 0.10 & 0.12 & 0.98 & 0.14 & 0.14 & 1.00 & 0.68 & 0.00 & 0.64\\
\bottomrule
\end{tabular}

%% file: tables/tab_exch_tasks.tex
\begin{tabular}{l rr r r r}
\toprule
feature & $N$ & levels & $p$ stratum & $p$ LR & max.\ dev.\ $m_1$\\
\midrule
airlines/AirportFrom & 27k & 283 & $<10^{-8}$ & $<10^{-9}$ & 0.066\\
airlines/Flight & 27k & 5,857 & $<10^{-5}$ & $<10^{-4}$ & 0.039\\
amazon/MGR\_ID & 33k & 4,243 & 0.088 & 0.362 & 0.016\\
amazon/RESOURCE & 33k & 7,518 & $<10^{-3}$ & 0.067 & 0.013\\
amazon/ROLE\_FAMILY\_DESC & 33k & 2,358 & $<10^{-6}$ & $<10^{-6}$ & 0.025\\
avazu/device\_ip & 4.0M & 1,056,051 & $<10^{-300}$ & $<10^{-300}$ & 0.035\\
avazu/user & 4.0M & 1,549,515 & $<10^{-300}$ & $<10^{-300}$ & 0.068\\
bx/author & 384k & 62,112 & $<10^{-69}$ & $<10^{-28}$ & 0.037\\
bx/isbn & 434k & 185,973 & $<10^{-242}$ & $<10^{-190}$ & 0.095\\
bx/loc & 434k & 21,335 & $<10^{-33}$ & $<10^{-14}$ & 0.046\\
bx/publisher & 384k & 11,575 & $<10^{-17}$ & 0.370 & 0.010\\
bx/user & 434k & 77,805 & $<10^{-91}$ & $<10^{-6}$ & 0.081\\
bx\_rated/isbn & 1.1M & 340,556 & $<10^{-300}$ & $<10^{-47}$ & 0.046\\
bx\_rated/user & 1.1M & 105,283 & $<10^{-300}$ & $<10^{-300}$ & 0.229\\
churn/pseudo\_id & 5k & 4,393 & 0.079 & 0.646 & 0.041\\
churn/state & 5k & 51 & -- & 0.700 & --\\
click\_small/advertiser\_id & 40k & 6,064 & $<10^{-8}$ & $<10^{-13}$ & 0.061\\
click\_small/user\_id & 40k & 30,114 & $<10^{-3}$ & 0.212 & 0.062\\
criteo/C1 & 4.6M & 879,152 & $<10^{-7}$ & 0.012 & 0.002\\
criteo/C20 & 4.6M & 965,565 & $<10^{-7}$ & 0.027 & 0.002\\
criteo/C22 & 4.6M & 796,068 & $<10^{-5}$ & 0.462 & 0.002\\
employee\_salaries/position\_title & 9k & 385 & $<10^{-4}$ & $<10^{-3}$ & 0.231\\
movielens/movie\_id & 1.0M & 3,706 & $<10^{-52}$ & $<10^{-88}$ & 0.184\\
movielens/user\_id & 1.0M & 6,040 & $<10^{-40}$ & $<10^{-67}$ & 0.032\\
\bottomrule
\end{tabular}

%% file: tables/tab_mi_bounds.tex
\begin{tabular}{l r r r r r r r r}
\toprule
feature & $F_1/N$ & pred.\ bound & stratified & combined & singletons & baseline & NPMLE & lik.\ (exch.)\\
\midrule
airlines/AirportFrom & 0.00 & 0.012 & [0.017, 0.019] & [0.017, 0.019] & 0.000 & 0.013 & 0.015 & [0.013, 0.030]\\
airlines/Flight & 0.04 & 0.002 & [0.021, 0.050] & [0.021, 0.050] & 0.024 & 0.016 & 0.017 & [0.013, 0.023]\\
amazon/MGR\_ID & 0.03 & 0.048 & [0.059, 0.085] & [0.059, 0.085] & 0.005 & 0.075 & 0.071 & [0.068, 0.086]\\
amazon/RESOURCE & 0.11 & 0.010 & [0.012, 0.054] & [0.012, 0.054] & 0.029 & 0.031 & 0.026 & [0.025, 0.041]\\
amazon/ROLE\_FAMILY\_DESC & 0.02 & 0.025 & [0.041, 0.057] & [0.041, 0.057] & 0.004 & 0.035 & 0.036 & [0.055, 0.079]\\
avazu/device\_ip & 0.15 & 0.037 & [0.068, 0.163] & [0.068, 0.163] & 0.075 & 0.088 & 0.089 & --\\
avazu/user & 0.23 & 0.047 & [0.083, 0.228] & [0.083, 0.228] & 0.121 & 0.135 & 0.130 & --\\
bx/author & 0.09 & 0.018 & [0.032, 0.100] & [0.032, 0.100] & 0.064 & 0.035 & 0.034 & --\\
bx/isbn & 0.30 & 0.012 & [0.031, 0.242] & [0.031, 0.242] & 0.206 & 0.043 & 0.042 & --\\
bx/loc & 0.02 & 0.046 & [0.056, 0.074] & [0.056, 0.074] & 0.014 & 0.057 & 0.056 & --\\
bx/publisher & 0.01 & 0.009 & [0.020, 0.031] & [0.020, 0.031] & 0.010 & 0.012 & 0.013 & --\\
bx/user & 0.10 & 0.083 & [0.100, 0.181] & [0.100, 0.181] & 0.072 & 0.121 & 0.119 & --\\
bx\_rated/isbn & 0.17 & 0.007 & [0.023, 0.144] & [0.023, 0.144] & 0.117 & 0.021 & 0.023 & --\\
bx\_rated/user & 0.05 & 0.145 & [0.121, 0.166] & [0.145, 0.166] & 0.034 & 0.169 & 0.167 & --\\
churn/pseudo\_id & 0.77 & 0.000 & [0.004, 0.333] & [0.004, 0.333] & 0.314 & 0.018 & 0.021 & [0.000, 0.084]\\
churn/state & 0.00 & 0.002 & [0.005, 0.006] & [0.005, 0.006] & 0.000 & 0.004 & 0.005 & [0.002, 0.073]\\
click\_small/advertiser\_id & 0.07 & 0.016 & [0.025, 0.057] & [0.025, 0.057] & 0.026 & 0.016 & 0.027 & [0.020, 0.031]\\
click\_small/user\_id & 0.75 & 0.003 & [0.003, 0.451] & [0.003, 0.451] & 0.358 & 0.099 & 0.091 & [0.042, 0.214]\\
criteo/C1 & 0.16 & 0.003 & [0.000, 0.027] & [0.003, 0.027] & 0.026 & 0.010 & 0.006 & --\\
criteo/C20 & 0.18 & 0.003 & [0.000, 0.030] & [0.003, 0.030] & 0.028 & 0.010 & 0.007 & --\\
criteo/C22 & 0.14 & 0.003 & [0.000, 0.024] & [0.003, 0.024] & 0.023 & 0.009 & 0.006 & --\\
movielens/movie\_id & 0.00 & 0.088 & [0.087, 0.091] & [0.088, 0.091] & 0.000 & 0.090 & 0.090 & --\\
movielens/user\_id & 0.00 & 0.059 & [0.068, 0.071] & [0.068, 0.071] & 0.000 & 0.062 & 0.062 & --\\
\bottomrule
\end{tabular}

%% file: tables/tab_downstream.tex
\begin{tabular}{l rr rr rr rr}
\toprule
 & \multicolumn{4}{c}{raw (8 columns)} & \multicolumn{4}{c}{raw + pairs (36 columns)} \\
\cmidrule(lr){2-5}\cmidrule(lr){6-9}
 & \multicolumn{2}{c}{logistic} & \multicolumn{2}{c}{boosting} & \multicolumn{2}{c}{logistic} & \multicolumn{2}{c}{boosting} \\
encoder & log-loss & $\Delta$ & log-loss & $\Delta$ & log-loss & $\Delta$ & log-loss & $\Delta$ \\
\midrule
baseline & 0.1609 & & 0.1583 & & 0.1424 & & 0.1421 & \\
NPMLE & 0.1597 & $-1.2^{*}$ & & & 0.1405 & $-1.9^{*}$ & 0.1430 & $+0.9$ \\
freq.\ prior & 0.1607 & $-0.2^{*}$ & & & 0.1411 & $-1.3^{*}$ & 0.1420 & $-0.1$ \\
GLMM mode & 0.1611 & $+0.2$ & & & 0.1469 & $+4.5^{*}$ & 0.1473 & $+5.2^{*}$ \\
GLMM mean & 0.1599 & $-1.0^{*}$ & & & 0.1411 & $-1.3^{*}$ & 0.1414 & $-0.8$ \\
fixed $m=20$ & 0.1624 & $+1.4^{*}$ & & & 0.1459 & $+3.5^{*}$ & 0.1423 & $+0.1$ \\
$m{=}1$ (CatBoost) & 0.1635 & $+2.6^{*}$ & & & 0.1431 & $+0.7$ & 0.1415 & $-0.6$ \\
\midrule
baseline + count & 0.1602 & $-0.7$ & 0.1534 & $-4.9^{*}$ & 0.1391 & $-3.3^{*}$ & 0.1403 & $-1.8^{*}$ \\
NPMLE + count & \textbf{0.1591} & $-1.8^{*}$ & \textbf{0.1534} & $-4.9^{*}$ & \textbf{0.1379} & $-4.5^{*}$ & \textbf{0.1399} & $-2.2^{*}$ \\
\midrule
one-hot logistic & 0.1558 & & & & & & & \\
native CatBoost & \multicolumn{8}{l}{0.1366 (raw columns; builds its own combinations)} \\
\bottomrule
\end{tabular}

%% file: tables/tab_oracle_sensitivity.tex
\begin{tabular}{ll r r r r rrrrrr}
\toprule
pool & population & levels & corr & baseline gap & at $n_x\le1$ & closed: +unseen & +Robbins & freq.\ prior & NPMLE & NPMLE +R & GLMM mean\\
\midrule
MovieLens movies & head $\geq100$ & 2,019 & +0.41 & 7.9 & 64\% & 43 $\pm$ 1\% & 59 $\pm$ 2\% & 81 $\pm$ 2\% & 14 $\pm$ 2\% & 41 $\pm$ 2\% & 22 $\pm$ 1\%\\
 & head $\geq50$ & 2,514 & +0.39 & 8.6 & 67\% & 45 $\pm$ 1\% & 63 $\pm$ 2\% & 83 $\pm$ 2\% & 19 $\pm$ 2\% & -- & --\\
 & head + tail & 3,706 & +0.35 & 8.7 & 68\% & 45 $\pm$ 2\% & 65 $\pm$ 2\% & 85 $\pm$ 3\% & 28 $\pm$ 2\% & 59 $\pm$ 3\% & 35 $\pm$ 1\%\\
\addlinespace
MovieLens users & head $\geq100$ & 2,945 & -0.27 & 2.1 & 57\% & 32 $\pm$ 2\% & 36 $\pm$ 4\% & 81 $\pm$ 2\% & -9 $\pm$ 3\% & 9 $\pm$ 6\% & 12 $\pm$ 1\%\\
 & head $\geq50$ & 4,297 & -0.26 & 2.6 & 63\% & 38 $\pm$ 1\% & 53 $\pm$ 2\% & 87 $\pm$ 2\% & -7 $\pm$ 3\% & -- & --\\
 & head + tail & 6,040 & -0.22 & 2.6 & 61\% & 35 $\pm$ 2\% & 56 $\pm$ 2\% & 88 $\pm$ 2\% & 0 $\pm$ 4\% & 28 $\pm$ 5\% & 19 $\pm$ 2\%\\
\addlinespace
Book-Crossing users, rated & head $\geq100$ & 1,847 & -0.24 & 6.1 & 47\% & 28 $\pm$ 1\% & 34 $\pm$ 2\% & 50 $\pm$ 2\% & 21 $\pm$ 2\% & 35 $\pm$ 2\% & 23 $\pm$ 1\%\\
 & head $\geq50$ & 3,427 & -0.27 & 7.2 & 60\% & 42 $\pm$ 1\% & 55 $\pm$ 2\% & 68 $\pm$ 2\% & 31 $\pm$ 2\% & -- & --\\
 & head + tail & 105,283 & -0.21 & 13.2 & 86\% & 68 $\pm$ 0\% & 85 $\pm$ 0\% & 96 $\pm$ 0\% & 87 $\pm$ 0\% & 95 $\pm$ 0\% & 85 $\pm$ 0\%\\
\addlinespace
Book-Crossing users, $r\geq8$ & head $\geq100$ & 495 & -0.03 & 3.9 & 16\% & -9 $\pm$ 3\% & -9 $\pm$ 3\% & 0 $\pm$ 0\% & -72 $\pm$ 16\% & -79 $\pm$ 17\% & -1 $\pm$ 2\%\\
 & head $\geq50$ & 1,332 & +0.02 & 2.3 & 17\% & -10 $\pm$ 3\% & -12 $\pm$ 4\% & 0 $\pm$ 0\% & -69 $\pm$ 21\% & -- & --\\
 & head + tail & 77,805 & -0.02 & 1.2 & 40\% & 21 $\pm$ 1\% & 39 $\pm$ 1\% & 57 $\pm$ 1\% & 70 $\pm$ 1\% & 71 $\pm$ 2\% & 49 $\pm$ 1\%\\
\addlinespace
Avazu device\_ip & head $\geq100$ & 2,307 & +0.02 & 2.2 & 12\% & 0 $\pm$ 1\% & -1 $\pm$ 1\% & -1 $\pm$ 1\% & 27 $\pm$ 5\% & 24 $\pm$ 5\% & -5 $\pm$ 2\%\\
 & head $\geq50$ & 6,701 & -0.02 & 1.4 & 3\% & -1 $\pm$ 0\% & -2 $\pm$ 1\% & -1 $\pm$ 1\% & 43 $\pm$ 3\% & -- & --\\
 & head + tail & 1,056,051 & -0.14 & 1.3 & 23\% & 18 $\pm$ 0\% & 23 $\pm$ 0\% & 28 $\pm$ 0\% & 67 $\pm$ 1\% & 85 $\pm$ 1\% & -40 $\pm$ 0\%\\
\addlinespace
Criteo C1 & head $\geq100$ & 1,982 & -0.02 & 0.7 & 12\% & -11 $\pm$ 4\% & -14 $\pm$ 5\% & 0 $\pm$ 0\% & -17 $\pm$ 9\% & -34 $\pm$ 10\% & 9 $\pm$ 4\%\\
 & head $\geq50$ & 3,893 & +0.00 & 0.5 & 14\% & -8 $\pm$ 3\% & -11 $\pm$ 5\% & 0 $\pm$ 0\% & -3 $\pm$ 7\% & -- & --\\
 & head + tail & 879,151 & -0.01 & 0.1 & 57\% & 39 $\pm$ 1\% & 56 $\pm$ 2\% & 29 $\pm$ 12\% & 67 $\pm$ 3\% & 67 $\pm$ 3\% & 59 $\pm$ 2\%\\
\bottomrule
\end{tabular}

%% file: tables/tab_wtl.tex
\begin{tabular}{lcccccccccc}
\toprule
 & baseline & +unseen & +Robbins & freq. & NPMLE & NPMLE+R & GLMM mode & GLMM mean & CV $m$ & $m{=}20$\\
\midrule
baseline & -- & 0/10 & 0/11 & 0/8 & 0/13 & 0/13 & 6/8 & 1/15 & 0/3 & 15/1\\
+unseen & 10/0 & -- & 0/9 & 4/4 & 2/7 & 0/10 & 9/5 & 4/9 & 10/0 & 17/0\\
+Robbins & 11/0 & 9/0 & -- & 5/2 & 4/5 & 1/7 & 9/3 & 4/7 & 11/0 & 17/0\\
freq. & 8/0 & 4/4 & 2/5 & -- & 4/3 & 1/7 & 10/0 & 5/3 & 7/0 & 16/0\\
NPMLE & 13/0 & 7/2 & 5/4 & 3/4 & -- & 0/7 & 11/0 & 4/0 & 13/0 & 16/0\\
NPMLE+R & 13/0 & 10/0 & 7/1 & 7/1 & 7/0 & -- & 11/0 & 7/0 & 13/0 & 17/0\\
GLMM mode & 8/6 & 5/9 & 3/9 & 0/10 & 0/11 & 0/11 & -- & 0/9 & 6/8 & 10/4\\
GLMM mean & 15/1 & 9/4 & 7/4 & 3/5 & 0/4 & 0/7 & 9/0 & -- & 13/2 & 17/1\\
CV $m$ & 3/0 & 0/10 & 0/11 & 0/7 & 0/13 & 0/13 & 8/6 & 2/13 & -- & 16/0\\
$m{=}20$ & 1/15 & 0/17 & 0/17 & 0/16 & 0/16 & 0/17 & 4/10 & 1/17 & 0/16 & --\\
\bottomrule
\end{tabular}

%% file: tables/tab_bakeoff.tex
\begin{tabular}{l rr rrrrrrrrrr}
\toprule
feature & $\frac{F_1+2F_2}{N}$ & gain & +unseen & +Robbins & NPMLE & NPMLE +Robbins & freq.\ prior & GLMM mode & GLMM mean & $m{=}20$ & \begin{tabular}[b]{@{}c@{}}NPMLE \\ at $n_x{=}0$\end{tabular} & \begin{tabular}[b]{@{}c@{}}freq. \\ at $n_x{=}0$\end{tabular} \\
\midrule
Avazu user (ip+model) & 0.37 & 0.0458 & -2.6$^{*}$ & -3.2$^{*}$ & -2.6$^{*}$ & \textbf{-3.3$^{*}$} & -2.8$^{*}$ & +31.1$^{*}$ & -1.3$^{*}$ & +41.0$^{*}$ & 79\% & 92\% \\
Avazu device\_ip & 0.24 & 0.0363 & -1.4$^{*}$ & -2.2$^{*}$ & -1.9$^{*}$ & \textbf{-3.1$^{*}$} & -0.8$^{*}$ & +16.0$^{*}$ & +0.3$^{*}$ & +31.2$^{*}$ & 36\% & 135\% \\
Criteo C1 & 0.19 & 0.0028 & -1.8$^{*}$ & -2.0$^{*}$ & -2.9$^{*}$ & -3.0$^{*}$ & -1.8$^{*}$ & +11.4$^{*}$ & \textbf{-3.0$^{*}$} & +3.4$^{*}$ & 61\% & 77\% \\
BX user, $r\geq8$ & 0.15 & 0.0827 & -0.1$^{*}$ & -0.3$^{*}$ & -0.7$^{*}$ & \textbf{-0.8$^{*}$} & -0.5$^{*}$ & -0.4$^{*}$ & -0.5$^{*}$ & +13.1$^{*}$ & 2\% & -16\% \\
BX user, rated & 0.07 & 0.1356 & -4.8$^{*}$ & -5.8$^{*}$ & -6.9$^{*}$ & \textbf{-7.2$^{*}$} & -7.1$^{*}$ & -6.4$^{*}$ & -6.5$^{*}$ & +9.5$^{*}$ & 66\% & 69\% \\
BX ISBN, $r\geq8$ & 0.42 & 0.0105 & -9.2$^{*}$ & \textbf{-10.7$^{*}$} & -5.0$^{*}$ & -9.1$^{*}$ & -10.6$^{*}$ & -3.3$^{*}$ & -4.6$^{*}$ & +5.3$^{*}$ & 112\% & 86\% \\
Amazon MGR\_ID & 0.06 & 0.0468 & +0.1 & +0.1 & -1.6 & -1.5 & +0.0 & +10.4$^{*}$ & \textbf{-1.9$^{*}$} & +22.6$^{*}$ & -- & -- \\
Amazon RESOURCE & 0.20 & 0.0094 & -1.8 & -3.0 & -8.9 & -9.3 & +0.1 & +14.7$^{*}$ & -8.2$^{*}$ & -0.1 & -- & -- \\
employee\_salaries$^\dagger$ & 0.02 & 0.6154 & -0.4 & -0.5 & -0.7$^{*}$ & -0.8 & -0.6$^{*}$ & +0.7 & -0.4 & +16.2$^{*}$ & 32\% & 47\% \\
\bottomrule
\end{tabular}

%% file: tables/tab_fixed_m.tex
\begin{tabular}{l r r r r r r}
\toprule
feature & $\hat s$ & CV $m$ & $m=20$ (mnats) & $m=20$ (\%) & $m=1$ (mnats) & $m=1$ (\%)\\
\midrule
airlines/AirportFrom & 25.8 & 20.5 & -0.06 & -1 & +2.96 & +27\\
airlines/Flight & 29.7 & 28.5 & +0.12 & -- & +69.80 & --\\
amazon/MGR\_ID & 3.55 & 3.77 & +10.58 & +23 & +6.71 & +14\\
amazon/RESOURCE & 17.2 & 15.8 & -0.01 & -0 & +17.53 & +186\\
amazon/ROLE\_FAMILY\_DESC & 4.38 & 3.29 & +4.81 & +19 & +1.60 & +6\\
avazu/device\_ip & 3.83 & 2.97 & +11.32 & +31 & +6.55 & +18\\
avazu/user & 2.73 & 2.53 & +18.79 & +41 & +7.18 & +16\\
bx/author & 13.1 & 10.4 & +0.54 & +3 & +15.55 & +90\\
bx/isbn & 12.3 & 11.5 & +0.56 & +5 & +31.55 & +300\\
bx/loc & 7.16 & 6.22 & +2.00 & +4 & +5.19 & +11\\
bx/publisher & 19.6 & 12 & +0.00 & +0 & +3.57 & +40\\
bx/user & 3.94 & 3.83 & +10.82 & +13 & +9.41 & +11\\
bx\_rated/isbn & 21.4 & 17.5 & -0.03 & -1 & +33.08 & +553\\
bx\_rated/user & 2.74 & 2.53 & +12.87 & +9 & +2.90 & +2\\
churn/pseudo\_id & 34.6 & 34.7 & -0.47 & -- & +12.46 & --\\
churn/state & 109 & 106 & +1.19 & -- & +3.46 & --\\
click\_small/advertiser\_id & 15.2 & 17.2 & -0.01 & -0 & +19.58 & +133\\
click\_small/user\_id & 3.62 & 2.88 & +0.01 & -- & +0.08 & --\\
criteo/C1 & 43.3 & 39.3 & +0.09 & +3 & +4.41 & +157\\
criteo/C20 & 42.8 & 38.8 & +0.09 & +3 & +4.38 & +160\\
criteo/C22 & 43.7 & 39.8 & +0.10 & +4 & +4.51 & +154\\
employee\_salaries/position\_title & 0.922 & 0.798 & +99.45 & +16 & +0.17 & +0\\
movielens/movie\_id & 3.92 & 3.43 & +1.27 & +1 & +0.14 & +0\\
movielens/user\_id & 7.58 & 7.55 & +0.61 & +1 & +0.49 & +1\\
\bottomrule
\end{tabular}

%% file: tables/tab_oracle_gaps.tex
\begin{tabular}{l r r r r r r r r r r }
\toprule
pool & baseline & +unseen & +Robbins & freq.\ prior & NPMLE & NPMLE +Robbins & $m{=}20$ & CV $m$ & matched beta & blind oracle \\
\midrule
MovieLens movies & 7.9 (24\%) & 4.5 (14\%) & 3.3 (10\%) & 1.5 (5\%) & 6.8 (21\%) & 4.6 (14\%) & 15.9 (49\%) & 7.9 (25\%) & 7.7 (24\%) & 7.3 (23\%) \\
MovieLens users & 2.1 (12\%) & 1.4 (8\%) & 1.4 (7\%) & 0.4 (2\%) & 2.3 (13\%) & 1.9 (10\%) & 6.3 (35\%) & 2.1 (12\%) & 1.9 (10\%) & 1.9 (10\%) \\
Book-Crossing users, rated & 6.1 (8\%) & 4.4 (6\%) & 4.0 (5\%) & 3.0 (4\%) & 4.8 (6\%) & 3.9 (5\%) & 31.0 (41\%) & 5.8 (8\%) & 5.7 (7\%) & 4.2 (6\%) \\
Book-Crossing users, $r\geq8$ & 3.9 (5\%) & 4.2 (6\%) & 4.2 (6\%) & 3.9 (5\%) & 6.7 (9\%) & 7.0 (10\%) & 29.4 (41\%) & 3.9 (5\%) & 2.8 (4\%) & 1.3 (2\%) \\
Avazu device\_ip & 2.2 (22\%) & 2.2 (22\%) & 2.2 (22\%) & 2.2 (22\%) & 1.6 (16\%) & 1.7 (17\%) & 2.5 (26\%) & 2.1 (22\%) & 2.0 (20\%) & 1.1 (11\%) \\
Criteo C1 & 0.7 (59\%) & 0.8 (66\%) & 0.9 (68\%) & 0.7 (59\%) & 0.9 (69\%) & 1.0 (79\%) & 1.2 (92\%) & 0.8 (66\%) & 0.4 (34\%) & 0.3 (27\%) \\
\bottomrule
\end{tabular}

%% file: tables/tab_bakeoff_full.tex
\begin{tabular}{l rrrr rrrrrrrrrr}
\toprule
feature & $N$ & levels & $\frac{F_1+2F_2}{N}$ & gain & +unseen & +Robbins & NPMLE & NPMLE +Robbins & freq.\ prior & GLMM mode & GLMM mean & $m{=}20$ & CV $m$ & $m{=}1$ (CatBoost) \\
\midrule
airlines/AirportFrom & 27k & 283 & 0.00 & 0.0111 & -0.8 & -0.9 & -3.1 & -3.7 & -7.7 & -4.7$^{*}$ & -4.5$^{*}$ & -0.6 & -0.0 & +26.7 \\
airlines/Flight & 27k & 5,857 & 0.11 & 0.0022 & -5.8 & -4.4 & +4.9 & +1.0 & -12.1 & -0.9 & -0.2 & +5.7 & -0.2 & +3171$^{*}$ \\
amazon/MGR\_ID & 33k & 4,243 & 0.06 & 0.0468 & +0.1 & +0.1 & -1.6 & -1.5 & +0.0 & +10.4$^{*}$ & \textbf{-1.9$^{*}$} & +22.6$^{*}$ & -0.0 & +14.3$^{*}$ \\
amazon/RESOURCE & 33k & 7,518 & 0.20 & 0.0094 & -1.8 & -3.0 & -8.9 & -9.3 & +0.1 & +14.7$^{*}$ & -8.2$^{*}$ & -0.1 & +0.2 & +186$^{*}$ \\
amazon/ROLE\_FAMILY\_DESC & 33k & 2,358 & 0.04 & 0.0248 & +0.1 & -0.3 & -0.5 & -0.3 & -1.4$^{*}$ & +3.7 & -1.6 & +19.4$^{*}$ & -0.7 & +6.5 \\
avazu/device\_ip & 4.0M & 1,056,051 & 0.24 & 0.0363 & -1.4$^{*}$ & -2.2$^{*}$ & -1.9$^{*}$ & \textbf{-3.1$^{*}$} & -0.8$^{*}$ & +16.0$^{*}$ & +0.3$^{*}$ & +31.2$^{*}$ & -0.9$^{*}$ & +18.1$^{*}$ \\
avazu/user & 4.0M & 1,549,515 & 0.37 & 0.0458 & -2.6$^{*}$ & -3.2$^{*}$ & -2.6$^{*}$ & \textbf{-3.3$^{*}$} & -2.8$^{*}$ & +31.1$^{*}$ & -1.3$^{*}$ & +41.0$^{*}$ & -0.1$^{*}$ & +15.7$^{*}$ \\
bx/author & 384k & 62,112 & 0.14 & 0.0173 & -1.8$^{*}$ & -2.9$^{*}$ & -2.4$^{*}$ & \textbf{-3.4$^{*}$} & -2.6$^{*}$ & -2.2$^{*}$ & -2.2$^{*}$ & +3.1$^{*}$ & -0.5 & +89.7$^{*}$ \\
bx/isbn & 434k & 185,973 & 0.42 & 0.0105 & -9.2$^{*}$ & \textbf{-10.7$^{*}$} & -5.0$^{*}$ & -9.1$^{*}$ & -10.6$^{*}$ & -3.3$^{*}$ & -4.6$^{*}$ & +5.3$^{*}$ & -0.1 & +300$^{*}$ \\
bx/loc & 434k & 21,335 & 0.03 & 0.0459 & -0.1$^{*}$ & -0.2$^{*}$ & -0.5$^{*}$ & \textbf{-0.5$^{*}$} & -0.4$^{*}$ & -0.3$^{*}$ & -0.3$^{*}$ & +4.3$^{*}$ & -0.1 & +11.3$^{*}$ \\
bx/publisher & 384k & 11,575 & 0.02 & 0.0088 & +0.0 & -0.3 & -0.4 & -0.6 & +0.0 & +0.0 & +0.1$^{*}$ & +0.0 & -0.9 & +40.4$^{*}$ \\
bx/user & 434k & 77,805 & 0.15 & 0.0827 & -0.1$^{*}$ & -0.3$^{*}$ & -0.7$^{*}$ & \textbf{-0.8$^{*}$} & -0.5$^{*}$ & -0.4$^{*}$ & -0.5$^{*}$ & +13.1$^{*}$ & -0.0 & +11.4$^{*}$ \\
bx\_rated/isbn & 1.1M & 340,556 & 0.28 & 0.0060 & -8.7$^{*}$ & -13.4$^{*}$ & -0.5 & \textbf{-15.1$^{*}$} & -2.2$^{*}$ & +0.6 & -0.1 & -0.6$^{*}$ & -1.0$^{*}$ & +553$^{*}$ \\
bx\_rated/user & 1.1M & 105,283 & 0.07 & 0.1356 & -4.8$^{*}$ & -5.8$^{*}$ & -6.9$^{*}$ & \textbf{-7.2$^{*}$} & -7.1$^{*}$ & -6.4$^{*}$ & -6.5$^{*}$ & +9.5$^{*}$ & -0.0 & +2.1$^{*}$ \\
churn/state & 5k & 51 & 0.00 & 0.0018 & +0.0 & +0.0 & +36.4 & +36.4 & +0.0 & +3.1 & +2.8 & +65.1 & -0.3 & +189 \\
churn/pseudo\_id & 5k & 4,393 & 0.96 & -0.0003 & n/a & n/a & n/a & n/a & n/a & n/a & n/a & n/a & n/a & n/a$^{*}$ \\
click\_small/advertiser\_id & 40k & 6,064 & 0.13 & 0.0147 & -4.9 & -6.2 & -6.7$^{*}$ & \textbf{-8.7$^{*}$} & -7.2$^{*}$ & -5.5 & -4.8$^{*}$ & -0.1 & -0.1 & +133$^{*}$ \\
click\_small/user\_id & 40k & 30,114 & 0.76 & 0.0022 & -30.2$^{*}$ & \textbf{-30.2$^{*}$} & -30.2$^{*}$ & -29.9$^{*}$ & -1.5 & +835 & -29.3$^{*}$ & +0.2 & -0.8 & +3.6 \\
criteo/C1 & 4.6M & 879,152 & 0.19 & 0.0028 & -1.8$^{*}$ & -2.0$^{*}$ & -2.9$^{*}$ & -3.0$^{*}$ & -1.8$^{*}$ & +11.4$^{*}$ & \textbf{-3.0$^{*}$} & +3.4$^{*}$ & -0.1 & +157$^{*}$ \\
criteo/C20 & 4.6M & 965,565 & 0.21 & 0.0027 & -1.9$^{*}$ & -2.0$^{*}$ & -3.1$^{*}$ & -3.1$^{*}$ & -1.7 & +12.9$^{*}$ & \textbf{-3.1$^{*}$} & +3.3$^{*}$ & -0.1 & +160$^{*}$ \\
criteo/C22 & 4.6M & 796,068 & 0.17 & 0.0029 & -1.3$^{*}$ & -1.4$^{*}$ & -2.5$^{*}$ & -2.5$^{*}$ & +0.0 & +9.5$^{*}$ & \textbf{-2.5$^{*}$} & +3.5$^{*}$ & -0.1 & +154$^{*}$ \\
employee\_salaries/position\_title & 9k & 385 & 0.02 & 0.6154 & -0.4 & -0.5 & -0.7$^{*}$ & -0.8 & -0.6$^{*}$ & +0.7 & -0.4 & +16.2$^{*}$ & -0.0 & +0.0 \\
movielens/movie\_id & 1.0M & 3,706 & 0.00 & 0.0879 & -0.0 & -0.0 & -0.1$^{*}$ & -0.1$^{*}$ & \textbf{-0.1$^{*}$} & -0.1$^{*}$ & -0.1$^{*}$ & +1.4$^{*}$ & -0.0 & +0.2$^{*}$ \\
movielens/user\_id & 1.0M & 6,040 & 0.00 & 0.0587 & +0.0 & +0.0 & -0.0 & -0.0 & \textbf{-0.1$^{*}$} & -0.1$^{*}$ & -0.1$^{*}$ & +1.0$^{*}$ & +0.0 & +0.8$^{*}$ \\
\bottomrule
\end{tabular}

%% file: tables/tab_bakeoff_abs.tex
\begin{tabular}{l rrrrrrrrrr}
\toprule
feature & +unseen & +Robbins & NPMLE & NPMLE +Robbins & freq.\ prior & GLMM mode & GLMM mean & $m{=}20$ & CV $m$ & $m{=}1$ (CatBoost)\\
\midrule
airlines/AirportFrom & $-0.09$\,{\scriptsize$\pm$0.33} & $-0.10$\,{\scriptsize$\pm$0.33} & $-0.34$\,{\scriptsize$\pm$0.76} & $-0.41$\,{\scriptsize$\pm$0.78} & $-0.85$\,{\scriptsize$\pm$0.79} & $-0.52$\,{\scriptsize$\pm$0.31} & $-0.50$\,{\scriptsize$\pm$0.30} & $-0.06$\,{\scriptsize$\pm$0.24} & $-0.01$\,{\scriptsize$\pm$0.26} & $+2.96$\,{\scriptsize$\pm$2.56}\\
airlines/Flight & $-0.13$\,{\scriptsize$\pm$0.39} & $-0.10$\,{\scriptsize$\pm$0.43} & $+0.11$\,{\scriptsize$\pm$0.29} & $+0.02$\,{\scriptsize$\pm$0.55} & $-0.27$\,{\scriptsize$\pm$0.52} & $-0.02$\,{\scriptsize$\pm$0.04} & $-0.01$\,{\scriptsize$\pm$0.01} & $+0.12$\,{\scriptsize$\pm$0.52} & $-0.00$\,{\scriptsize$\pm$0.07} & $+69.8$\,{\scriptsize$\pm$9.4}\\
amazon/MGR\_ID & $+0.03$\,{\scriptsize$\pm$0.07} & $+0.04$\,{\scriptsize$\pm$0.09} & $-0.74$\,{\scriptsize$\pm$0.76} & $-0.72$\,{\scriptsize$\pm$0.76} & $+0.00$\,{\scriptsize$\pm$0.00} & $+4.88$\,{\scriptsize$\pm$2.57} & $-0.87$\,{\scriptsize$\pm$0.66} & $+10.6$\,{\scriptsize$\pm$3.3} & $-0.02$\,{\scriptsize$\pm$0.20} & $+6.71$\,{\scriptsize$\pm$2.50}\\
amazon/RESOURCE & $-0.17$\,{\scriptsize$\pm$0.31} & $-0.29$\,{\scriptsize$\pm$0.44} & $-0.84$\,{\scriptsize$\pm$1.58} & $-0.88$\,{\scriptsize$\pm$1.63} & $+0.01$\,{\scriptsize$\pm$0.12} & $+1.39$\,{\scriptsize$\pm$1.07} & $-0.77$\,{\scriptsize$\pm$0.45} & $-0.01$\,{\scriptsize$\pm$0.19} & $+0.01$\,{\scriptsize$\pm$0.12} & $+17.5$\,{\scriptsize$\pm$5.0}\\
amazon/ROLE\_FAMILY\_DESC & $+0.02$\,{\scriptsize$\pm$0.10} & $-0.07$\,{\scriptsize$\pm$0.40} & $-0.12$\,{\scriptsize$\pm$0.56} & $-0.08$\,{\scriptsize$\pm$0.59} & $-0.35$\,{\scriptsize$\pm$0.29} & $+0.92$\,{\scriptsize$\pm$1.52} & $-0.40$\,{\scriptsize$\pm$0.41} & $+4.81$\,{\scriptsize$\pm$1.78} & $-0.17$\,{\scriptsize$\pm$0.46} & $+1.60$\,{\scriptsize$\pm$1.91}\\
avazu/device\_ip & $-0.49$\,{\scriptsize$\pm$0.06} & $-0.79$\,{\scriptsize$\pm$0.06} & $-0.69$\,{\scriptsize$\pm$0.09} & $-1.11$\,{\scriptsize$\pm$0.11} & $-0.30$\,{\scriptsize$\pm$0.03} & $+5.82$\,{\scriptsize$\pm$0.18} & $+0.13$\,{\scriptsize$\pm$0.04} & $+11.3$\,{\scriptsize$\pm$0.2} & $-0.32$\,{\scriptsize$\pm$0.06} & $+6.55$\,{\scriptsize$\pm$0.30}\\
avazu/user & $-1.20$\,{\scriptsize$\pm$0.09} & $-1.47$\,{\scriptsize$\pm$0.08} & $-1.21$\,{\scriptsize$\pm$0.05} & $-1.51$\,{\scriptsize$\pm$0.08} & $-1.30$\,{\scriptsize$\pm$0.09} & $+14.3$\,{\scriptsize$\pm$0.3} & $-0.61$\,{\scriptsize$\pm$0.06} & $+18.8$\,{\scriptsize$\pm$0.3} & $-0.04$\,{\scriptsize$\pm$0.02} & $+7.18$\,{\scriptsize$\pm$0.23}\\
bx/author & $-0.31$\,{\scriptsize$\pm$0.14} & $-0.50$\,{\scriptsize$\pm$0.17} & $-0.42$\,{\scriptsize$\pm$0.12} & $-0.59$\,{\scriptsize$\pm$0.18} & $-0.46$\,{\scriptsize$\pm$0.17} & $-0.38$\,{\scriptsize$\pm$0.07} & $-0.39$\,{\scriptsize$\pm$0.09} & $+0.54$\,{\scriptsize$\pm$0.13} & $-0.08$\,{\scriptsize$\pm$0.11} & $+15.5$\,{\scriptsize$\pm$1.3}\\
bx/isbn & $-0.96$\,{\scriptsize$\pm$0.20} & $-1.12$\,{\scriptsize$\pm$0.21} & $-0.53$\,{\scriptsize$\pm$0.09} & $-0.95$\,{\scriptsize$\pm$0.22} & $-1.11$\,{\scriptsize$\pm$0.18} & $-0.34$\,{\scriptsize$\pm$0.06} & $-0.48$\,{\scriptsize$\pm$0.05} & $+0.56$\,{\scriptsize$\pm$0.17} & $-0.01$\,{\scriptsize$\pm$0.03} & $+31.5$\,{\scriptsize$\pm$1.7}\\
bx/loc & $-0.04$\,{\scriptsize$\pm$0.04} & $-0.09$\,{\scriptsize$\pm$0.08} & $-0.23$\,{\scriptsize$\pm$0.14} & $-0.24$\,{\scriptsize$\pm$0.14} & $-0.19$\,{\scriptsize$\pm$0.15} & $-0.14$\,{\scriptsize$\pm$0.08} & $-0.13$\,{\scriptsize$\pm$0.10} & $+2.00$\,{\scriptsize$\pm$0.27} & $-0.03$\,{\scriptsize$\pm$0.04} & $+5.19$\,{\scriptsize$\pm$0.61}\\
bx/publisher & $+0.00$\,{\scriptsize$\pm$0.01} & $-0.02$\,{\scriptsize$\pm$0.05} & $-0.04$\,{\scriptsize$\pm$0.11} & $-0.05$\,{\scriptsize$\pm$0.12} & $+0.00$\,{\scriptsize$\pm$0.00} & $+0.00$\,{\scriptsize$\pm$0.02} & $+0.01$\,{\scriptsize$\pm$0.01} & $+0.00$\,{\scriptsize$\pm$0.01} & $-0.08$\,{\scriptsize$\pm$0.10} & $+3.57$\,{\scriptsize$\pm$0.76}\\
bx/user & $-0.06$\,{\scriptsize$\pm$0.06} & $-0.28$\,{\scriptsize$\pm$0.14} & $-0.60$\,{\scriptsize$\pm$0.18} & $-0.69$\,{\scriptsize$\pm$0.17} & $-0.45$\,{\scriptsize$\pm$0.20} & $-0.30$\,{\scriptsize$\pm$0.11} & $-0.40$\,{\scriptsize$\pm$0.15} & $+10.8$\,{\scriptsize$\pm$0.7} & $-0.00$\,{\scriptsize$\pm$0.02} & $+9.41$\,{\scriptsize$\pm$0.89}\\
bx\_rated/isbn & $-0.52$\,{\scriptsize$\pm$0.13} & $-0.80$\,{\scriptsize$\pm$0.15} & $-0.03$\,{\scriptsize$\pm$0.03} & $-0.90$\,{\scriptsize$\pm$0.16} & $-0.13$\,{\scriptsize$\pm$0.02} & $+0.03$\,{\scriptsize$\pm$0.03} & $-0.00$\,{\scriptsize$\pm$0.01} & $-0.03$\,{\scriptsize$\pm$0.02} & $-0.06$\,{\scriptsize$\pm$0.04} & $+33.1$\,{\scriptsize$\pm$1.1}\\
bx\_rated/user & $-6.57$\,{\scriptsize$\pm$0.37} & $-7.90$\,{\scriptsize$\pm$0.40} & $-9.30$\,{\scriptsize$\pm$0.36} & $-9.79$\,{\scriptsize$\pm$0.45} & $-9.57$\,{\scriptsize$\pm$0.47} & $-8.72$\,{\scriptsize$\pm$0.37} & $-8.76$\,{\scriptsize$\pm$0.33} & $+12.9$\,{\scriptsize$\pm$0.5} & $-0.02$\,{\scriptsize$\pm$0.03} & $+2.90$\,{\scriptsize$\pm$0.31}\\
churn/pseudo\_id & $+0.03$\,{\scriptsize$\pm$0.11} & $+0.03$\,{\scriptsize$\pm$0.11} & $-0.06$\,{\scriptsize$\pm$0.47} & $-0.03$\,{\scriptsize$\pm$0.48} & $+0.00$\,{\scriptsize$\pm$0.00} & $+1.27$\,{\scriptsize$\pm$3.31} & $+0.00$\,{\scriptsize$\pm$0.04} & $-0.47$\,{\scriptsize$\pm$1.05} & $+0.00$\,{\scriptsize$\pm$0.02} & $+12.5$\,{\scriptsize$\pm$8.2}\\
churn/state & $+0.00$\,{\scriptsize$\pm$0.00} & $+0.00$\,{\scriptsize$\pm$0.00} & $+0.67$\,{\scriptsize$\pm$1.04} & $+0.67$\,{\scriptsize$\pm$1.04} & $+0.00$\,{\scriptsize$\pm$0.00} & $+0.06$\,{\scriptsize$\pm$0.09} & $+0.05$\,{\scriptsize$\pm$0.09} & $+1.19$\,{\scriptsize$\pm$2.69} & $-0.01$\,{\scriptsize$\pm$0.05} & $+3.46$\,{\scriptsize$\pm$4.43}\\
click\_small/advertiser\_id & $-0.72$\,{\scriptsize$\pm$0.69} & $-0.91$\,{\scriptsize$\pm$0.84} & $-0.98$\,{\scriptsize$\pm$0.67} & $-1.28$\,{\scriptsize$\pm$1.00} & $-1.06$\,{\scriptsize$\pm$0.78} & $-0.81$\,{\scriptsize$\pm$0.87} & $-0.70$\,{\scriptsize$\pm$0.37} & $-0.01$\,{\scriptsize$\pm$0.37} & $-0.01$\,{\scriptsize$\pm$0.15} & $+19.6$\,{\scriptsize$\pm$4.7}\\
click\_small/user\_id & $-0.67$\,{\scriptsize$\pm$0.23} & $-0.67$\,{\scriptsize$\pm$0.24} & $-0.67$\,{\scriptsize$\pm$0.24} & $-0.67$\,{\scriptsize$\pm$0.25} & $-0.03$\,{\scriptsize$\pm$0.20} & $+18.6$\,{\scriptsize$\pm$23.3} & $-0.65$\,{\scriptsize$\pm$0.23} & $+0.01$\,{\scriptsize$\pm$0.20} & $-0.02$\,{\scriptsize$\pm$0.11} & $+0.08$\,{\scriptsize$\pm$0.39}\\
criteo/C1 & $-0.05$\,{\scriptsize$\pm$0.02} & $-0.06$\,{\scriptsize$\pm$0.02} & $-0.08$\,{\scriptsize$\pm$0.02} & $-0.08$\,{\scriptsize$\pm$0.02} & $-0.05$\,{\scriptsize$\pm$0.05} & $+0.32$\,{\scriptsize$\pm$0.03} & $-0.08$\,{\scriptsize$\pm$0.02} & $+0.09$\,{\scriptsize$\pm$0.03} & $-0.00$\,{\scriptsize$\pm$0.00} & $+4.41$\,{\scriptsize$\pm$0.14}\\
criteo/C20 & $-0.05$\,{\scriptsize$\pm$0.02} & $-0.05$\,{\scriptsize$\pm$0.02} & $-0.08$\,{\scriptsize$\pm$0.02} & $-0.08$\,{\scriptsize$\pm$0.02} & $-0.05$\,{\scriptsize$\pm$0.05} & $+0.35$\,{\scriptsize$\pm$0.03} & $-0.09$\,{\scriptsize$\pm$0.02} & $+0.09$\,{\scriptsize$\pm$0.03} & $-0.00$\,{\scriptsize$\pm$0.00} & $+4.38$\,{\scriptsize$\pm$0.15}\\
criteo/C22 & $-0.04$\,{\scriptsize$\pm$0.01} & $-0.04$\,{\scriptsize$\pm$0.01} & $-0.07$\,{\scriptsize$\pm$0.02} & $-0.07$\,{\scriptsize$\pm$0.02} & $+0.00$\,{\scriptsize$\pm$0.00} & $+0.28$\,{\scriptsize$\pm$0.03} & $-0.07$\,{\scriptsize$\pm$0.02} & $+0.10$\,{\scriptsize$\pm$0.03} & $-0.00$\,{\scriptsize$\pm$0.00} & $+4.51$\,{\scriptsize$\pm$0.16}\\
employee\_salaries/position\_title & $-2.26$\,{\scriptsize$\pm$2.76} & $-2.78$\,{\scriptsize$\pm$3.55} & $-4.22$\,{\scriptsize$\pm$2.94} & $-5.14$\,{\scriptsize$\pm$4.55} & $-3.96$\,{\scriptsize$\pm$2.43} & $+4.11$\,{\scriptsize$\pm$5.02} & $-2.30$\,{\scriptsize$\pm$1.99} & $+99.4$\,{\scriptsize$\pm$11.6} & $-0.14$\,{\scriptsize$\pm$0.73} & $+0.17$\,{\scriptsize$\pm$0.30}\\
movielens/movie\_id & $-0.02$\,{\scriptsize$\pm$0.03} & $-0.03$\,{\scriptsize$\pm$0.03} & $-0.08$\,{\scriptsize$\pm$0.06} & $-0.09$\,{\scriptsize$\pm$0.07} & $-0.12$\,{\scriptsize$\pm$0.10} & $-0.11$\,{\scriptsize$\pm$0.05} & $-0.11$\,{\scriptsize$\pm$0.05} & $+1.27$\,{\scriptsize$\pm$0.20} & $-0.00$\,{\scriptsize$\pm$0.01} & $+0.14$\,{\scriptsize$\pm$0.09}\\
movielens/user\_id & $+0.00$\,{\scriptsize$\pm$0.00} & $+0.00$\,{\scriptsize$\pm$0.00} & $-0.02$\,{\scriptsize$\pm$0.04} & $-0.02$\,{\scriptsize$\pm$0.04} & $-0.06$\,{\scriptsize$\pm$0.05} & $-0.05$\,{\scriptsize$\pm$0.03} & $-0.05$\,{\scriptsize$\pm$0.03} & $+0.61$\,{\scriptsize$\pm$0.14} & $+0.00$\,{\scriptsize$\pm$0.00} & $+0.49$\,{\scriptsize$\pm$0.15}\\
\bottomrule
\end{tabular}